\documentclass{article}

\PassOptionsToPackage{numbers, compress}{natbib}

\usepackage[preprint]{neurips_2022}
\def\neuripssubmission{} %

\usepackage[strict]{changepage}
\usepackage{afterpage}
\usepackage[utf8]{inputenc} %
\usepackage[T1]{fontenc}    %
\usepackage{hyperref}       %
\usepackage{fancyvrb}
\usepackage{xcolor}
\definecolor{LightGray}{gray}{0.98}
\usepackage{url}            %
\usepackage{booktabs}       %
\usepackage{amsfonts}       %
\usepackage{nicefrac}       %
\usepackage{microtype}      %
\usepackage{xcolor}         %
\usepackage{xspace}
\usepackage{caption}
\usepackage{multirow}
\usepackage{subcaption}
\usepackage{tablefootnote}
\usepackage{algorithm,algorithmicx,algpseudocode}
\usepackage{tikz}
\usepackage{xfrac}
\usepackage{soul}
\usepackage{placeins}
\usetikzlibrary{arrows.meta}
\usetikzlibrary{shapes,arrows,chains,positioning}
\hypersetup{
    colorlinks=true,
    linkcolor=blue,
    urlcolor=blue,
    citecolor=blue,
}

\usepackage{pgfplots, pgfplotstable}[compat=1.18]

\usepackage{enumitem}
\usepackage{wrapfig}
\usepackage{tabularx}

\newcommand{\name}{Imagen\xspace}

\newcommand{\benchmarkname}{DrawBench\xspace}
\newcommand{\ourmscocofid}{7.27\xspace}
\newcommand{\presectionspace}{-0.25cm}
\newcommand{\postsectionspace}{-0.2cm}

\usepackage{mathtools}
\usepackage[capitalise]{cleveref}
\newcommand{\defeq}{\coloneqq}

\newcommand{\E}{\mathbb{E}}

\newcommand{\Eb}[2]{\E_{#1}\!\left[#2\right]}

\newcommand{\bI}{\mathbf{I}}

\newcommand{\bzero}{\mathbf{0}}

\newcommand{\bc}{\mathbf{c}}

\newcommand{\bx}{\mathbf{x}}

\newcommand{\bz}{\mathbf{z}}

\newcommand{\bepsilon}{{\boldsymbol{\epsilon}}}
\newcommand{\bmu}{{\boldsymbol{\mu}}}

\title{Photorealistic Text-to-Image Diffusion Models \\ with Deep Language Understanding}

\author{%
  Chitwan Saharia\thanks{Equal contribution.}\:,\; William Chan\footnotemark[1]\:,\; Saurabh Saxena\thanks{Core contribution.}\;,\; Lala Li\footnotemark[2]\,,\; Jay Whang\footnotemark[2]\,, \\
  \textbf{Emily Denton,\; Seyed Kamyar Seyed Ghasemipour,\; Burcu Karagol Ayan,} \\
  \textbf{S. Sara Mahdavi,\; Rapha Gontijo Lopes,\; Tim Salimans,} \\
  \textbf{Jonathan Ho\footnotemark[2]\;, David J Fleet\footnotemark[2]\;, Mohammad Norouzi\footnotemark[1]\;} \\[.2cm]
  \texttt{\{sahariac,williamchan,mnorouzi\}@google.com} \\
  \texttt{\{srbs,lala,jwhang,jonathanho,davidfleet\}@google.com}\\[.2cm]
  Google Research, Brain Team \\
  Toronto, Ontario, Canada
}

\begin{document}

\maketitle

\begin{abstract}
We present \name, a text-to-image diffusion model with an unprecedented degree of photorealism and a deep level of language understanding.
\name builds on the power of large transformer language models in understanding text and hinges on the strength of diffusion models in high-fidelity image generation.
Our key discovery is that generic large language models (e.g. T5), pretrained on text-only corpora, are surprisingly
effective at encoding text for image synthesis: increasing the size of the language model in \name boosts both sample fidelity and image-text
alignment much more than increasing the size of the image diffusion model.
\name achieves a new state-of-the-art FID score of \ourmscocofid on the COCO dataset, without ever training on COCO, and human raters find \name samples to be on par with the COCO data itself in image-text alignment.
To assess text-to-image models in greater depth, we introduce \benchmarkname, a comprehensive and challenging benchmark for text-to-image models. 
With \benchmarkname, we compare \name with recent methods including VQ-GAN+CLIP, Latent Diffusion Models, GLIDE and DALL-E~2, and find that human raters prefer \name over other models in side-by-side comparisons, both in terms of sample quality and image-text alignment. See \href{https://imagen.research.google/}{\texttt{imagen.research.google}} for an overview of the results.

\end{abstract}

\vspace{\presectionspace}
\section{Introduction}
\begin{figure}[p]
\vspace*{-2cm}

\newgeometry{left=0cm,top=0cm,right=0cm,bottom=0cm}%
\setlength{\tabcolsep}{2.0pt}
\captionsetup[subfigure]{labelformat=empty}
\hspace*{-7.8cm}
\setlength{\tabcolsep}{2.0pt}
\centering
\begin{tabular}{ccc}
\begin{subfigure}[t]{0.31\textwidth} \centering \includegraphics[width=\textwidth]{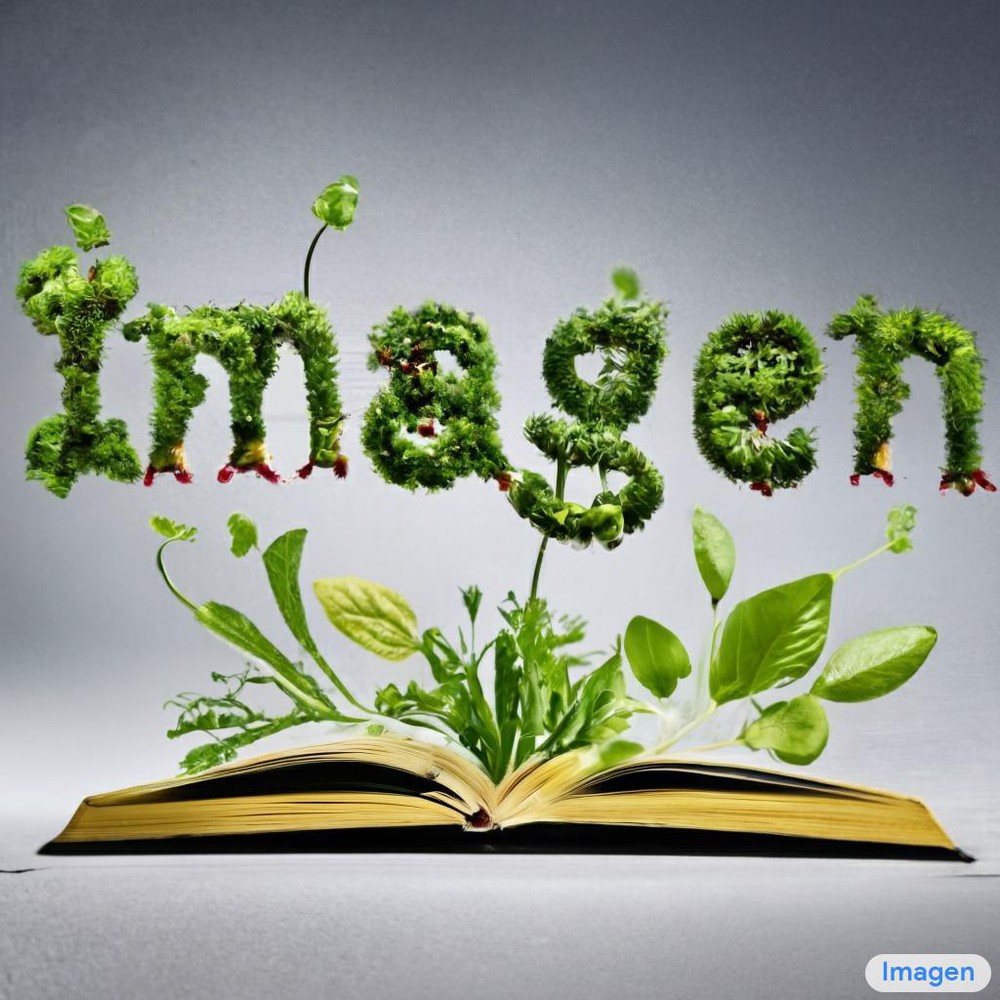} \caption{Sprouts in the shape of text `Imagen' coming out of a fairytale book.} \end{subfigure} &
\begin{subfigure}[t]{0.31\textwidth} \centering \includegraphics[width=\textwidth]{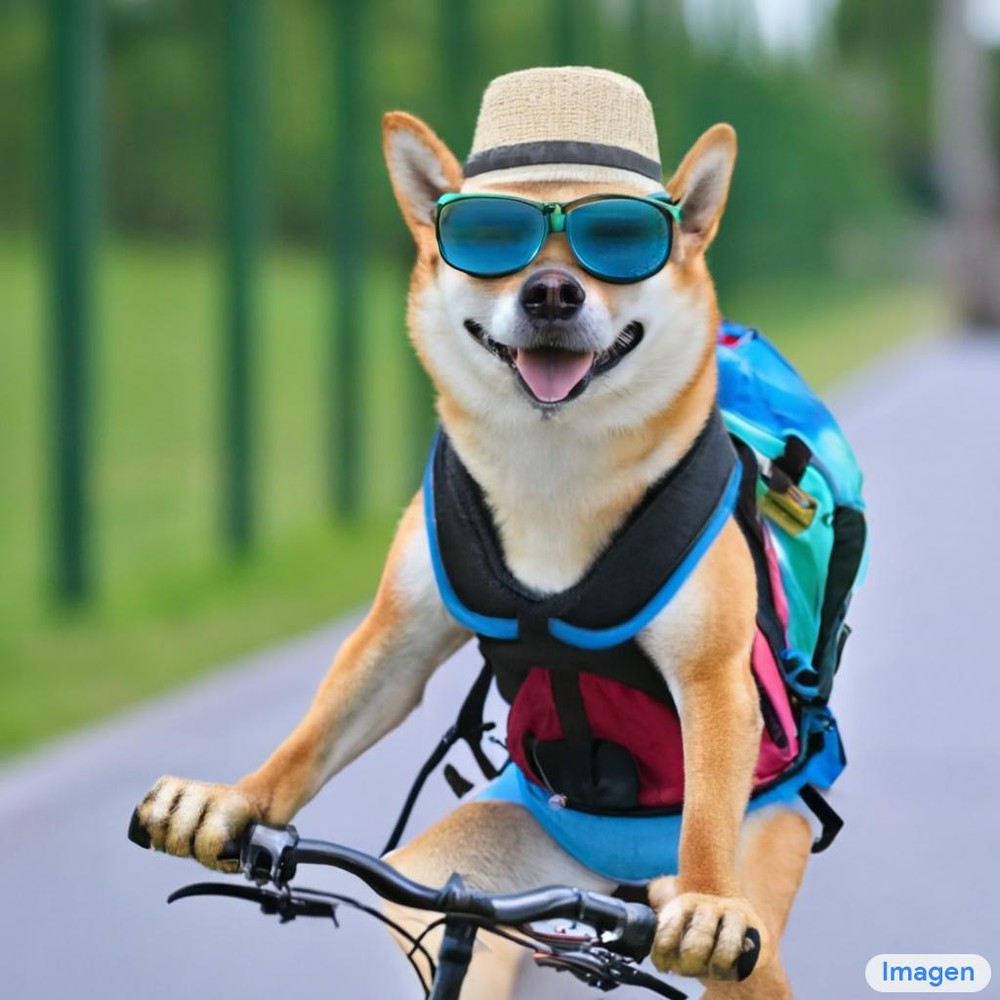} \caption{A photo of a Shiba Inu dog with a backpack riding a bike. It is wearing sunglasses and a beach hat.} \end{subfigure} &
\begin{subfigure}[t]{0.31\textwidth} \centering \includegraphics[width=\textwidth]{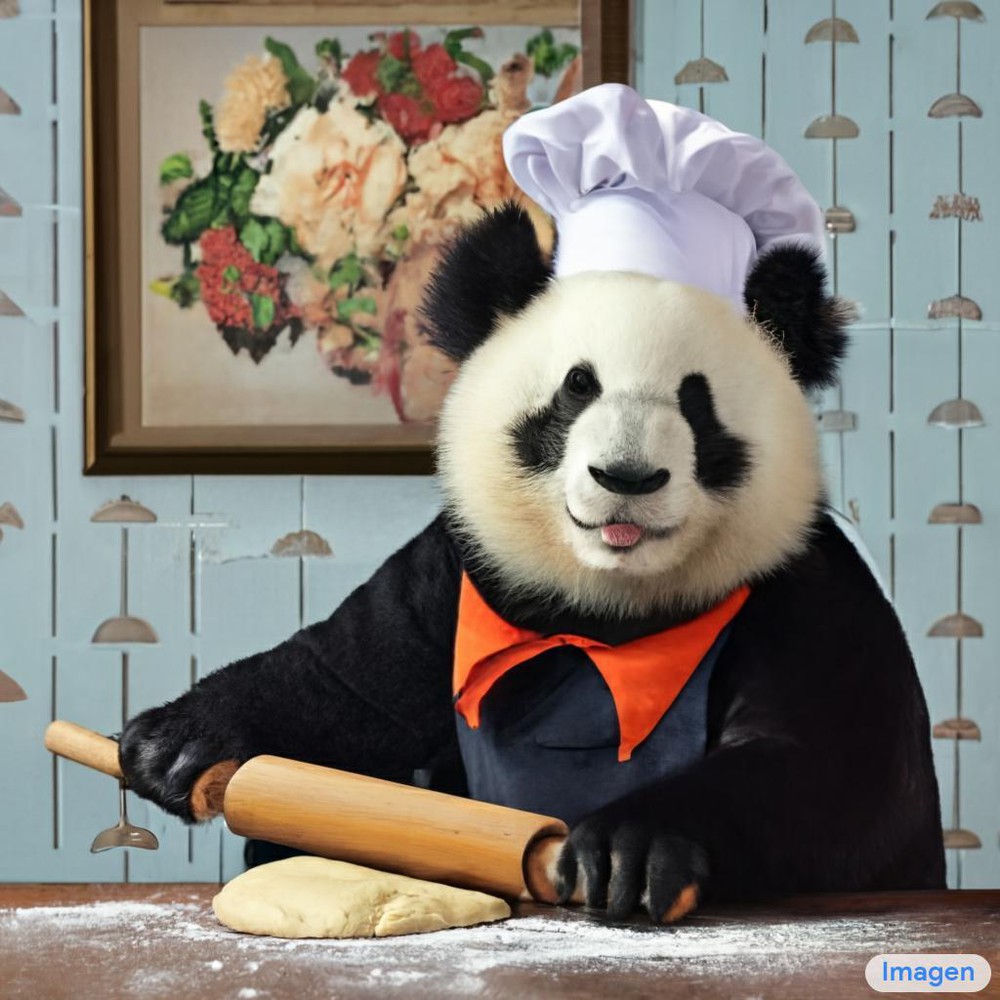} \caption{A high contrast portrait of a very happy fuzzy panda dressed as a chef in a high end kitchen making dough. There is a painting of flowers on the wall behind him.} \end{subfigure} \\
\addlinespace
\begin{subfigure}[t]{0.31\textwidth} \centering \includegraphics[width=\textwidth]{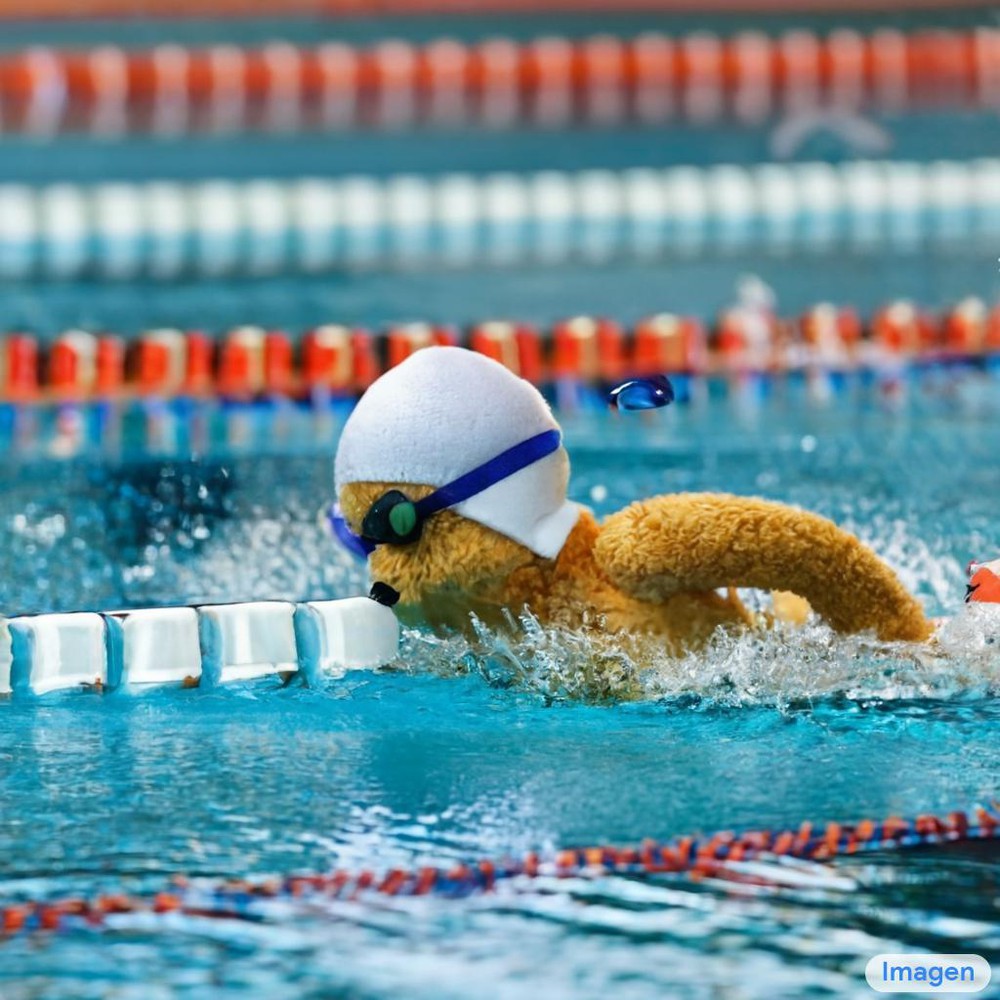} \caption{Teddy bears swimming at the Olympics 400m Butterfly event.} \end{subfigure} &
\begin{subfigure}[t]{0.31\textwidth} \centering \includegraphics[width=\textwidth]{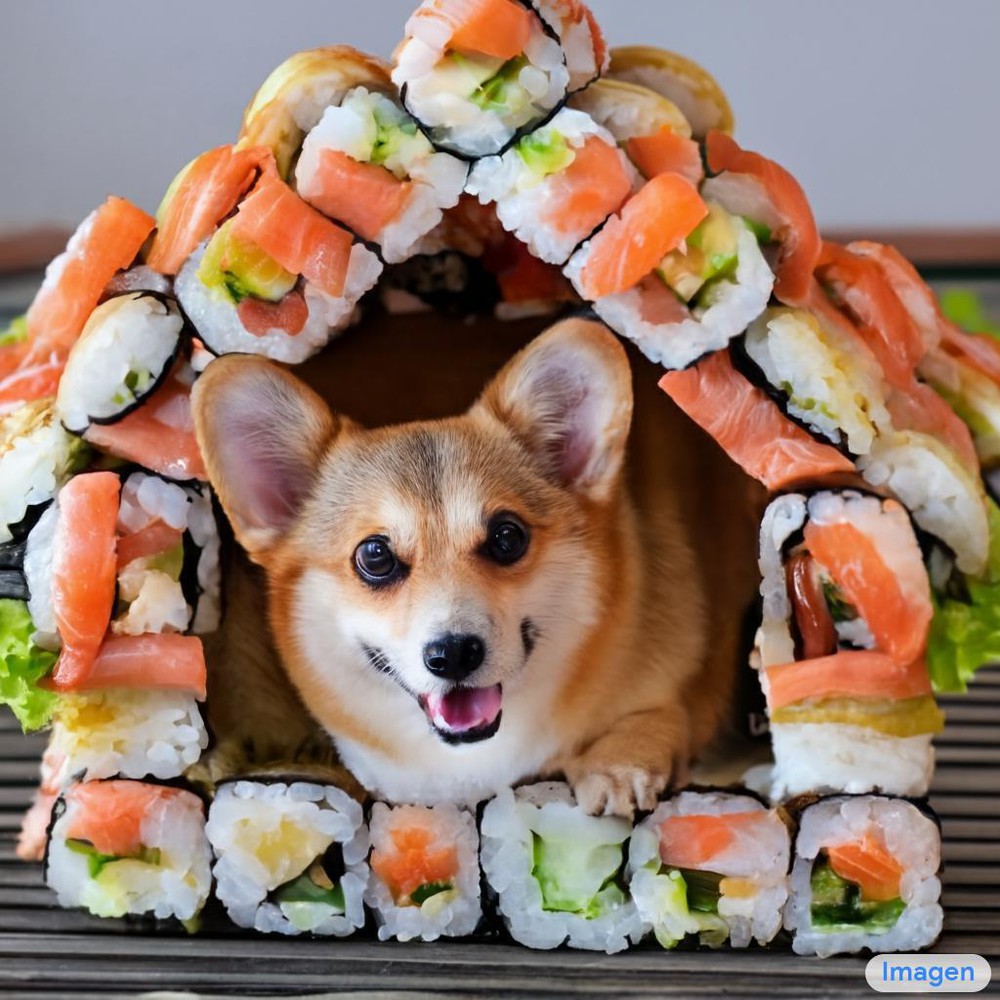} \caption{A cute corgi lives in a house made out of sushi.} \end{subfigure} &
\begin{subfigure}[t]{0.31\textwidth} \centering \includegraphics[width=\textwidth]{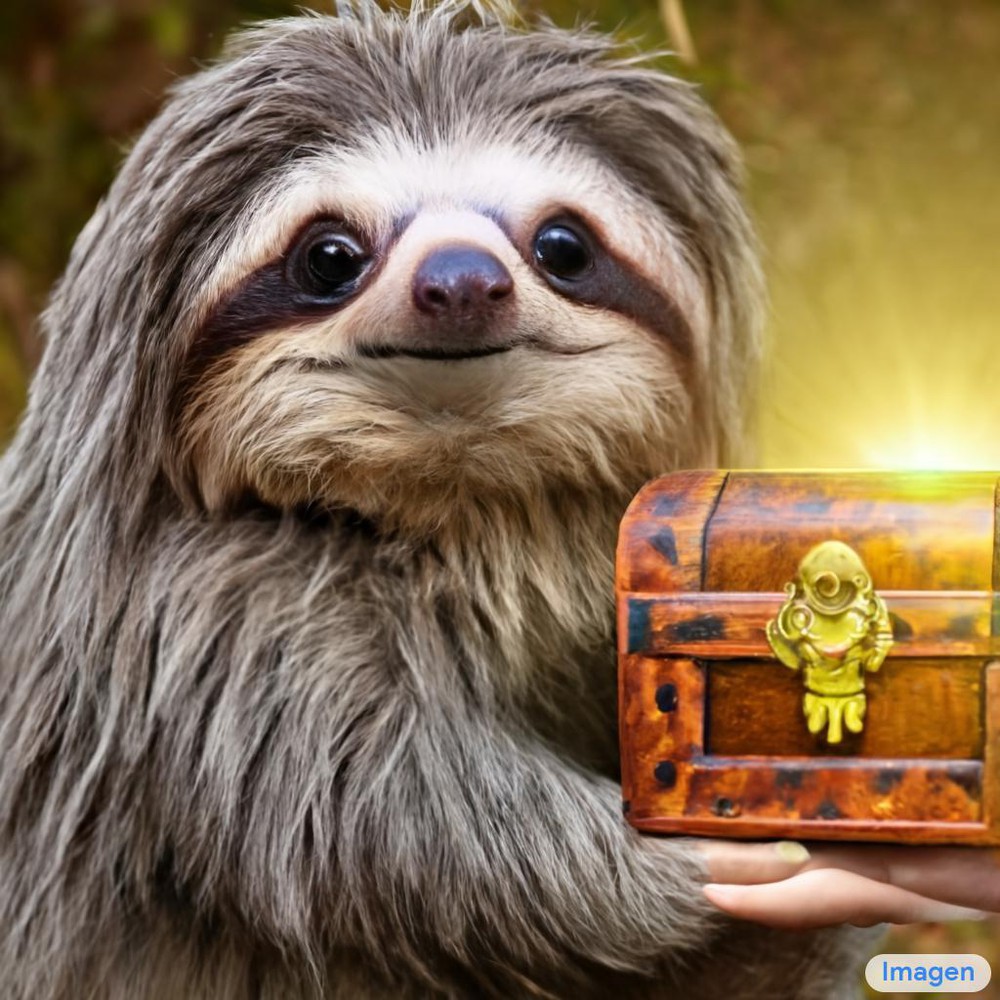} \caption{A cute sloth holding a small treasure chest. A bright golden glow is coming from the chest.} \end{subfigure} \\
\addlinespace
\begin{subfigure}[t]{0.31\textwidth} \centering \includegraphics[width=\textwidth]{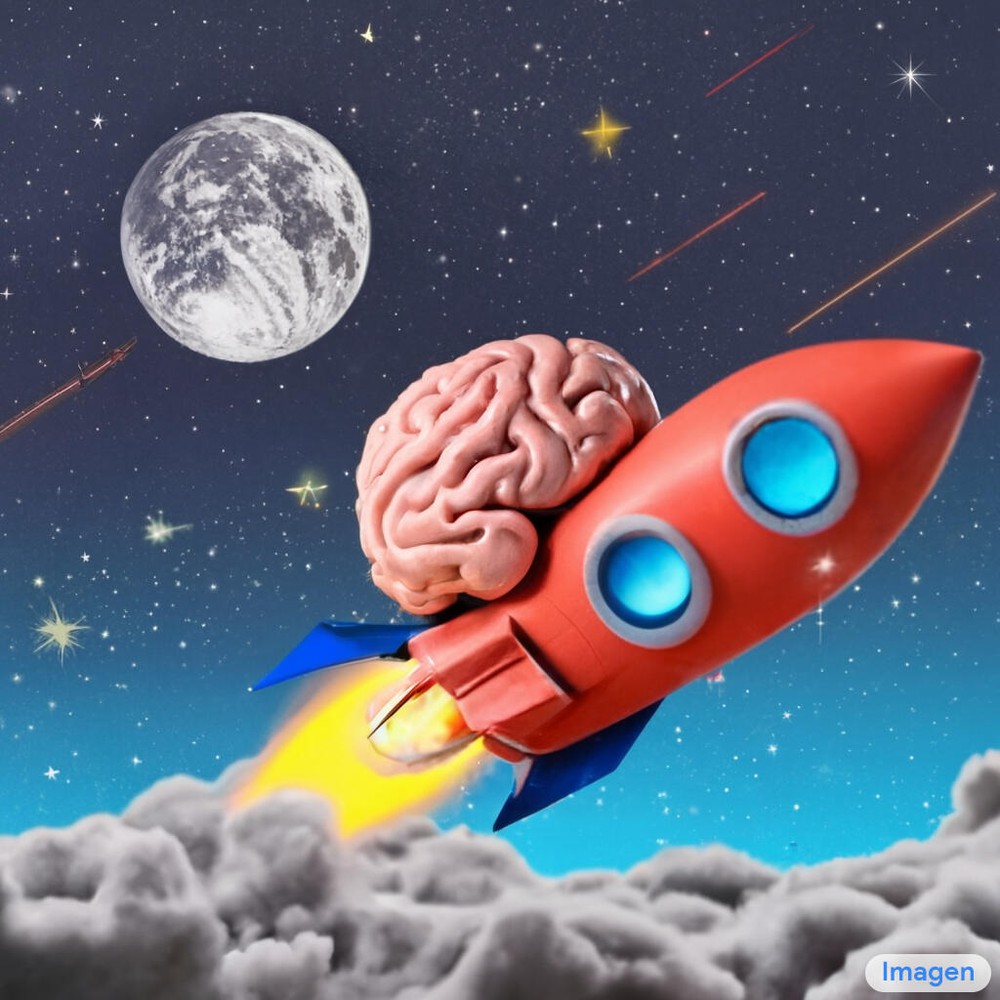} \caption{A brain riding a rocketship heading towards the moon.} \end{subfigure} &
\begin{subfigure}[t]{0.31\textwidth} \centering \includegraphics[width=\textwidth]{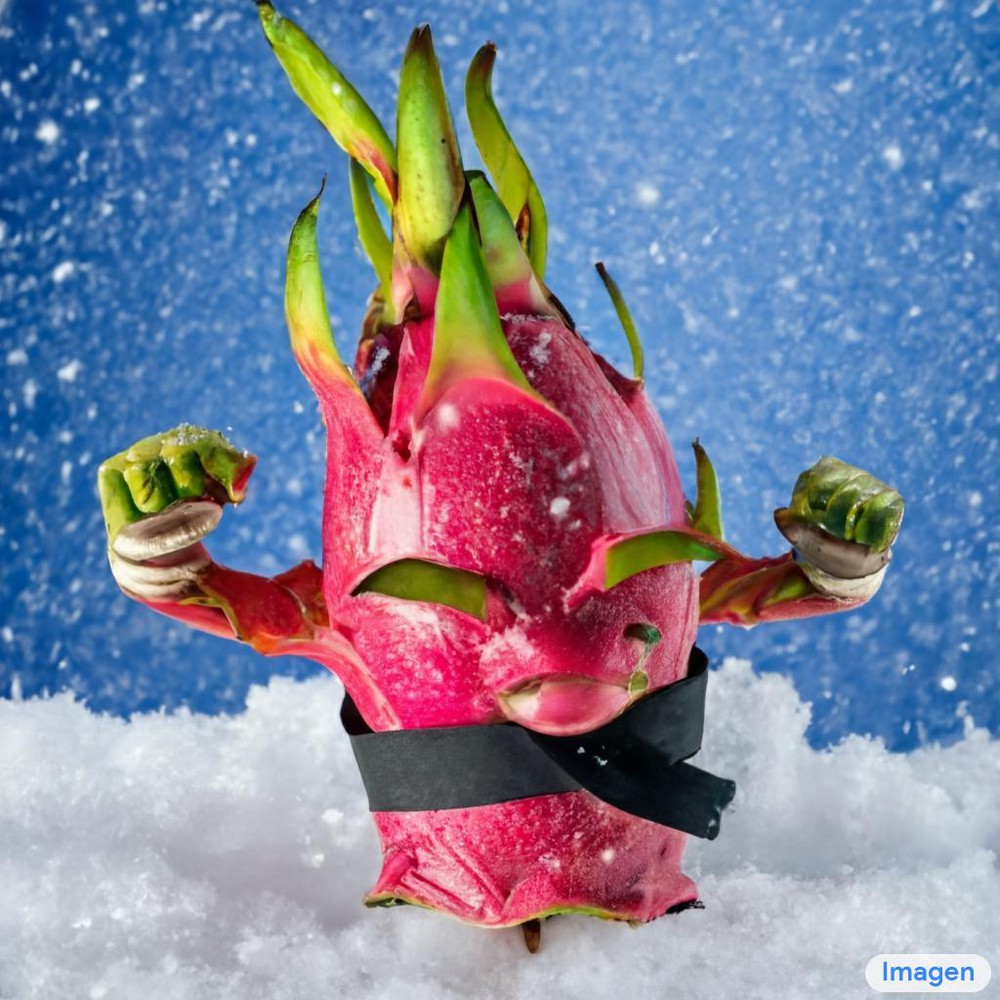} \caption{A dragon fruit wearing karate belt in the snow.} \end{subfigure} &
\begin{subfigure}[t]{0.31\textwidth} \centering \includegraphics[width=\textwidth]{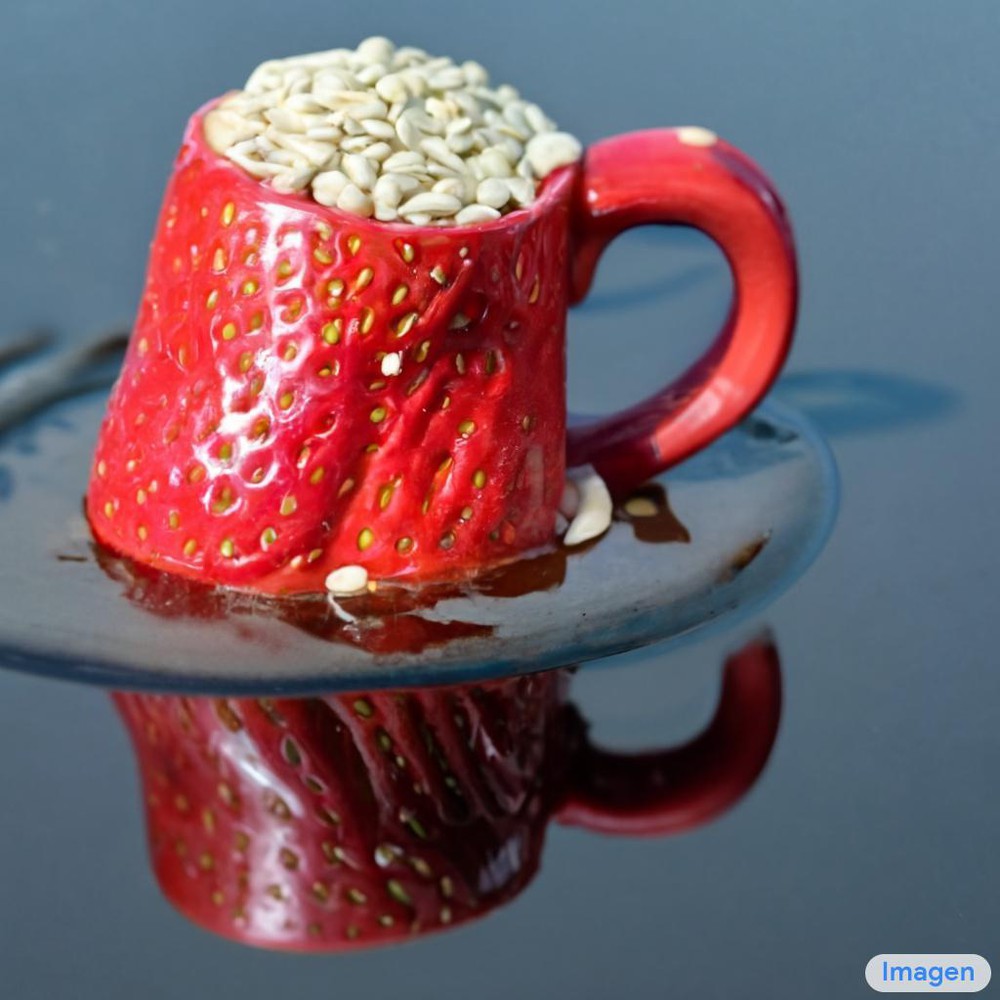} \caption{A strawberry mug filled with white sesame seeds. The mug is floating in a dark chocolate sea.} \end{subfigure}
\end{tabular}
\newgeometry{
textheight=9in,
textwidth=5.5in,
top=1in,
headheight=12pt,
headsep=25pt,
footskip=30pt
}%
\caption{Select $1024\times1024$ \name samples for various text inputs. We only include photorealistic images in this figure and leave artistic content to the Appendix, since generating photorealistic images is more challenging from a technical point of view. \cref{fig:full_page_collage_app1,fig:full_page_collage_app2,fig:full_page_collage_app3} show more samples.}\label{fig:full_page_collage}
\end{figure}

\vspace{\postsectionspace}
Multimodal learning has come into prominence recently, with text-to-image synthesis~\cite{ramesh-dalle, crowson2022vqgan, rombach-cvpr-2022} and image-text contrastive learning~\cite{radford-icml-2021,jia2021scaling,weston2011wsabie} at the forefront.
These models have transformed the research community and captured widespread public attention with creative image generation \cite{gafni2022make, ramesh-dalle2} and editing applications~\cite{fu2021language, nichol-glide, kim2021diffusionclip}.
To pursue this research direction further, we introduce \name, a text-to-image diffusion model that combines the power of transformer language models (LMs)~\cite{devlin-naacl-2019,raffel-jmlr-2020} with high-fidelity diffusion models~\cite{ho2020denoising, ho2021cascaded, dhariwal2021diffusion, nichol-glide} to deliver an unprecedented degree of photorealism and a deep level of language understanding in text-to-image synthesis.
In contrast to prior work that uses only image-text data for model training~\cite[e.g.,][]{ramesh-dalle, nichol-glide}, the key finding behind \name is that text embeddings from large LMs~\cite{raffel-jmlr-2020, devlin-naacl-2019},
pretrained on text-only corpora, are remarkably effective for text-to-image synthesis.
See \cref{fig:full_page_collage} for select samples.

\name comprises a frozen T5-XXL~\cite{raffel-jmlr-2020} encoder to map input text into a sequence of embeddings and a $64\!\times\!64$ image diffusion model, followed by two super-resolution diffusion models for generating $256\!\times\!256$ and $1024\!\times\!1024$ images (see \cref{fig:main_diagram}). All diffusion models are conditioned on the text embedding sequence and use classifier-free guidance \cite{ho2021classifierfree}. \name relies on new sampling techniques to allow usage of large guidance weights without sample quality degradation observed in prior work, resulting in images with higher fidelity and better image-text alignment than previously possible. 

While conceptually simple and easy to train, \name yields surprisingly strong results.  \name outperforms other methods on COCO~\cite{lin-eccv-2014} with zero-shot FID-30K of~\ourmscocofid, significantly outperforming prior work such as GLIDE~\cite{nichol-glide} (at 12.4) and the concurrent work of DALL-E~2~\cite{ramesh-dalle2} (at 10.4). Our zero-shot FID score is also better than state-of-the-art models trained on COCO, e.g., Make-A-Scene~\cite{gafni2022make} (at 7.6). 
Additionally, human raters indicate that generated samples from \name are on-par in image-text alignment to the reference images on COCO captions.

We introduce \benchmarkname, a new structured suite of text prompts for text-to-image evaluation. 
\benchmarkname enables deeper insights through a multi-dimensional evaluation of  text-to-image
models, with text prompts designed to probe different semantic properties of models.  These include compositionality, cardinality, spatial relations, the ability to handle complex text prompts or prompts with rare words, and they include creative prompts that push the limits of models' ability to generate highly implausible scenes well beyond the scope of the training data.
With \benchmarkname, extensive human evaluation shows that \name outperforms other recent methods~\cite{rombach-cvpr-2022, crowson2022vqgan, ramesh-dalle2} by a significant margin. 
We further demonstrate some of the clear advantages of the use of large pre-trained language models~\cite{raffel-jmlr-2020} over multi-modal embeddings such as CLIP~\cite{radford-icml-2021} as a text encoder for \name.

Key contributions of the paper include:
\begin{enumerate}[noitemsep,nolistsep,leftmargin=0.5cm]
    \item We discover that large frozen language models trained only on text data are surprisingly very effective text encoders for text-to-image generation, and that scaling the size of frozen text encoder improves sample quality significantly more than scaling the size of image diffusion model.
    \item We introduce \emph{dynamic thresholding}, a %
    new diffusion sampling technique to leverage high guidance weights and generating more photorealistic and detailed images than previously possible.
    \item We highlight several important diffusion architecture design choices and propose \mbox{\emph{Efficient U-Net}}, a new architecture variant which is simpler, converges faster and is more memory efficient. 
    \item We achieve a new state-of-the-art COCO FID of \ourmscocofid. Human raters find \name to be on-par with the reference images in terms of image-text alignment. 
    \item We introduce \benchmarkname, a new comprehensive and challenging evaluation benchmark for the text-to-image task. On \benchmarkname human evaluation, we find \name to outperform all other work, including the concurrent work of DALL-E 2 \cite{ramesh-dalle2}.
\end{enumerate}

\vspace{-0.25cm}
\section{\name}
\vspace{-0.3cm}
\name consists of a text encoder that maps text to a sequence of embeddings and a cascade of conditional diffusion models that map these embeddings to images of increasing resolutions (see \cref{fig:main_diagram}). In the following subsections, we describe each of these components in detail.

\vspace{-0.3cm}
\subsection{Pretrained text encoders}
\vspace{\postsectionspace}

Text-to-image models need powerful semantic text encoders to capture the complexity and compositionality of arbitrary natural language text inputs. Text encoders trained on paired image-text data are standard in current text-to-image models; they can be trained from scratch \cite{nichol-glide, ramesh-dalle} or pretrained on image-text data \cite{ramesh-dalle2} (e.g., CLIP~\cite{radford-icml-2021}). The image-text training objectives suggest that these text encoders may encode visually semantic and meaningful representations especially relevant for the text-to-image generation task. Large language models can be another models of choice to encode text for text-to-image generation. Recent progress in large language models (e.g., BERT~\cite{devlin-naacl-2019}, GPT~\cite{radford-gpt,radford-gpt2,brown-neurips-2020}, T5~\cite{raffel-jmlr-2020}) have led to leaps in textual understanding and generative capabilities. Language models are trained on text only corpus significantly larger than paired image-text data, thus being exposed to a very rich and wide distribution of text. These models are also generally much larger than text encoders in current image-text models~\cite{radford-icml-2021,jia2021scaling,yu2022coca} (e.g. PaLM \cite{palm-2022} has 540B parameters, while CoCa \cite{yu2022coca} has a $\approx$~1B parameter text encoder).

It thus becomes natural to explore both families of text encoders for the text-to-image task. \name explores pretrained text encoders: BERT \cite{devlin-naacl-2019}, T5 \cite{raffel-icml-2017} and CLIP \cite{radford-arxiv-2017}. For simplicity, we freeze the weights of these text encoders. Freezing has several advantages such as offline computation of embeddings, resulting in negligible computation or memory footprint during training of the text-to-image model. In our work, we find that there is a clear conviction that scaling the text encoder size improves the quality of text-to-image generation. We also find that while T5-XXL and CLIP text encoders perform similarly on simple benchmarks such as MS-COCO, human evaluators prefer T5-XXL encoders over CLIP text encoders in both image-text alignment and image fidelity on \benchmarkname, a set of challenging and compositional prompts. We refer the reader to Section \ref{sec:short_further_analysis} for summary of our findings, and Appendix \ref{sec:frozen_text_encoders} for detailed ablations.

\subsection{Diffusion models and classifier-free guidance}
\vspace{-0.15cm}
Here we give a brief introduction to diffusion models; a precise description is  in~\cref{sec:app_background_diffusion}. Diffusion models \citep{sohl2015deep, ho2020denoising, song2019generative} are a class of generative models that convert Gaussian noise into samples from a learned data distribution via an iterative denoising process. These models can be conditional, for example on class labels, text, or low-resolution images~\citep[e.g.][]{dhariwal2021diffusion,ho2021cascaded,saharia2021image,sahariac-palette,whang2021deblurring,nichol-glide,ramesh-dalle2}.
A diffusion model $\hat\bx_\theta$ is trained on a denoising objective of the form
\begin{align}
    \Eb{\bx,\bc,\bepsilon,t}{w_t \|\hat\bx_\theta(\alpha_t \bx + \sigma_t \bepsilon, \bc) - \bx \|^2_2}
\end{align}
where $(\bx, \bc)$ are data-conditioning pairs, $t \sim \mathcal{U}([0, 1])$, $\bepsilon \sim \mathcal{N}(\bzero, \bI)$, and $\alpha_t, \sigma_t, w_t$ are functions of $t$ that influence sample quality. Intuitively, $\hat\bx_\theta$ is trained to denoise $\bz_t \defeq \alpha_t \bx + \sigma_t \bepsilon$ into $\bx$ using a squared error loss, weighted to emphasize certain values of $t$. 
Sampling such as the ancestral sampler~\citep{ho2020denoising} and DDIM~\citep{song2020denoising} start from pure noise $\bz_1 \sim \mathcal{N}(\bzero, \bI)$ and iteratively generate points $\bz_{t_1}, \dotsc, \bz_{t_T}$, where $1 = t_1 > \cdots > t_T = 0$, that gradually decrease in noise content. These points are functions of the $\bx$-predictions $\hat{\bx}^t_0 \defeq \hat\bx_\theta(\bz_t, \bc)$. %

Classifier guidance \cite{dhariwal2021diffusion} is a technique to improve sample quality while reducing diversity in conditional diffusion models using gradients from a pretrained model $p(\bc|\bz_t)$ during sampling.
\emph{Classifier-free guidance}~\citep{ho2021classifierfree} is an alternative technique that avoids this pretrained model by instead jointly training a single diffusion model on conditional and unconditional objectives via randomly dropping~$\bc$ during training (e.g. with 10\% probability). %
Sampling is performed using the adjusted $\bx$-prediction $(\bz_t - \sigma\tilde\bepsilon_\theta)/\alpha_t$, where
\begin{align}
    \tilde{\bepsilon}_\theta(\bz_t, \bc) = w\bepsilon_\theta(\bz_t, \bc) + (1-w)\bepsilon_{\theta}(\bz_t). \label{eq:classifier_free_score}
\end{align}
Here, $\bepsilon_\theta(\bz_t, \bc)$ and $\bepsilon_{\theta}(\bz_t)$ are conditional and unconditional $\bepsilon$-predictions, given by $\bepsilon_\theta \defeq (\bz_t - \alpha_t\hat\bx_\theta)/\sigma_t$, and $w$ is the \emph{guidance weight}. Setting $w = 1$ disables classifier-free guidance, while increasing $w > 1$ strengthens the effect of guidance. \name depends critically on classifier-free guidance for effective text conditioning.

\subsection{Large guidance weight samplers}
\vspace{-0.15cm}
We corroborate the results of recent text-guided diffusion work~\cite{dhariwal2021diffusion, nichol-glide, ramesh-dalle2} and find that increasing the classifier-free guidance weight improves image-text alignment, but damages image fidelity producing highly saturated and unnatural images~\citep{ho2021classifierfree}.
We find that this is due to a train-test mismatch arising from high guidance weights. At each sampling step $t$, the $\bx$-prediction $\hat{\bx}^t_0$ must be within the same bounds as training data $\bx$, i.e. within $[-1, 1]$, but we find empirically that high guidance weights cause $\bx$-predictions to exceed these bounds. This is a train-test mismatch, and since the diffusion model is iteratively applied on its own output throughout sampling, the sampling process produces unnatural images and sometimes even diverges. To counter this problem, we investigate \emph{static thresholding} and \emph{dynamic thresholding}. See Appendix  \cref{fig:clip_vs_scaled_clip_impl} for reference implementation of the techniques and Appendix \cref{fig:clip_noclip_scaledclip} for visualizations of their effects.

\textbf{Static thresholding}: We refer to elementwise clipping the $\bx$-prediction to $[-1, 1]$ as \emph{static thresholding}. This method was in fact used but not emphasized in previous work~\citep{ho2020denoising}, and to our knowledge its importance has not been investigated in the context of guided sampling. We discover that static thresholding is essential to sampling with large guidance weights and prevents generation of blank images. Nonetheless, static thresholding still results in over-saturated and less detailed images as the guidance weight further increases.

\textbf{Dynamic thresholding}: We introduce a new \emph{dynamic thresholding} method: at each sampling step we set $s$ to a certain percentile absolute pixel value in $\hat{\bx}^t_0$, and if $s>1$, then we threshold $\hat{\bx}^t_0$ to the range $[-s, s]$ and then divide by $s$. Dynamic thresholding pushes saturated pixels (those near -1 and 1) inwards, thereby actively preventing pixels from saturation at each step. We find that dynamic thresholding results in significantly better photorealism as well as better image-text alignment, especially when using very large guidance weights.

\vspace{\presectionspace}
\subsection{Robust cascaded diffusion models}
\vspace{\postsectionspace}
\name utilizes a pipeline of a base $64\times64$ model, and two text-conditional super-resolution diffusion models to upsample a $64\times64$ generated image into a $256\times256$ image, and then to $1024\times1024$ image. Cascaded diffusion models with noise conditioning augmentation \citep{ho2021cascaded} have been extremely effective in progressively generating high-fidelity images. Furthermore, making the super-resolution models aware of the amount of noise added, via noise level conditioning, significantly improves the sample quality and helps improving the robustness of the super-resolution models to handle artifacts generated by lower resolution models \cite{ho2021cascaded}.
\name uses noise conditioning augmentation for both the super-resolution models. We find this to be a critical for generating high fidelity images.

Given a conditioning low-resolution image and augmentation level (a.k.a $\mathrm{aug\_level}$) (e.g., strength of Gaussian noise or blur), we corrupt the low-resolution image with the augmentation (corresponding to $\mathrm{aug\_level}$), and condition the diffusion model on $\mathrm{aug\_level}$. During training, $\mathrm{aug\_level}$ is chosen randomly, while during inference, we sweep over its different values to find the best sample quality. In our case, we use Gaussian noise as a form of augmentation, and apply variance preserving Gaussian noise augmentation resembling the forward process used in diffusion models~(\cref{sec:app_background}). The augmentation level is specified using $\mathrm{aug\_level} \in [0, 1]$. See \cref{fig:cond_aug_impl} for reference pseudocode.

\subsection{Neural network architecture}
\vspace{-0.15cm}
\label{sec:architecture}
\textbf{Base model}: We adapt the U-Net architecture from \cite{nichol2021improved} for our base $64\times64$ text-to-image diffusion model. The network is conditioned on text embeddings via a pooled embedding vector, added to the diffusion timestep embedding similar to the class embedding conditioning method used in \cite{dhariwal2021diffusion, ho2021cascaded}. We further condition on the entire sequence of text embeddings by adding cross attention \cite{rombach-cvpr-2022} over the text embeddings at multiple resolutions. We study various methods of text conditioning in \cref{sec:text_conditioning_ablation}. Furthermore, we found Layer Normalization \cite{ba2016layer} for text embeddings in the attention and pooling layers to help considerably improve performance.

\textbf{Super-resolution models}: For $64\times64 \rightarrow 256\times256$ super-resolution, we use the U-Net model adapted from \cite{nichol2021improved, sahariac-palette}. We make several modifications to this U-Net model for improving memory efficiency, inference time and convergence speed (our variant is 2-3x faster in steps/second over the U-Net used in \cite{nichol2021improved, sahariac-palette}). We call this variant \textit{Efficient U-Net} (See Appendix \ref{sec:unet_vs_efficient_unet_appendix} for more details and comparisons). Our $256\times256 \rightarrow 1024\times1024$ super-resolution model trains on $64\times64 \rightarrow 256\times 256$ crops of the $1024\times1024$ image. To facilitate this, we remove the self-attention layers, however we keep the text cross-attention layers which we found to be critical. During inference, the model receives the full $256\times256$ low-resolution images as inputs, and returns upsampled $1024\times1024$ images as outputs. Note that we use text cross attention for both our super-resolution models.

\begin{figure}[tb]
\centering
\captionsetup[subfigure]{font=scriptsize,labelformat=empty}

\setlength{\tabcolsep}{0.2pt}
\begin{tabular}{cc@{\hspace{.25cm}}cc@{\hspace{.25cm}}cc}

\includegraphics[width=0.16\textwidth]{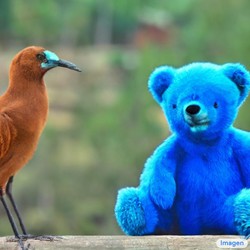} &
\includegraphics[width=0.16\textwidth]{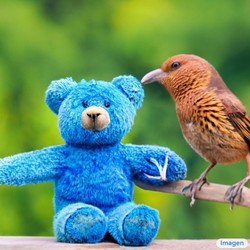} &
\includegraphics[width=0.16\textwidth]{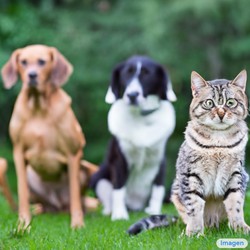} &
\includegraphics[width=0.16\textwidth]{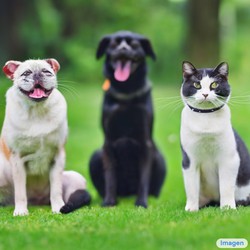} &
\includegraphics[width=0.16\textwidth]{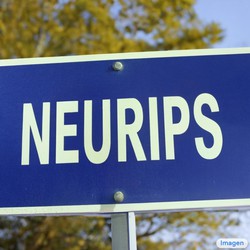} &
\includegraphics[width=0.16\textwidth]{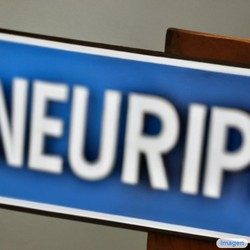} \\ [-4pt]

\multicolumn{2}{c}{\scriptsize{A brown bird and a blue bear.}} &
\multicolumn{2}{c}{\scriptsize{One cat and two dogs sitting on the grass.}} &
\multicolumn{2}{c}{\scriptsize{A sign that says 'NeurIPS'.}} \\[4pt]

\includegraphics[width=0.16\textwidth]{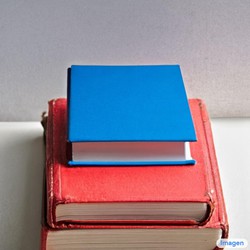} &
\includegraphics[width=0.16\textwidth]{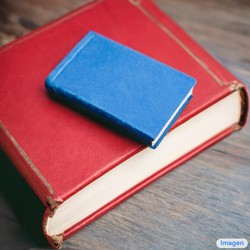} &
\includegraphics[width=0.16\textwidth]{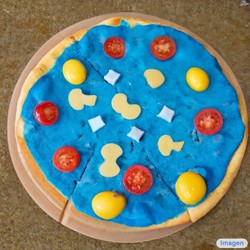} &
\includegraphics[width=0.16\textwidth]{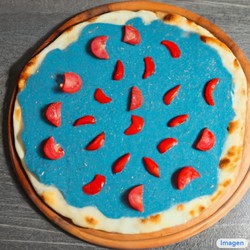} &
\includegraphics[width=0.16\textwidth]{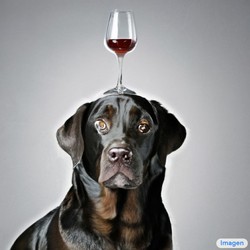} &
\includegraphics[width=0.16\textwidth]{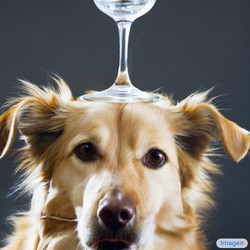} \\[-4pt]

\multicolumn{2}{c}{\scriptsize{A small blue book sitting on a large red book.}} &
\multicolumn{2}{c}{\scriptsize{A blue coloured pizza.}} &
\multicolumn{2}{c}{\scriptsize{A wine glass on top of a dog.}} \\[4pt]

\includegraphics[width=0.16\textwidth]{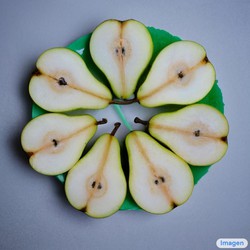} &
\includegraphics[width=0.16\textwidth]{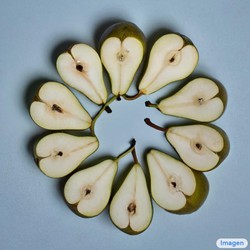} &
\includegraphics[width=0.16\textwidth]{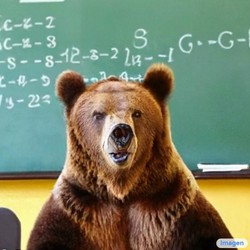} &
\includegraphics[width=0.16\textwidth]{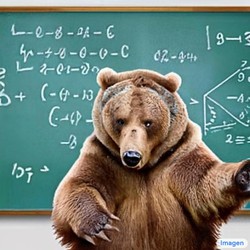} &
\includegraphics[width=0.16\textwidth]{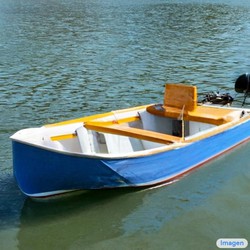} &
\includegraphics[width=0.16\textwidth]{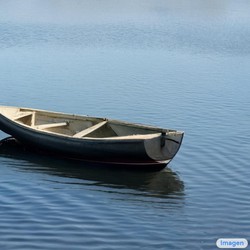} \\[-4pt]

\multicolumn{2}{c}{\scriptsize{A pear cut into seven pieces}} &
\multicolumn{2}{c}{\scriptsize{A photo of a confused grizzly bear}} &
\multicolumn{2}{c}{\scriptsize{A small vessel propelled on water}} \\[-3pt]

\multicolumn{2}{c}{\scriptsize{arranged in a ring.}} &
\multicolumn{2}{c}{\scriptsize{in calculus class.}} &
\multicolumn{2}{c}{\scriptsize{by oars, sails, or an engine.}} \\
\end{tabular}

\vspace{-0.2cm}
\caption{\small Non-cherry picked \name samples for different categories of prompts from \benchmarkname.}
\label{fig:drawbench_samples}
\vspace*{-1em}
\end{figure}

\vspace{\presectionspace}
\section{Evaluating Text-to-Image Models}
\vspace{\postsectionspace}

The COCO \cite{lin-eccv-2014} validation set is the standard benchmark for evaluating text-to-image models for both the supervised \cite{zhou2021lafite, gafni2022make} and the zero-shot setting \cite{ramesh-dalle, nichol-glide}. The key automated performance metrics used are FID \cite{heusel2017gans} to measure image fidelity, and CLIP score \cite{hessel2021clipscore, radford-icml-2021} to measure image-text alignment. Consistent with previous works, we report zero-shot FID-30K, for which 30K prompts are drawn randomly from the validation set, and the model samples generated on these prompts are compared with reference images from the full validation set. Since guidance weight is an important ingredient to control image quality and text alignment, we report most of our ablation results using trade-off (or \textit{pareto}) curves between CLIP and FID scores across a range of guidance weights. 

Both FID and CLIP scores have limitations, for example FID is not fully aligned with perceptual quality~\cite{parmar-cvpr-2022}, and CLIP is ineffective at counting~\cite{radford-icml-2021}. Due to these limitations, we use human evaluation to assess image quality and caption similarity, with ground truth reference caption-image pairs as a baseline. We use two experimental paradigms:
\begin{enumerate}[noitemsep,nolistsep,leftmargin=0.5cm]
    \item To probe image quality, the rater is asked to select between the model generation and reference image using the question: ``Which image is more photorealistic (looks more real)?''. We report the percentage of times raters choose model generations over reference images (the \emph{preference rate}). 
    
    \item To probe alignment, human raters are shown an image and a prompt and asked ``Does the caption accurately describe the above image?''. They must respond with ``yes'', ``somewhat'', or ``no''. These responses are scored as 100, 50, and 0, respectively. These ratings are obtained independently for model samples and  reference images, and both are reported.
\end{enumerate}

For both cases we use 200 randomly chosen image-caption pairs from the COCO validation set. Subjects were shown batches of 50 images. We also used interleaved ``control" trials, and only include rater data from those who correctly answered at least 80\% of the control questions. This netted 73 and 51 ratings per image for image quality and image-text alignment evaluations, respectively.

\textbf{DrawBench}:
While COCO is a valuable benchmark, it is increasingly clear that it has a limited spectrum of prompts that do not readily provide insight into differences between models (e.g., see Sec.\  \ref{sec:experiments_mscoco}). 
Recent work by \citep{dalleval} proposed a new evaluation set called PaintSkills to systematically evaluate visual reasoning skills and social biases beyond COCO. 
With similar motivation, we introduce \emph{\benchmarkname}, a comprehensive and challenging set of prompts that support the evaluation and comparison of text-to-image models.
\benchmarkname contains 11 categories of prompts, testing different capabilities of models such as the ability to faithfully render different colors, numbers of objects, spatial relations, text in the scene, and unusual interactions between objects. Categories also include complex prompts, including long, intricate textual descriptions, rare words, and also misspelled prompts. We also include sets of prompts collected from DALL-E \cite{ramesh-dalle}, Gary Marcus et al. \cite{marcus-arxiv-2022} and \href{https://reddit.com/r/dalle2/}{\color{black}{Reddit}}.
Across these 11 categories,  \benchmarkname comprises 200 prompts in total, striking a good balance between the desire for a large, comprehensive dataset, and small enough that human evaluation remains feasible.
(Appendix \ref{appendix:benchmark} provides a more detailed description of \benchmarkname. \cref{fig:drawbench_samples} shows example prompts from \benchmarkname with \name samples.)

We use \benchmarkname to directly compare different models. To this end, human raters are presented with two sets of images, one from Model A and one from Model B, each of which has 8 samples. Human raters are asked to compare Model A and Model B on sample fidelity and image-text alignment. They respond with one of three choices: Prefer Model A; Indifferent; or Prefer Model B.

\vspace{\presectionspace}
\section{Experiments}
\vspace{\postsectionspace}
\label{sec:experiments}
\Cref{sec:experiments_trainingdetails} describes training details, \cref{sec:experiments_mscoco,sec:experiments_drawbench} analyze results on MS-COCO and  \benchmarkname, and \cref{sec:short_further_analysis} summarizes our ablation studies and key findings. For all experiments below, the images are fair random samples from \name with no post-processing or re-ranking.

\subsection{Training details}
\vspace{-0.15cm}
\label{sec:experiments_trainingdetails}
Unless specified, we train a 2B parameter model for the $64\times64$ text-to-image synthesis, and 600M and 400M parameter models for $64\times64 \rightarrow 256\times256$ and $256\times256 \rightarrow 1024\times1024$ for super-resolution respectively. We use a batch size of 2048 and 2.5M training steps for all models. We use 256 TPU-v4 chips for our base $64\times64$ model, and 128 TPU-v4 chips for both super-resolution models. We do not find over-fitting to be an issue, and we believe further training might improve overall performance. We use Adafactor for our base $64\times64$ model, because initial comparisons with Adam suggested similar performance with much smaller memory footprint for Adafactor. For super-resolution models, we use Adam as we found Adafactor to hurt model quality in our initial ablations. For classifier-free guidance, we joint-train unconditionally via zeroing out the text embeddings with 10\% probability for all three models. We train on a combination of internal datasets, with $\approx$~460M image-text pairs, and the publicly available Laion dataset \cite{schuhmann2021laion}, with $\approx$~400M image-text pairs. There are limitations in our training data, and we refer the reader to \cref{sec:limitations} for details. See \cref{sec:impl_details_appendix} for more implementation details.

\subsection{Results on COCO}
\label{sec:experiments_mscoco}
\begin{table}[t]
\small

\parbox{.55\textwidth}{
\centering
\caption{MS-COCO $256\times256$ FID-30K. We use a guidance weight of 1.35 for our $64\times64$ model, and a guidance weight of 8.0 for our super-resolution model.}
\vspace{.2cm}
\label{tab:zero_shot_mscoco}
\begin{tabular}{lcc}
\toprule
\multirow{2}{*}{\bfseries{Model}} & \multirow{2}{*}{\bfseries{FID-30K}} & \bfseries{Zero-shot}  \\
& &  \bfseries{FID-30K} \\
\midrule
AttnGAN \citep{xu-cvpr-2018} & 35.49 & \\
DM-GAN \citep{zhu2019dm} & 32.64 & \\
DF-GAN \citep{tao2020df} & 21.42 & \\
DM-GAN + CL \citep{ye2021improving} & 20.79 & \\
XMC-GAN \citep{zhang2021cross} & 9.33 & \\
LAFITE \citep{zhou2021lafite} & 8.12 & \\
Make-A-Scene \citep{gafni2022make} & 7.55 & \\
\midrule
DALL-E \citep{ramesh-dalle} & & 17.89  \\
LAFITE \citep{zhou2021lafite} & & 26.94 \\
GLIDE \citep{nichol-glide} & & 12.24 \\
DALL-E 2 \citep{ramesh-dalle2} & & 10.39 \\
\midrule
\textbf{\name (Our Work)} & & \textbf{\ourmscocofid} \\
\bottomrule
\end{tabular}
}
\hfill
\parbox{.42\textwidth}{
\setlength{\tabcolsep}{3pt}
\centering
\caption{COCO $256\times256$ human evaluation comparing model outputs and original images. For the bottom part (no people), we filter out prompts containing one of \texttt{man}, \texttt{men}, \texttt{woman}, \texttt{women}, \texttt{person}, \texttt{people}, \texttt{child}, \texttt{adult}, \texttt{adults}, \texttt{boy}, \texttt{boys}, \texttt{girl}, \texttt{girls}, \texttt{guy}, \texttt{lady}, \texttt{ladies}, \texttt{someone}, \texttt{toddler}, \texttt{(sport) player}, \texttt{workers}, \texttt{spectators}.}
\vspace{.2cm}
\label{tab:mscoco_human_eval}
\begin{tabular}{lcc}
    \toprule
    \bfseries{Model} & \bfseries{Photorealism $\uparrow$} & \bfseries{Alignment $\uparrow$}  \\
    \midrule
    \textit{Original} \\
    \quad Original & 50.0\% & 91.9 $\pm$ 0.42 \\
    \quad \name & 39.5 $\pm$ 0.75\% & 91.4 $\pm$ 0.44 \\ 
    \midrule
    \textit{No people} \\
    \quad Original & 50.0\% & 92.2 $\pm$ 0.54 \\
    \quad \name & 43.9 $\pm$ 1.01\% & 92.1 $\pm$ 0.55 \\
    \bottomrule
    \end{tabular}
}
\end{table}

\label{sec:zero_shot_mscoco}
We evaluate \name on the COCO validation set using FID score, similar to~\citep{ramesh-dalle, nichol-glide}.
\Cref{tab:zero_shot_mscoco} displays the results. \name achieves state of the art \emph{zero-shot} FID on COCO at \ourmscocofid,  outperforming the concurrent work of DALL-E~2~\cite{ramesh-dalle2} and even models trained on COCO.
\Cref{tab:mscoco_human_eval} reports the human evaluation to test image quality and alignment on the COCO validation set. 
We report results on the original COCO validation set, as well as a filtered version in which all reference data with people have been removed. For photorealism, \name achieves 39.2\% preference rate indicating high image quality  generation.  On the set with no people, there is a boost in preference rate of \name to 43.6\%,  indicating \name's limited ability to generate photorealistic people. On caption similarity, \name's score is on-par with the original reference images, suggesting \name's ability to generate images that align well with COCO captions.

\subsection{Results on \benchmarkname}
\vspace{-0.15cm}
\label{sec:experiments_drawbench}

\begin{figure}[tb]
    \centering
    \begin{tabularx}{\linewidth}{XXXX}
        \centering
        \begin{subfigure}[t]{0.25\textwidth}
        \centering
        \hspace*{-9pt}\begin{tikzpicture}

\definecolor{google_blue}{RGB}{66,133,244}
\definecolor{google_red}{RGB}{219,68,55}
\definecolor{google_yellow}{RGB}{194,144,0}
\definecolor{google_green}{RGB}{15,157,88}

\begin{axis}[
name=fidelity,
width=1.3\textwidth,
height=1.35\textwidth,
ybar,
ymin=0.0,
ymax=1.0,
xtick={1, 2},
xtick style={draw=none},
xticklabels={Alignment, Fidelity},
every tick label/.append style={font=\scriptsize},
ytick={0,0.5,1},
yticklabels={0\%, 50\%, 100\%},
legend style={font=\tiny},
legend cell align=left,
legend image code/.code={
    \draw [#1] (0cm, -0.1cm) rectangle (0.2cm, 0.25cm);
},
legend style={
draw=none,
at={(0.5, 1.0)},anchor=south,
legend columns=3,
/tikz/every even column/.append style={column sep=0.2cm}
},
legend image code/.code={
    \draw [#1] (0cm, -0.1cm) rectangle (0.2cm, 0.25cm);
},
bar width=0.4cm,
enlarge x limits={abs=0.4},
]

\addplot+[
    fill=google_blue,
    fill opacity=0.2,
    draw=google_blue,
    error bars/error bar style={google_blue},
    error bars/.cd,
    y dir=both,
    y explicit
] coordinates {
(1.0,0.6198) +- (1.0, 0.030088091985619042)
(2.0,0.6601) +- (2.0, 0.026296085131583567)
};

\addplot+[
fill=google_green,fill opacity=0.2,draw=google_green,
    error bars/error bar style={google_green},
    error bars/.cd,
    y dir=both,
    y explicit
] coordinates {
(1.0,0.3802) +- (1.0, 0.025511315633144172)
(2.0,0.3399) +- (2.0, 0.020835148016479652)
};

\legend{\name, DALL-E 2}
\end{axis}

\end{tikzpicture}    
        \label{fig:drawit_vs_dalle2_drawbench}
        \end{subfigure} & 
        \begin{subfigure}[t]{0.25\textwidth}
        \centering
        \begin{tikzpicture}

\definecolor{google_blue}{RGB}{66,133,244}
\definecolor{google_red}{RGB}{219,68,55}
\definecolor{google_yellow}{RGB}{194,144,0}
\definecolor{google_green}{RGB}{15,157,88}

\begin{axis}[
name=fidelity,
width=1.3\textwidth,
height=1.35\textwidth,
ybar,
ymin=0.0,
ymax=1.0,
xtick={1, 2},
xtick style={draw=none},
xticklabels={Alignment, Fidelity},
every tick label/.append style={font=\scriptsize},
ytick={0,0.5,1},
yticklabels={},
legend style={font=\tiny},
legend cell align=left,
legend image code/.code={
    \draw [#1] (0cm, -0.1cm) rectangle (0.2cm, 0.25cm);
},
legend style={
draw=none,
at={(0.5, 1.0)},anchor=south,
legend columns=3,
/tikz/every even column/.append style={column sep=0.2cm}
},
bar width=0.4cm,
enlarge x limits={abs=0.4},
]
\addplot+[
    fill=google_blue,
    fill opacity=0.2,
    draw=google_blue,
    error bars/error bar style={google_blue},
    error bars/.cd,
    y dir=both,
    y explicit
] coordinates {
(1.0,0.5945) +- (1.0, 0.02145839465123118)
(2.0,0.604) +- (2.0, 0.02598245656099196)
};

\addplot+[
fill=google_green,fill opacity=0.2,draw=google_green,
    error bars/error bar style={google_green},
    error bars/.cd,
    y dir=both,
    y explicit
] coordinates {
(1.0,0.405) +- (1.0, 0.021982563782935984)
(2.0,0.396) +- (2.0, 0.02406971888310527)
};

\legend{\name, GLIDE}
\end{axis}

\end{tikzpicture}    
        \label{fig:drawit_vs_glide_drawbench}
        \end{subfigure} & 
        \begin{subfigure}[t]{0.25\textwidth}
        \centering
        \begin{tikzpicture}

\definecolor{google_blue}{RGB}{66,133,244}
\definecolor{google_red}{RGB}{219,68,55}
\definecolor{google_yellow}{RGB}{194,144,0}
\definecolor{google_green}{RGB}{15,157,88}

\begin{axis}[
name=fidelity,
width=1.3\textwidth,
height=1.35\textwidth,
ybar,
ymin=0.0,
ymax=1.0,
xtick={1, 2},
xtick style={draw=none},
xticklabels={Alignment, Fidelity},
every tick label/.append style={font=\scriptsize},
ytick={0,0.5,1},
yticklabels={},
legend style={font=\tiny},
legend cell align=left,
legend image code/.code={
    \draw [#1] (0cm, -0.1cm) rectangle (0.2cm, 0.25cm);
},
legend style={
draw=none,
at={(0.5, 1.0)},anchor=south,
legend columns=3,
/tikz/every even column/.append style={column sep=0.2cm}
},
bar width=0.4cm,
enlarge x limits={abs=0.4},
]

\addplot+[
    fill=google_blue,
    fill opacity=0.2,
    draw=google_blue,
    error bars/error bar style={google_blue},
    error bars/.cd,
    y dir=both,
    y explicit
] coordinates {
(1.0,0.7603) +- (1.0, 0.03168142241293373)
(2.0,0.7855) +- (2.0, 0.02625504163936866)
};

\addplot+[
fill=google_green,fill opacity=0.2,draw=google_green,
    error bars/error bar style={google_green},
    error bars/.cd,
    y dir=both,
    y explicit
] coordinates {
(1.0,0.2396) +- (1.0, 0.025106797640769264)
(2.0,0.2145) +- (2.0, 0.02559702857990766)
};

\legend{\name, VQGAN+CLIP}
\end{axis}

\end{tikzpicture}
        \label{fig:drawit_vs_vaganclip_drawbench}
        \end{subfigure} &
        \begin{subfigure}[t]{0.25\textwidth}
        \centering
        \begin{tikzpicture}

\definecolor{google_blue}{RGB}{66,133,244}
\definecolor{google_red}{RGB}{219,68,55}
\definecolor{google_yellow}{RGB}{194,144,0}
\definecolor{google_green}{RGB}{15,157,88}

\begin{axis}[
name=fidelity,
width=1.3\textwidth,
height=1.35\textwidth,
ybar,
ymin=0.0,
ymax=1.0,
xtick={1, 2},
xtick style={draw=none},
xticklabels={Alignment, Fidelity},
every tick label/.append style={font=\scriptsize},
ytick={0,0.5,1},
yticklabels={},
legend style={font=\tiny},
legend cell align=left,
legend image code/.code={
    \draw [#1] (0cm, -0.1cm) rectangle (0.2cm, 0.25cm);
},
legend style={
draw=none,
at={(0.5, 1.0)},anchor=south,
legend columns=3,
/tikz/every even column/.append style={column sep=0.2cm}
},
bar width=0.4cm,
enlarge x limits={abs=0.4},
]

\addplot+[
    fill=google_blue,
    fill opacity=0.2,
    draw=google_blue,
    error bars/error bar style={google_blue},
    error bars/.cd,
    y dir=both,
    y explicit
] coordinates {
(1.0,0.7837) +- (1.0, 0.028497994689556643)
(2.0,0.6652) +- (2.0, 0.02454522731413525)
};

\addplot+[
fill=google_green,fill opacity=0.2,draw=google_green,
    error bars/error bar style={google_green},
    error bars/.cd,
    y dir=both,
    y explicit
] coordinates {
(1.0,0.2136) +- (1.0, 0.01993785825157147)
(2.0,0.3348) +- (2.0, 0.025427213103106437)
};

\legend{\name, Latent Diffusion}
\end{axis}

\end{tikzpicture}
        \label{fig:drawit_vs_latentdiffusion_drawbench}
        \end{subfigure} \\
    \end{tabularx}
    \caption{Comparison between \name and DALL-E 2~\citep{ramesh-dalle2}, GLIDE~\citep{nichol-glide}, VQ-GAN+CLIP~\citep{crowson2022vqgan} and Latent Diffusion~\citep{rombach-cvpr-2022} on \benchmarkname: User preference rates (with 95\% confidence intervals) for image-text alignment and image fidelity.}
    \label{fig:drawbench-comparison}
    \vspace{-1em}
\end{figure}
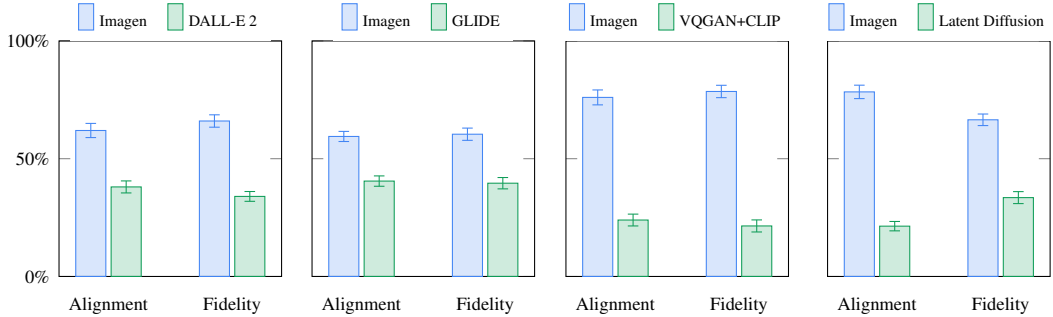

Using \benchmarkname, we compare \name with \mbox{DALL-E 2} (the public version)~\cite{ramesh-dalle2}, \mbox{GLIDE}~\cite{nichol-glide}, \href{https://github.com/CompVis/latent-diffusion}{\color{black}{Latent Diffusion}}~\cite{rombach-cvpr-2022}, and \href{https://github.com/EleutherAI/vqgan-clip}{\color{black}{CLIP-guided VQ-GAN}}~\cite{crowson2022vqgan}. %
\cref{fig:drawbench-comparison} shows the human evaluation results for pairwise comparison of \name with each of the three models. We report the percentage of time raters prefer Model A, Model B, or are indifferent for both image fidelity and image-text alignment. We aggregate the scores across all the categories and raters. We find the human raters to exceedingly prefer \name over all others models in both image-text alignment and image fidelity. We refer the reader to \cref{sec:benchmark_comparison} for a more detailed category wise comparison and qualitative comparison.

\subsection{Analysis of \name}
\begin{figure}[t]
    \centering
        \begin{subfigure}[t]{0.33\textwidth}
        \centering
        \begin{tikzpicture}[scale=0.55]
\definecolor{google_blue}{RGB}{66,133,244}
\definecolor{google_red}{RGB}{219,68,55}
\definecolor{google_yellow}{RGB}{194,144,0}
\definecolor{google_green}{RGB}{15,157,88}

\begin{axis}[
width=8cm,
height=6cm,
  xlabel=CLIP Score,
  ylabel=FID-10K,
  legend columns=1,
  legend pos=north west,
  legend cell align=left,
  y label style={at={(axis description cs:0.03,.5)}},
  legend style={font=\footnotesize},
  xmin=0.216,
  ymin=7.5,
  ymax=29,
  xmax=0.295,
  every axis plot/.append style={very thick},
  yticklabel style={xshift=-5pt}
]

\addplot +[mark=*, solid, google_yellow, mark options={scale=0.4}] coordinates {(0.2186, 17.1279)(0.2272, 14.9052)(0.2332, 14.4287)(0.2379, 14.0655)(0.2418, 14.1945)(0.2499, 16.339)(0.2543, 18.8731)(0.2563, 21.2413)(0.2586, 22.5724)(0.2596, 23.7404)(0.2604, 24.7911)(0.2609, 25.5612)(0.2615, 26.0648)};

\addplot +[mark=*, solid, google_green, mark options={scale=0.4}] coordinates {(0.2242, 15.7681)(0.2338, 13.4837)(0.2412, 12.5508)(0.2461, 12.2109)(0.25, 12.7537)(0.2594, 15.1684)(0.2638, 17.8411)(0.2666, 19.9616)(0.2683, 21.6277)(0.2691, 22.8355)(0.2701, 23.7544)(0.271, 24.3638)(0.2714, 25.1119)};

\addplot +[mark=*, solid, google_red, mark options={scale=0.4}] coordinates {(0.2316, 14.4068)(0.2422, 12.2246)(0.2499, 11.2804)(0.2552, 11.3188)(0.2593, 11.6776)(0.2685, 15.0103)(0.2731, 17.9712)(0.2749, 20.454)(0.2767, 22.132)(0.2774, 23.5054)(0.2783, 24.6424)(0.2788, 25.485)(0.2792, 26.1941)};

\addplot +[mark=*, solid, google_blue, mark options={scale=0.4}] coordinates {(0.2392, 13.33)(0.2508, 11.0229)(0.2584, 10.688)(0.2637, 11.2378)(0.268, 12.0375)(0.2764, 16.3064)(0.2797, 19.7633)(0.2822, 22.2284)(0.283, 24.3269)(0.2836, 25.5622)(0.2843, 26.6161)(0.2845, 27.1637)(0.2847, 28.0816)};

\legend{T5-Small, T-Large, T5-XL, T5-XXL}
\end{axis}
\end{tikzpicture}
        \caption{Impact of encoder size.}
        \label{fig:encoder_size_main_paper}
        \end{subfigure}%
        \begin{subfigure}[t]{0.33\textwidth}
        \centering
        \begin{tikzpicture}[scale=0.55]

\definecolor{google_blue}{RGB}{66,133,244}
\definecolor{google_red}{RGB}{219,68,55}
\definecolor{google_yellow}{RGB}{194,144,0}
\definecolor{google_green}{RGB}{15,157,88}

\begin{axis}[width=8cm, height=6cm,
  xlabel=CLIP Score,
  ylabel=FID-10K,
  legend columns=1,
  legend pos=north west,
  y label style={at={(axis description cs:0.03,.5)}},
  legend style={font=\footnotesize},
  xmin=0.24,
  ymin=7.5,
  ymax=29,
  xmax=0.295,
  every axis plot/.append style={very thick},
]
\addplot +[mark=*, solid, google_blue, mark options={scale=0.4}] coordinates {(0.2451, 11.3483)(0.2572, 9.4933)(0.2652, 9.4017)(0.2699, 10.2998)(0.2738, 11.3422)(0.281, 16.4762)(0.2837, 19.563)(0.285, 21.9303)(0.286, 23.5618)(0.2862, 24.8776)(0.2865, 25.403)(0.2862, 26.2234)(0.2865, 26.5037)};
\addplot +[mark=*, solid, google_red, mark options={scale=0.4}] coordinates {(0.2479, 11.1512)(0.2599, 9.4931)(0.2672, 9.7014)(0.2721, 10.7983)(0.2756, 11.8326)(0.2822, 17.2267)(0.2847, 20.8375)(0.2861, 22.856)(0.2867, 24.6328)(0.2869, 25.723)(0.2869, 26.5768)(0.2868, 26.8506)(0.2873, 27.3912)};
\addplot +[mark=*, solid, google_green, mark options={scale=0.4}] coordinates {(0.2511, 10.4663)(0.2627, 9.4659)(0.27, 9.9582)(0.2748, 11.2437)(0.2781, 12.6562)(0.2836, 17.3972)(0.2863, 20.673)(0.2869, 22.4689)(0.2873, 23.7)(0.2873, 24.6056)(0.2876, 25.4052)(0.2876, 26.0697)(0.2876, 26.2802)};
\addplot +[mark=*, solid, google_yellow, mark options={scale=0.4}] coordinates {(0.2535, 10.1896)(0.2645, 9.3721)(0.2715, 9.9115)(0.2761, 11.2531)(0.28, 12.6531)(0.2844, 17.0384)(0.2866, 20.061)(0.2873, 21.9676)(0.2879, 23.0162)(0.2881, 23.936)(0.288, 24.1819)(0.288, 25)(0.2879, 25.5056)};

\legend{300M, 500M, 1B, 2B}
\end{axis}
\end{tikzpicture}
        \caption{Impact of U-Net size.}
        \label{fig:unet_main_paper}
        \end{subfigure}%
        \begin{subfigure}[t]{0.33\textwidth}
        \centering
        \begin{tikzpicture}[scale=0.55]

\definecolor{google_blue}{RGB}{66,133,244}
\definecolor{google_red}{RGB}{219,68,55}
\definecolor{google_yellow}{RGB}{194,144,0}
\definecolor{google_green}{RGB}{15,157,88}

\begin{axis}[
width=8cm,
height=6cm,
  xlabel=CLIP Score,
  ylabel=FID@10K,
  legend columns=1,
  legend pos=north west,
  y label style={at={(axis description cs:0.03,.5)}},
  legend style={font=\footnotesize},
  legend cell align=left,
  xmin=0.255,
  ymin=7.5,
  ymax=27,
  xmax=0.295,
  every axis plot/.append style={very thick}
]
\addplot +[mark=*, solid, google_blue, mark options={scale=0.4}] coordinates {(0.2573, 9.5976)(0.2686, 8.3913)(0.2755, 9.1741)(0.2794, 10.3238)(0.282, 11.6418)(0.2867, 15.9009)(0.2881, 18.2998)(0.288, 19.4781)(0.2881, 20.3822)(0.2877, 20.9476)(0.2872, 21.4995)(0.2869, 22.526)(0.2866, 23.1381)};

\addplot +[mark=*, solid, google_red, mark options={scale=0.4}] coordinates {(0.2574, 9.2972)(0.2692, 8.0746)(0.2756, 8.8401)(0.2798, 10.0002)(0.2827, 11.3142)(0.2884, 14.7455)(0.2903, 16.8102)(0.2913, 18.092)(0.2921, 19.6666)(0.292, 21.2374)(0.2923, 22.3307)(0.292, 24.3588)(0.2918, 25.9931)};

\legend{static thresholding, dynamic thresholding}
\end{axis}
\end{tikzpicture}
        \caption{Impact of thresholding.}
        \label{fig:clip_vs_scaled_clip_main_paper}
        \end{subfigure}
    \caption{Summary of some of the critical findings of \name with pareto curves sweeping over different guidance values. See \cref{sec:experiments_analysis} for more details.}
    \label{fig:main_ablations}
    \vspace{-1em}
\end{figure}
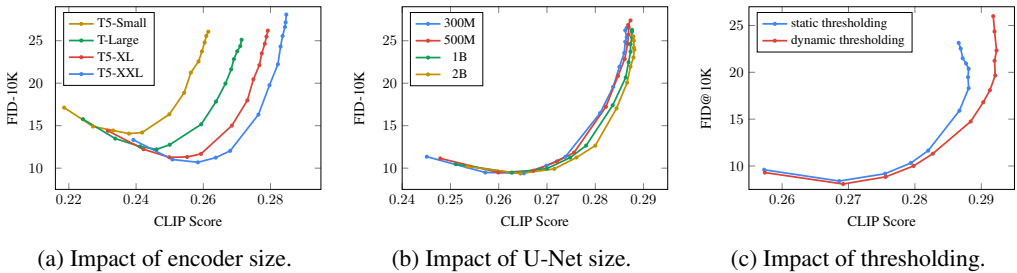

\vspace{-0.15cm}
\label{sec:short_further_analysis}
For a detailed analysis of \name see \cref{sec:experiments_analysis}. Key findings are discussed in \cref{fig:main_ablations} and below.

\textbf{Scaling text encoder size is extremely effective.} We observe that scaling the size of the text encoder leads to consistent improvement in both image-text alignment and image fidelity. \name trained with our largest text encoder, T5-XXL (4.6B parameters), yields the best results (\cref{fig:encoder_size_main_paper}).

\textbf{Scaling text encoder size is more important than U-Net size.} While scaling the size of the diffusion model U-Net improves sample quality, we found scaling the text encoder size to be significantly more impactful than the U-Net size (\cref{fig:unet_main_paper}).

\textbf{Dynamic thresholding is critical.} We show that dynamic thresholding results in samples with significantly better photorealism and alignment with text, over static or no thresholding, especially under the presence of large classifier-free guidance weights (\cref{fig:clip_vs_scaled_clip_main_paper}).

\textbf{Human raters prefer T5-XXL over CLIP on \benchmarkname.} The models trained with T5-XXL and CLIP text encoders perform similarly on the COCO validation set in terms of CLIP and FID scores. However, we find that human raters prefer T5-XXL over CLIP on \benchmarkname across all 11 categories.

\textbf{Noise conditioning augmentation is critical.} We show that training the super-resolution models with noise conditioning augmentation leads to better CLIP and FID scores. We also show that noise conditioning augmentation enables stronger text conditioning for the super-resolution model, resulting in improved CLIP and FID scores at higher guidance weights. Adding noise to the low-res image during inference along with the use of large guidance weights allows the super-resolution models to generate diverse upsampled outputs while removing artifacts from the low-res image.

\textbf{Text conditioning method is critical.} We observe that conditioning over the sequence of text embeddings with cross attention significantly outperforms simple mean or attention based pooling in both sample fidelity as well as image-text alignment.

\textbf{Efficient U-Net is critical.} Our Efficient U-Net implementation uses less memory, converges faster, and has better sample quality with faster inference.

\vspace{\presectionspace}
\section{Related Work}
\vspace{\postsectionspace}
Diffusion models have seen wide success in image generation~\cite{ho2020denoising,nichol2021improved,saharia2021image,dhariwal2021diffusion,ho2021cascaded,sahariac-palette}, outperforming GANs in fidelity and diversity, without training instability and mode collapse issues~\cite{brock2018large,dhariwal2021diffusion,ho2021cascaded}.
Autoregressive models \cite{mansimov-iclr-2016}, GANs \cite{xu-cvpr-2018,zhang2021cross}, VQ-VAE Transformer-based methods~\cite{ramesh-dalle,gafni2022make}, and diffusion models have seen remarkable progress in text-to-image~\cite{rombach-cvpr-2022,nichol-glide,rombach-cvpr-2022}, including the concurrent \mbox{DALL-E 2}~\cite{ramesh-dalle2}, which uses a diffusion prior on CLIP text latents and cascaded diffusion models to generate high resolution $1024\times1024$ images; we believe \name is much simpler, as \name does not need to learn a latent prior, yet achieves better results in both MS-COCO FID and human evaluation on \benchmarkname.
GLIDE \cite{nichol-glide} also uses cascaded diffusion models for text-to-image, but we use large pretrained frozen language models, which we found to be instrumental to both image fidelity and image-text alignment.
XMC-GAN \cite{zhang2021cross} also uses BERT as a text encoder, but we scale to much larger text encoders and demonstrate the effectiveness thereof.
The use of cascaded models is also popular throughout the literature \cite{denton-nips-2015,menick-iclr-2019} and has been used with success in diffusion models to generate high resolution images \cite{dhariwal2021diffusion,ho2021cascaded}. %

\vspace{\presectionspace}
\section{Conclusions, Limitations and Societal Impact}
\label{sec:limitations}
\vspace{\postsectionspace}

\name showcases the effectiveness of frozen large pretrained language models as text encoders for the text-to-image generation using diffusion models. Our observation that scaling the size of these language models have significantly more impact than scaling the U-Net size on overall performance encourages future research directions on exploring even bigger language models as text encoders. Furthermore, through \name we re-emphasize the importance of classifier-free guidance, and we introduce dynamic thresholding, which allows usage of much higher guidance weights than seen in previous works. With these novel components, \name produces $1024\times1024$ samples with unprecedented photorealism and alignment with text.

Our primary aim with \name is to advance research on generative methods, using text-to-image synthesis as a test bed. While end-user applications of generative methods remain largely out of scope, we recognize the potential downstream applications of this research are varied and may impact society in complex ways. On the one hand, generative models have a great potential to complement, extend, and augment human creativity \cite{hughes2021}. Text-to-image generation models, in particular, have the potential to extend image-editing capabilities  and lead to the development of new tools for creative practitioners. On the other hand, generative methods can be leveraged for malicious purposes, including harassment and misinformation spread \cite{franks2019}, and raise many concerns regarding social and cultural exclusion and bias \cite{Srinivasan2021, Sequeira2021, Steed2021}. 
These considerations inform our decision to not to release code or a public demo. In future work we will explore a framework for responsible externalization that balances the value of external auditing with the risks of unrestricted open-access.

Another ethical challenge relates to the large scale data requirements of text-to-image models, which have have led researchers to rely heavily on large, mostly uncurated, web-scraped datasets. 
While this approach has enabled rapid algorithmic advances in recent years, datasets of this nature have been critiqued and contested along various ethical dimensions. For example, public and academic discourse regarding appropriate use of public data has raised concerns regarding data subject awareness and consent \cite{MegaPixels, dulhantychrisIssuesComputerVision2020, Scheuerman2021DoDH, Paullada2021}. Dataset audits have revealed these datasets tend to reflect social stereotypes, oppressive viewpoints, and derogatory, or otherwise harmful, associations to marginalized identity groups \cite{prabhu2020, birhane2021laionaudit}. 
Training text-to-image models on this data risks reproducing these associations and causing significant representational harm that would disproportionately impact individuals and communities already experiencing marginalization, discrimination and exclusion within society. As such, there are a multitude of data challenges that must be addressed before text-to-image models like \name can be safely integrated into user-facing applications. While we do not directly address these challenges in this work, an awareness of the limitations of our training data guide our decision not to release \name for public use. We strongly caution against the use text-to-image generation methods for any user-facing tools without close care and attention to the contents of the training dataset. 

\name's training data was drawn from several pre-existing datasets of image and English alt-text pairs. 
\ifdefined\neuripssubmission
A subset of this data was filtered to removed noise and undesirable content, such as pornographic imagery and toxic language. 
\else
400 million examples came from FIT400M\footnote{https://github.com/google-research/babeldraw/data_cards/fit400m.md}, a cleaned version of the internal Alt-Text dataset. This data was filtered to removed noise and undesirable content, such as pornographic imagery and toxic language. 
\fi
However, a recent audit of one of our data sources, LAION-400M \cite{schuhmann2021laion}, uncovered a wide range of inappropriate content including pornographic imagery, racist slurs, and harmful social stereotypes \cite{birhane2021laionaudit}. This finding informs our assessment that \name is not suitable for public use at this time and also demonstrates the value of rigorous dataset audits and comprehensive dataset documentation (e.g. \cite{gebruDatasheetsDatasets2020, datacards}) in informing consequent decisions about the model's appropriate and safe use. \name also relies on text encoders trained on uncurated web-scale data, and thus inherits the social biases and limitations of large language models \cite{bordia-naacl-2017, bender2021, weidinger2021}.

 While we leave an in-depth empirical analysis of social and cultural biases encoded by \name to future work, our small scale internal assessments reveal several limitations that guide our decision not to release \name at this time.
 First, all generative models, including \name, 
 \name, may run into danger of dropping modes of the data distribution, which may further compound the social consequence of dataset bias. Second, \name exhibits serious limitations when generating images depicting people. Our human evaluations found \name obtains significantly higher preference rates when evaluated on images that do not portray people, indicating  a degradation in image fidelity. Finally, our preliminary assessment also suggests \name  encodes several social biases and stereotypes, including an overall bias towards generating images of people with lighter skin tones and a tendency for images portraying different professions to align with Western gender stereotypes. Even when we focus generations away from people, our preliminary analysis indicates \name encodes a range of social and cultural biases when generating images of activities, events, and objects. 

While there has been extensive work auditing image-to-text and image labeling models for forms of social bias (e.g. \cite{gendershades, Burns2018, Steed2021}), there has been comparatively less work on social bias evaluation methods for text-to-image models, with the recent exception of \citep{dalleval}. We believe this is a critical avenue for future research and we intend to explore benchmark evaluations for social and cultural bias in future work---for example, exploring whether it is possible to generalize the normalized pointwise mutual information metric \cite{npmi} to the measurement of biases in image generation models. There is also a great need to develop a conceptual vocabulary around potential harms of text-to-image models that could guide the development of evaluation metrics and inform responsible model release.  We aim to address these challenges in future work.

\section{Acknowledgements}

We give thanks to Ben Poole for reviewing our manuscript, early discussions, and providing many helpful comments and suggestions throughout the project. Special thanks to Kathy Meier-Hellstern, Austin Tarango, and Sarah Laszlo for helping us incorporate important responsible AI practices around this project. We appreciate valuable feedback and support from Elizabeth Adkison, Zoubin Ghahramani, Jeff Dean, Yonghui Wu, and Eli Collins. We are grateful to Tom Small for designing the Imagen watermark. We thank Jason Baldridge, Han Zhang, and Kevin Murphy for initial discussions and feedback. We acknowledge hard work and support from Fred Alcober, Hibaq Ali, Marian Croak, Aaron Donsbach, Tulsee Doshi, Toju Duke, Douglas Eck, Jason Freidenfelds, Brian Gabriel, Molly FitzMorris, David Ha, Philip Parham, Laura Pearce, Evan Rapoport, Lauren Skelly, Johnny Soraker, Negar Rostamzadeh, Vijay Vasudevan, Tris Warkentin, Jeremy Weinstein, and Hugh Williams for giving us advice along the project and assisting us with the publication process. We thank Victor Gomes and Erica Moreira for their consistent and critical help with TPU resource allocation. We also give thanks to Shekoofeh Azizi, Harris Chan, Chris A. Lee, and Nick Ma for volunteering a considerable amount of their time for testing out DrawBench. We thank Aditya Ramesh, Prafulla Dhariwal, and Alex Nichol for allowing us to use DALL-E~2 samples and providing us with GLIDE samples. We are thankful to Matthew Johnson and Roy Frostig for starting the JAX project and to the whole JAX team for building such a fantastic system for high-performance machine learning research. Special thanks to Durk Kingma, Jascha Sohl-Dickstein, Lucas Theis and the Toronto Brain team for helpful discussions and spending time Imagening!

\bibliographystyle{plainnat}
\bibliography{main}

\FloatBarrier
\newpage
\appendix
\renewcommand\thefigure{A.\arabic{figure}} \setcounter{figure}{0}
\renewcommand\thetable{A.\arabic{table}} \setcounter{table}{0}

\begin{figure}[p]
\vspace*{-2cm}

\newgeometry{left=0cm,top=0cm,right=0cm,bottom=0cm}%
\setlength{\tabcolsep}{2.0pt}
\captionsetup[subfigure]{labelformat=empty}
\hspace*{-7.8cm}
\setlength{\tabcolsep}{2.0pt}
\centering
\begin{tabular}{ccc}
\begin{subfigure}[t]{0.31\textwidth} \centering \includegraphics[width=\textwidth]{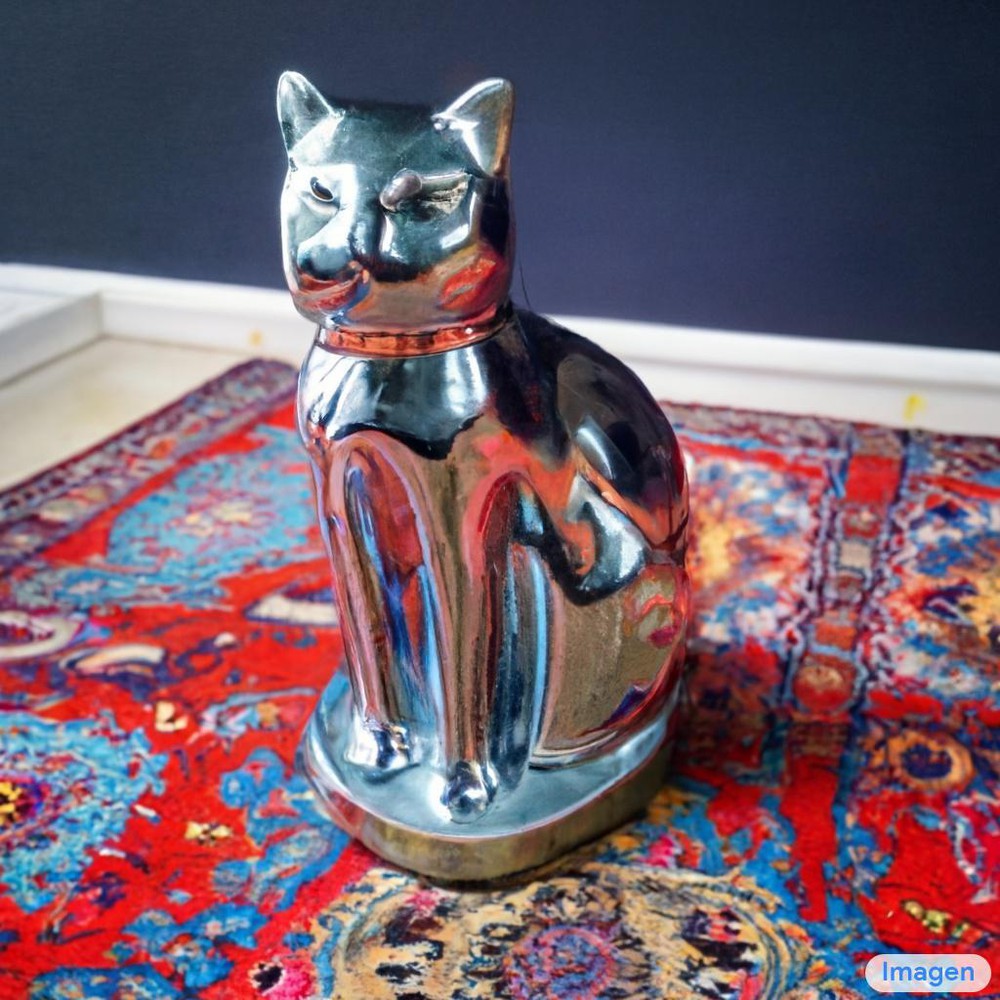} \caption{A chromeplated cat sculpture placed on a Persian rug.} \end{subfigure} &
\begin{subfigure}[t]{0.31\textwidth} \centering \includegraphics[width=\textwidth]{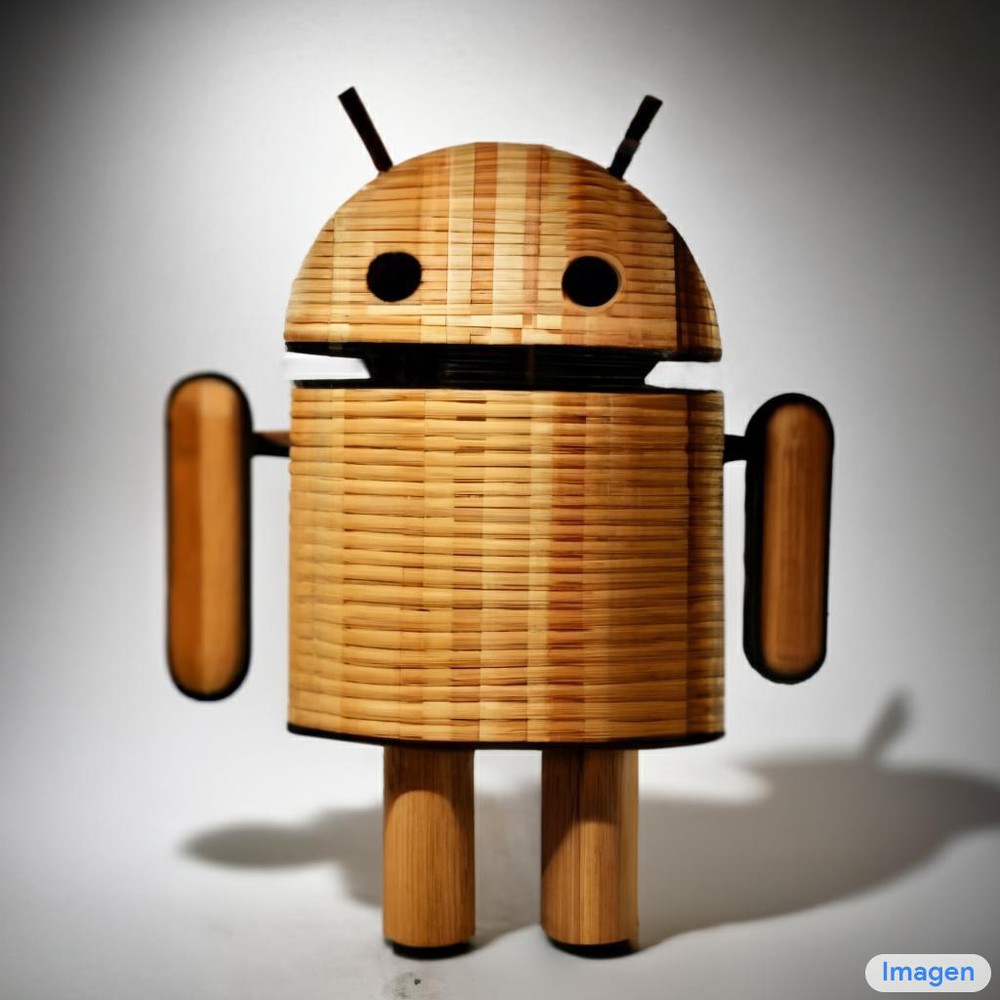} \caption{Android Mascot made from bamboo.} \end{subfigure} &
\begin{subfigure}[t]{0.31\textwidth} \centering \includegraphics[width=\textwidth]{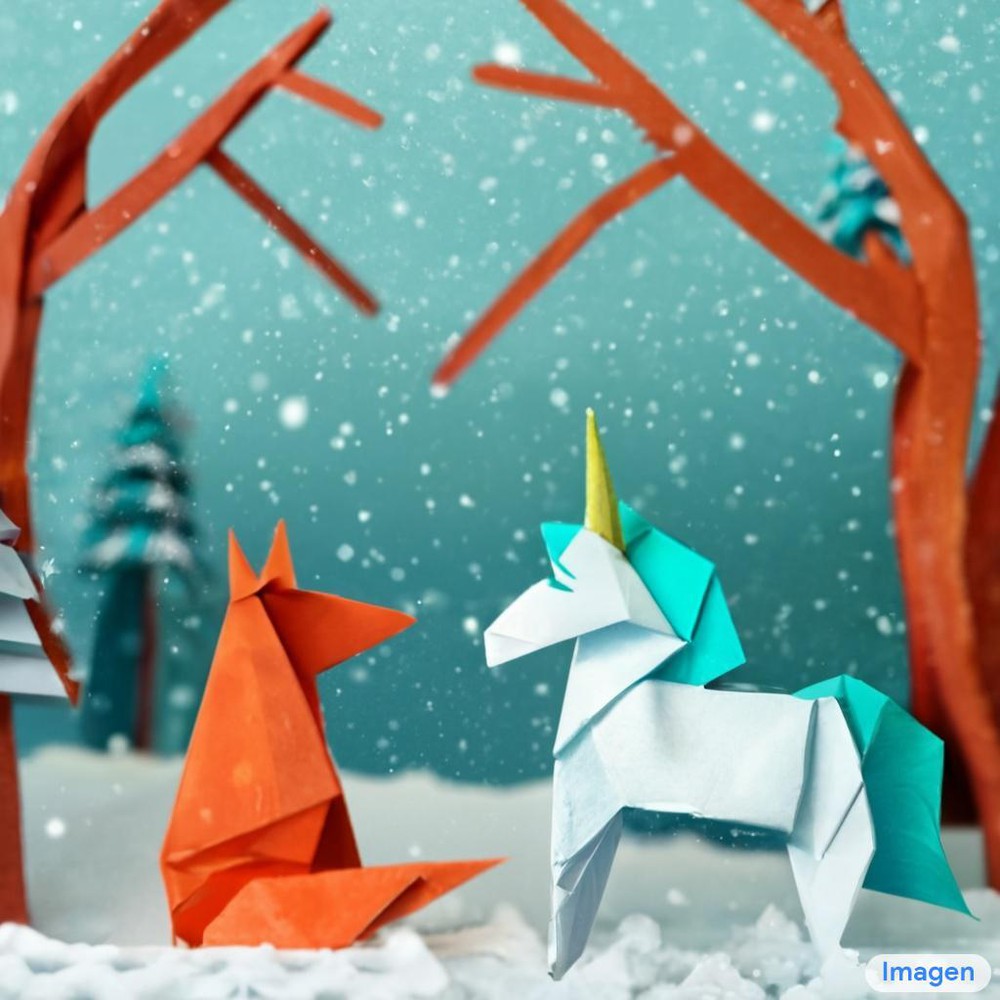} \caption{Intricate origami of a fox and a unicorn in a snowy forest.} \end{subfigure} \\
\addlinespace
\begin{subfigure}[t]{0.31\textwidth} \centering \includegraphics[width=\textwidth]{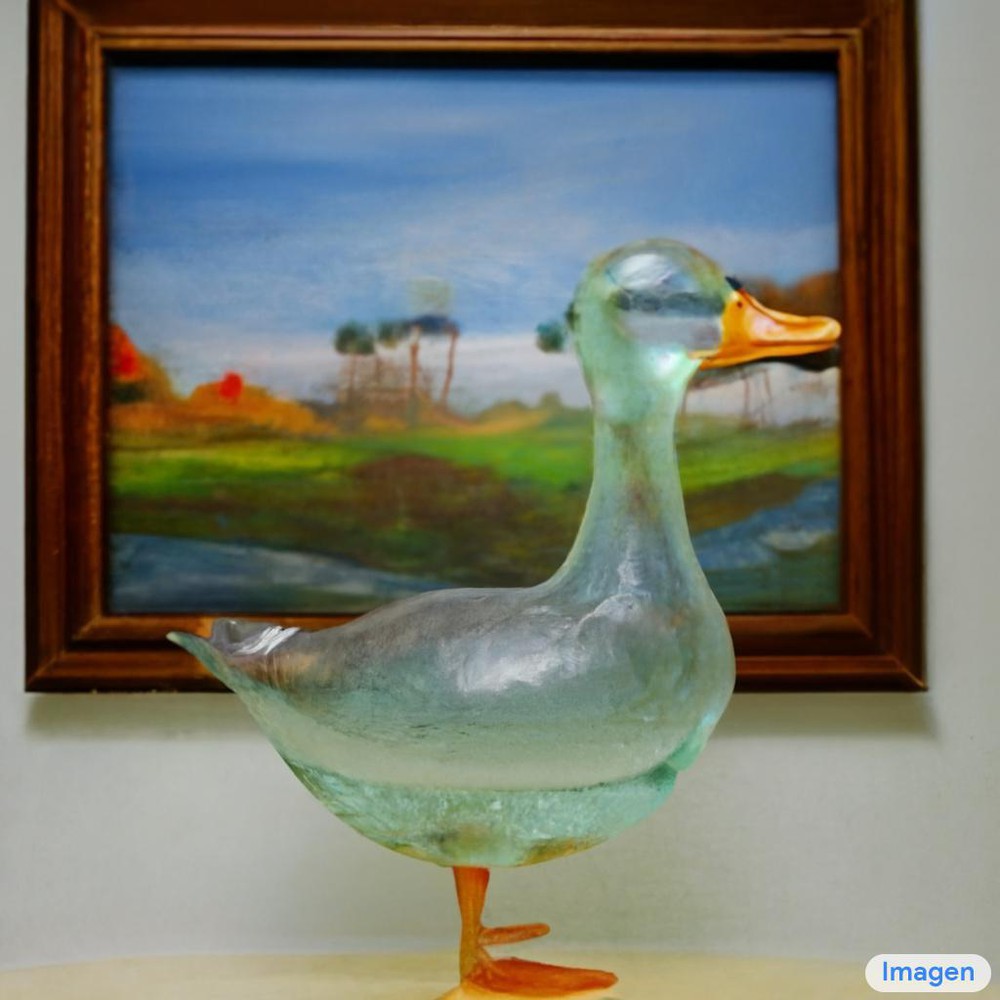} \caption{A transparent sculpture of a duck made out of glass.} \end{subfigure} &
\begin{subfigure}[t]{0.31\textwidth} \centering \includegraphics[width=\textwidth]{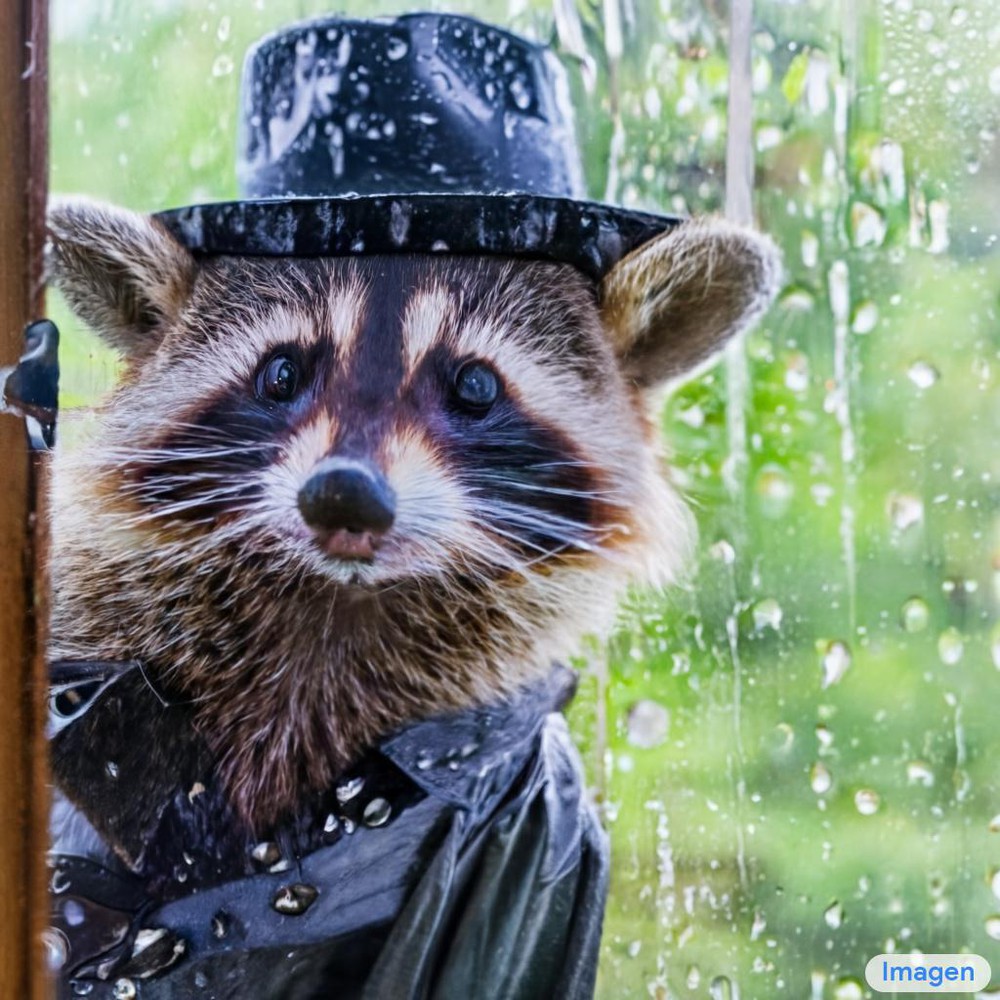} \caption{A raccoon wearing cowboy hat and black leather jacket is behind the backyard window. Rain droplets on the window.} \end{subfigure} &
\begin{subfigure}[t]{0.31\textwidth} \centering \includegraphics[width=\textwidth]{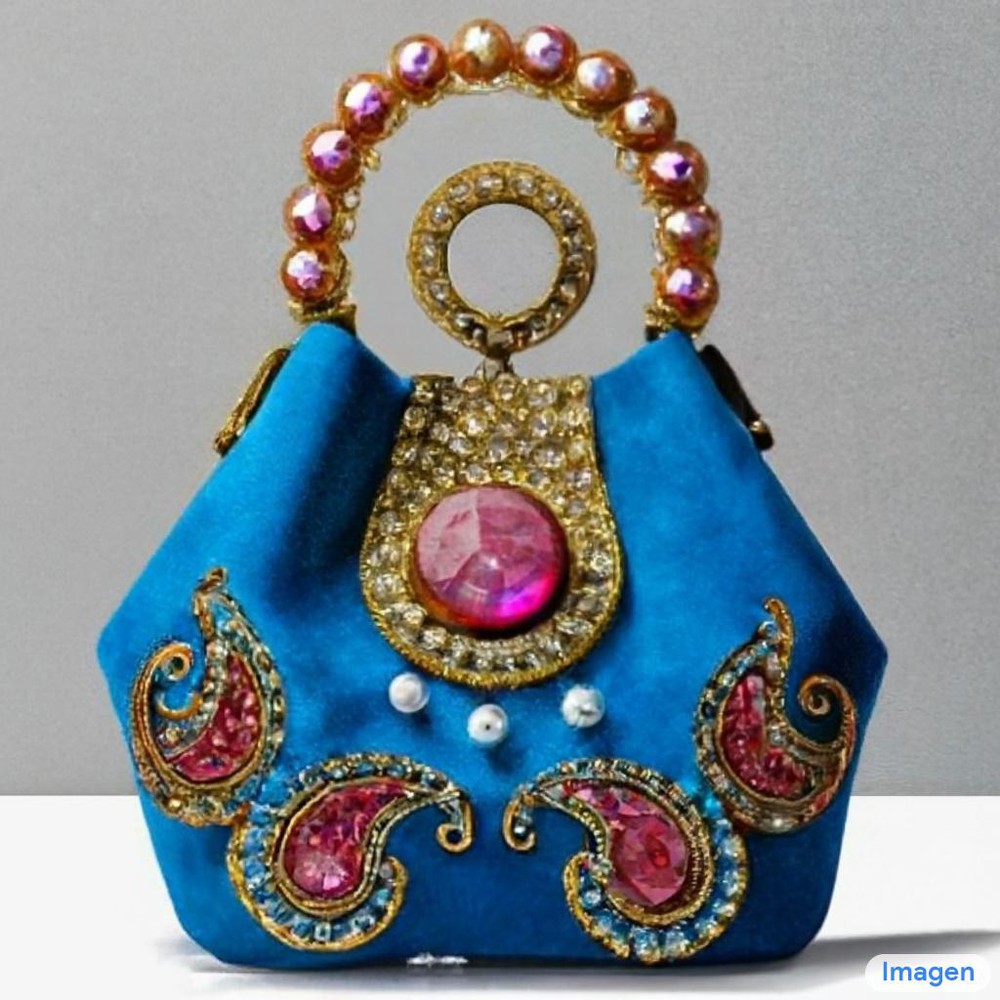} \caption{A bucket bag made of blue suede. The bag is decorated with intricate golden paisley patterns. The handle of the bag is made of rubies and pearls.} \end{subfigure} \\
\addlinespace
\begin{subfigure}[t]{0.31\textwidth} \centering \includegraphics[width=\textwidth]{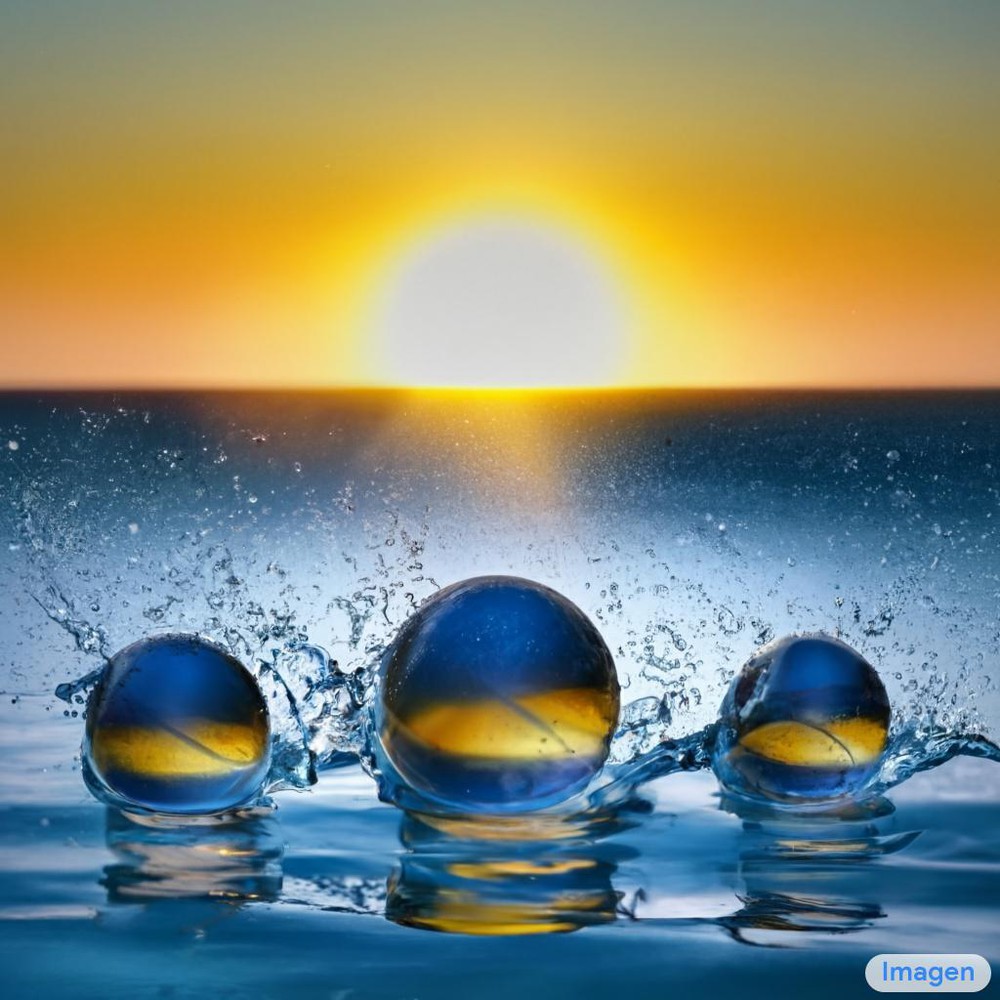} \caption{Three spheres made of glass falling into ocean. Water is splashing. Sun is setting.} \end{subfigure} &
\begin{subfigure}[t]{0.31\textwidth} \centering \includegraphics[width=\textwidth]{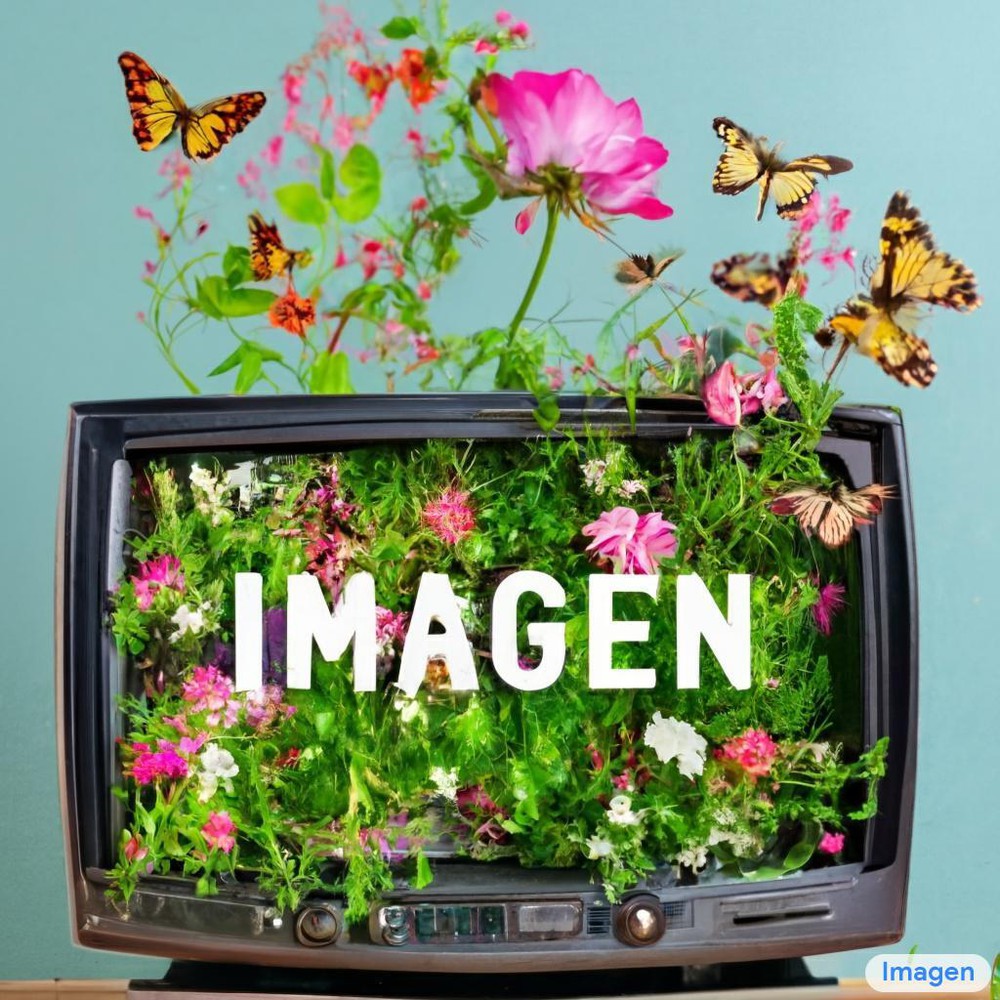} \caption{Vines in the shape of text 'Imagen' with flowers and butterflies bursting out of an old TV.} \end{subfigure} &
\begin{subfigure}[t]{0.31\textwidth} \centering \includegraphics[width=\textwidth]{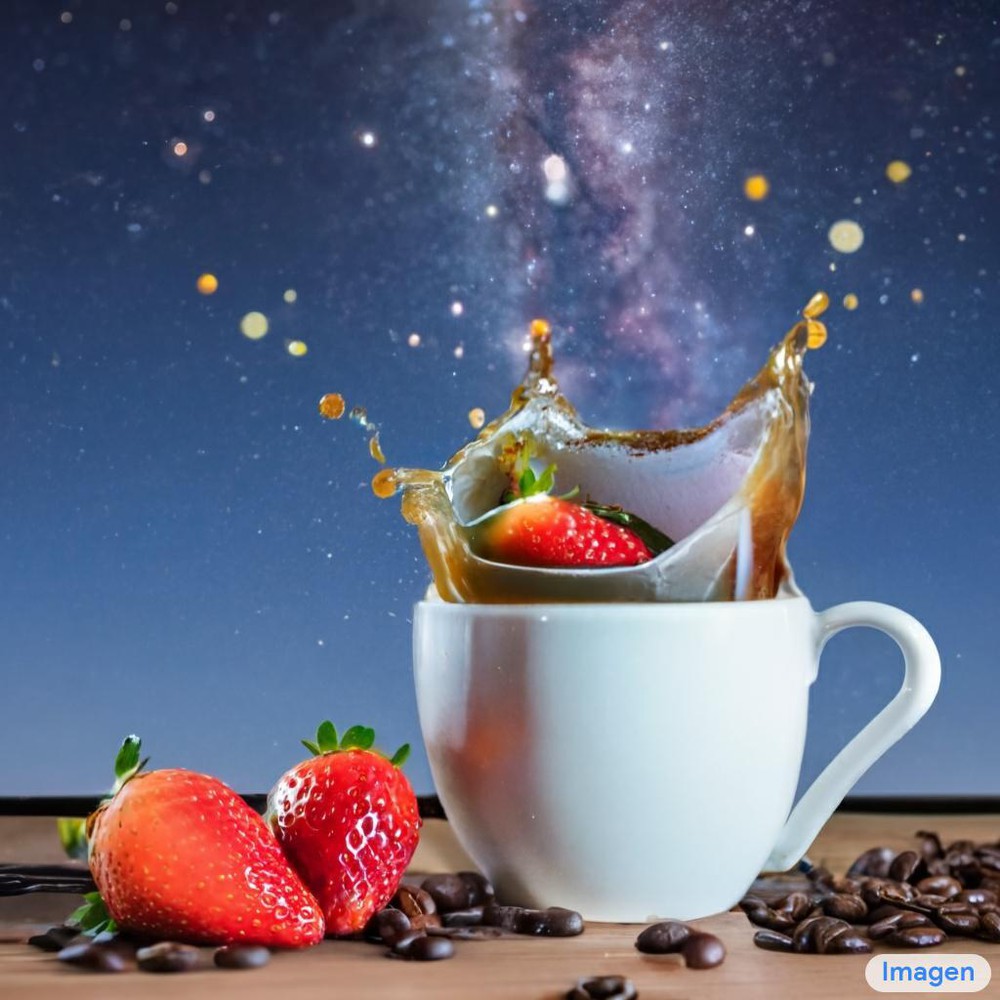} \caption{A strawberry splashing in the coffee in a mug under the starry sky.} \end{subfigure}
\end{tabular}
\newgeometry{
textheight=9in,
textwidth=5.5in,
top=1in,
headheight=12pt,
headsep=25pt,
footskip=30pt
}%
\caption{Select $1024\times1024$ \name samples for various text inputs.}\label{fig:full_page_collage_app1}
\end{figure}%

\begin{figure}[p]
\vspace*{-2cm}

\newgeometry{left=0cm,top=0cm,right=0cm,bottom=0cm}%
\setlength{\tabcolsep}{2.0pt}
\captionsetup[subfigure]{labelformat=empty}
\hspace*{-7.8cm}
\setlength{\tabcolsep}{2.0pt}
\centering
\begin{tabular}{ccc}
\begin{subfigure}[t]{0.31\textwidth} \centering \includegraphics[width=\textwidth]{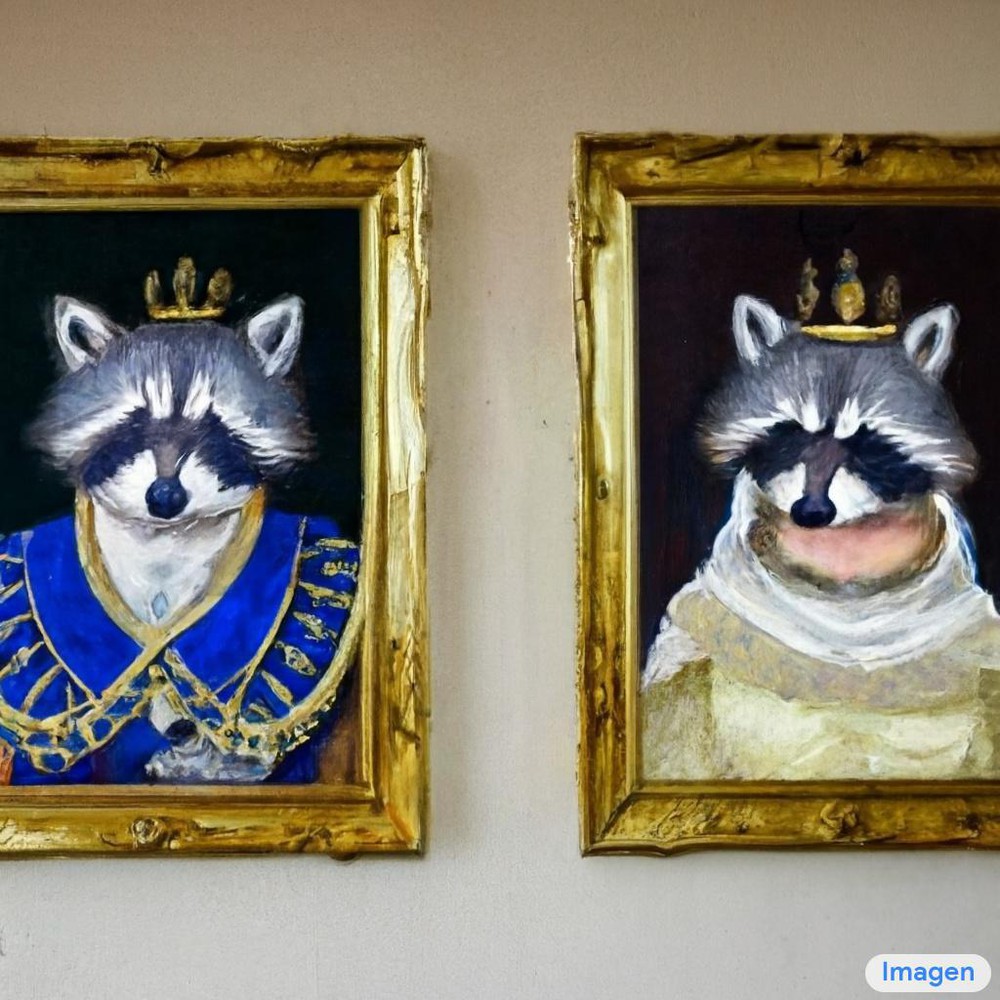} \caption{A wall in a royal castle. There are two paintings on the wall. The one on the left a detailed oil painting of the royal raccoon king. The one on the right a detailed oil painting of the royal raccoon queen.} \end{subfigure} &
\begin{subfigure}[t]{0.31\textwidth} \centering \includegraphics[width=\textwidth]{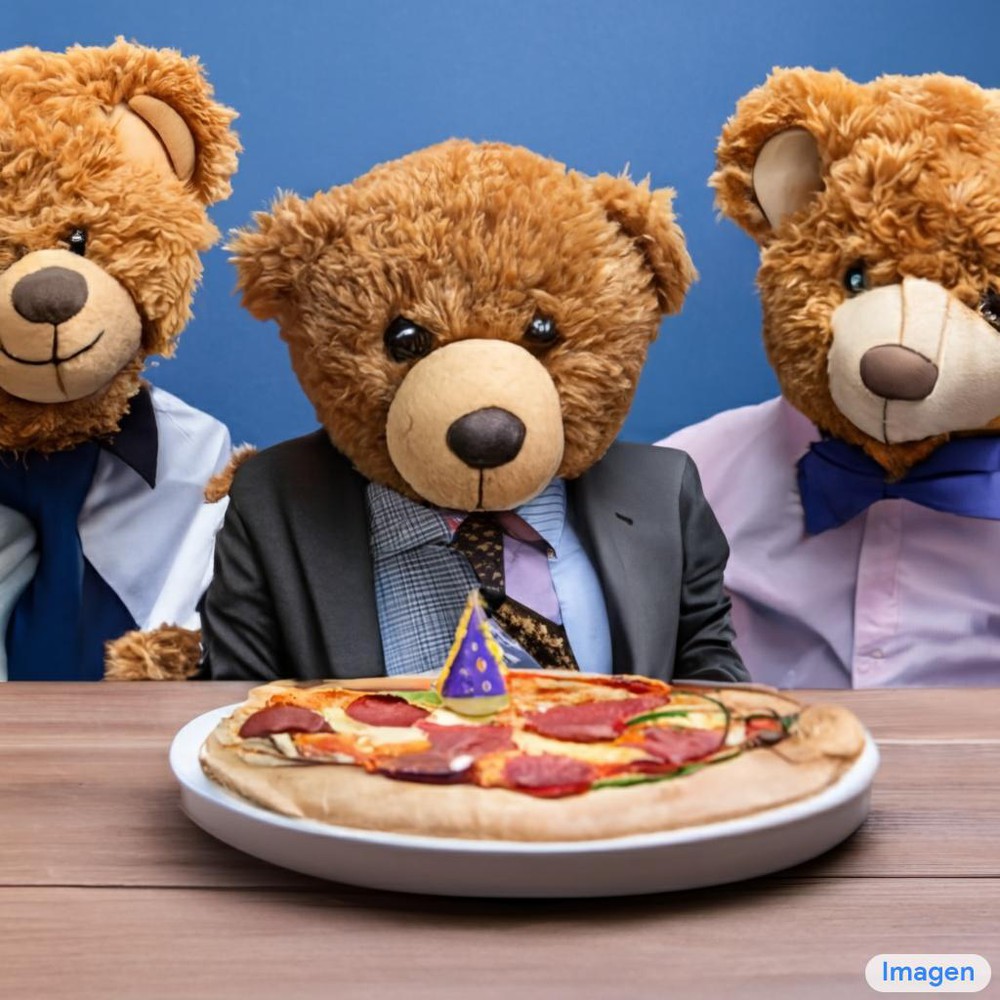} \caption{A group of teddy bears in suit in a corporate office celebrating the birthday of their friend. There is a pizza cake on the desk.} \end{subfigure} &
\begin{subfigure}[t]{0.31\textwidth} \centering \includegraphics[width=\textwidth]{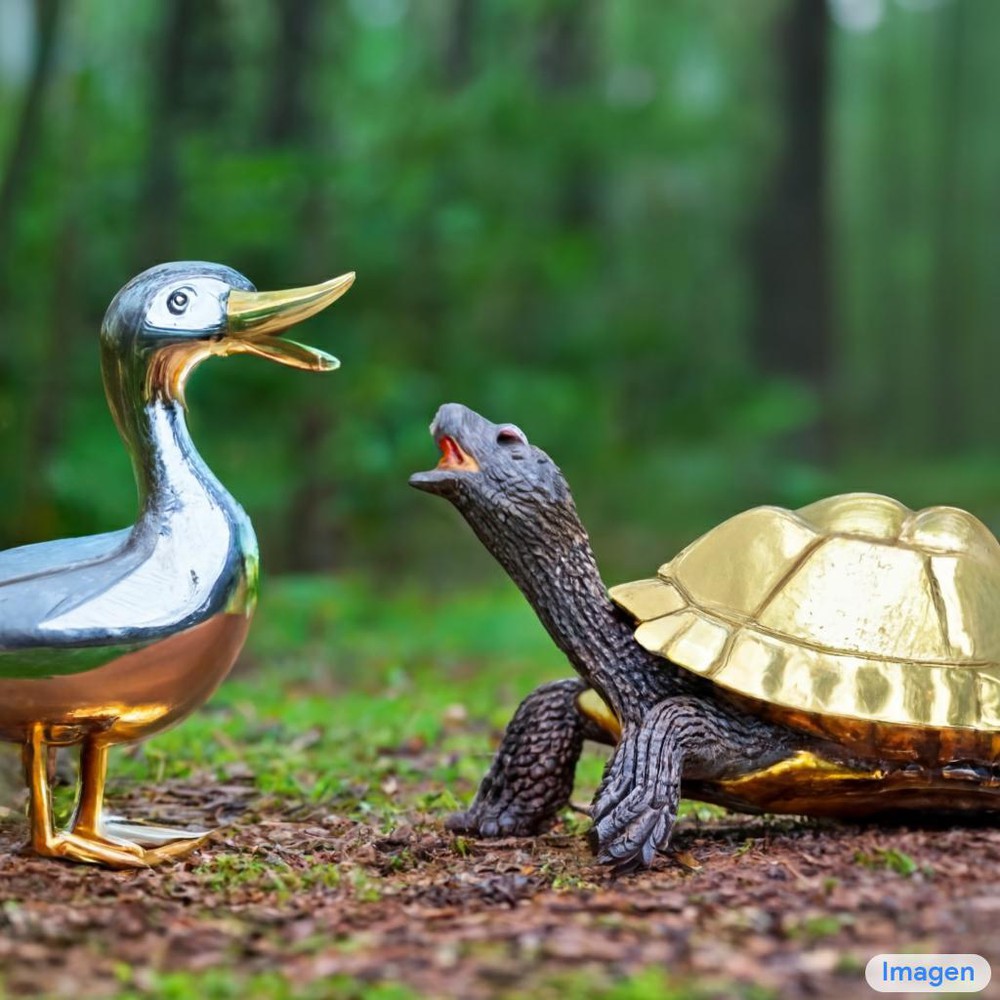} \caption{A chrome-plated duck with a golden beak arguing with an angry turtle in a forest.} \end{subfigure} \\
\addlinespace
\begin{subfigure}[t]{0.31\textwidth} \centering \includegraphics[width=\textwidth]{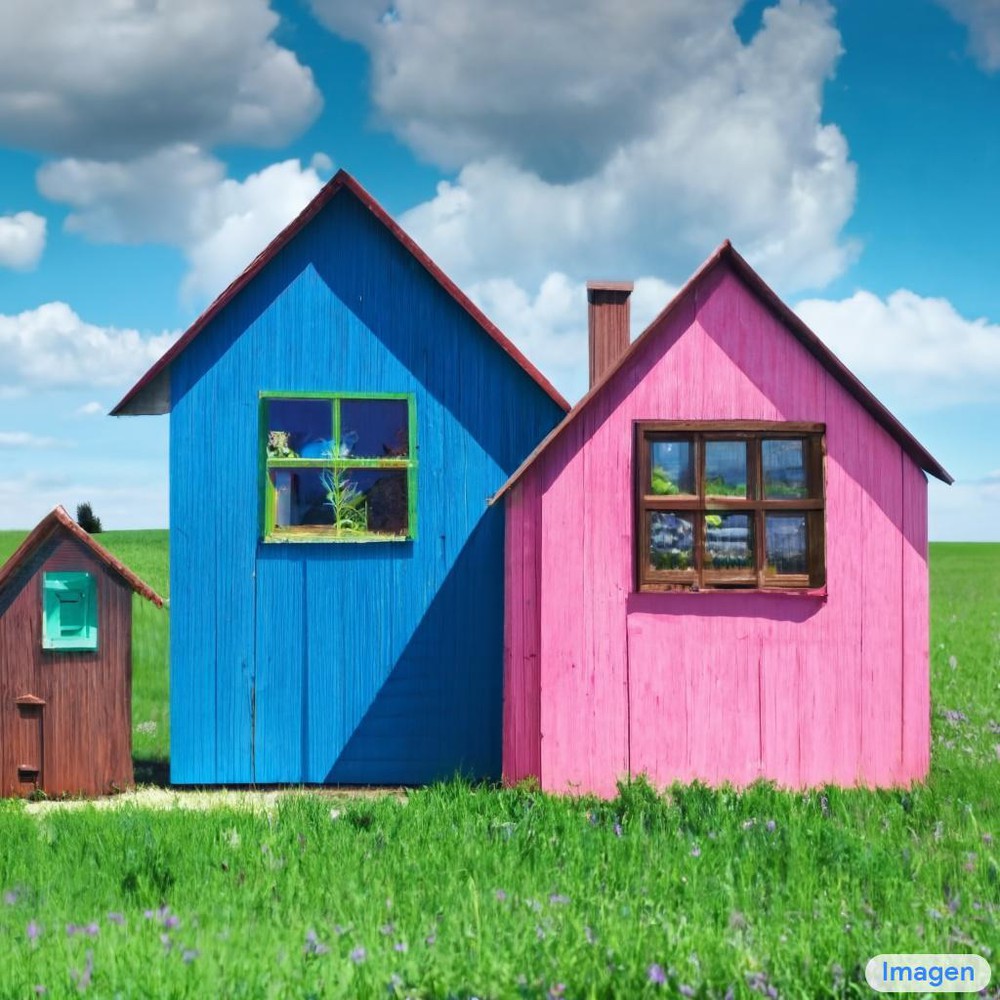} \caption{A family of three houses in a meadow. The Dad house is a large blue house. The Mom house is a large pink house. The Child house is a small wooden shed.} \end{subfigure} &
\begin{subfigure}[t]{0.31\textwidth} \centering \includegraphics[width=\textwidth]{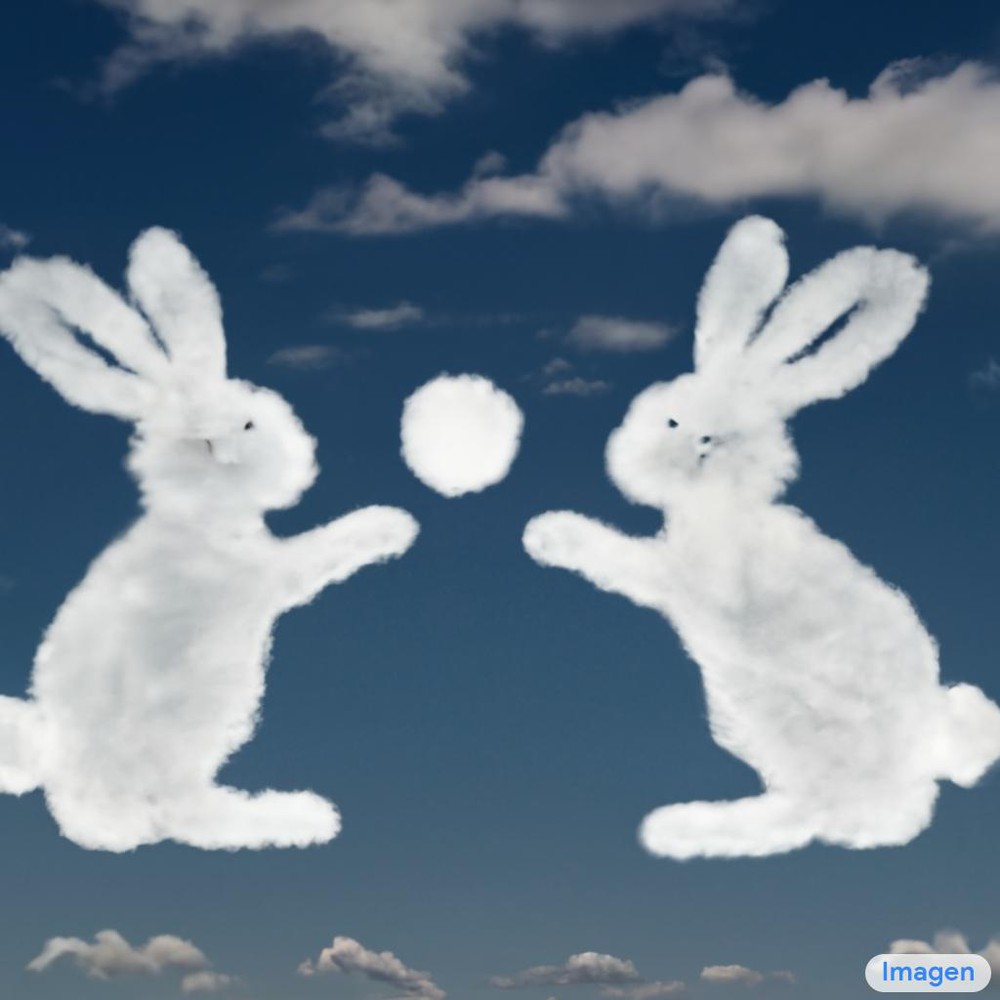} \caption{A cloud in the shape of two bunnies playing with a ball. The ball is made of clouds too.} \end{subfigure} &
\begin{subfigure}[t]{0.31\textwidth} \centering \includegraphics[width=\textwidth]{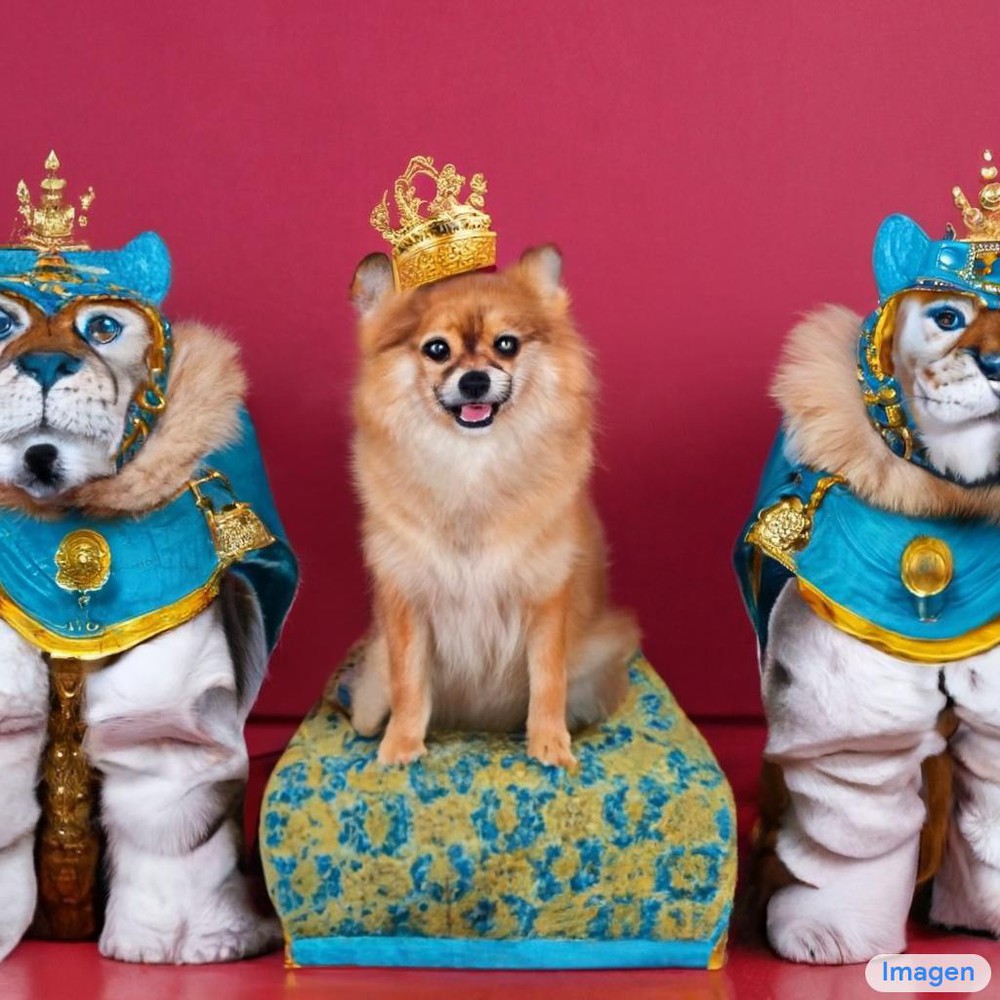} \caption{A Pomeranian is sitting on the Kings throne wearing a crown. Two tiger soldiers are standing next to the throne.} \end{subfigure} \\
\addlinespace
\begin{subfigure}[t]{0.31\textwidth} \centering \includegraphics[width=\textwidth]{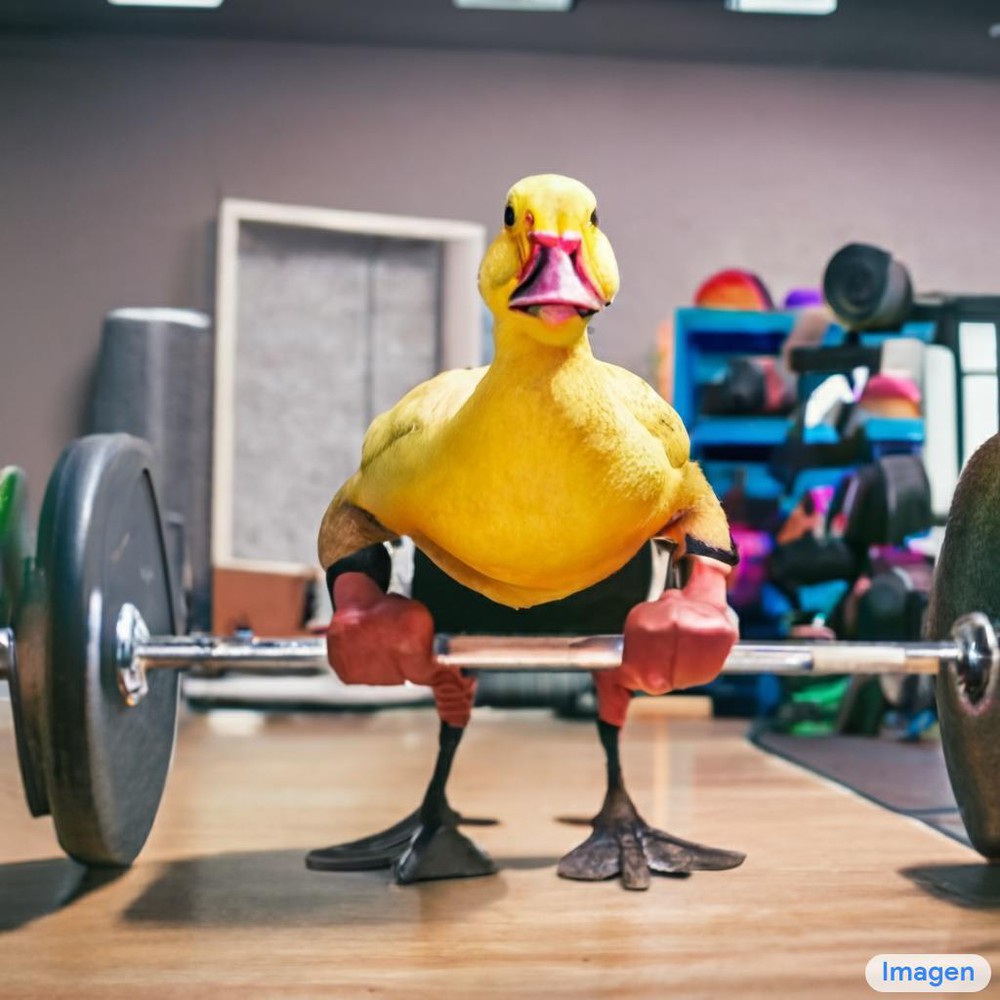} \caption{An angry duck doing heavy weightlifting at the gym.} \end{subfigure} &
\begin{subfigure}[t]{0.31\textwidth} \centering \includegraphics[width=\textwidth]{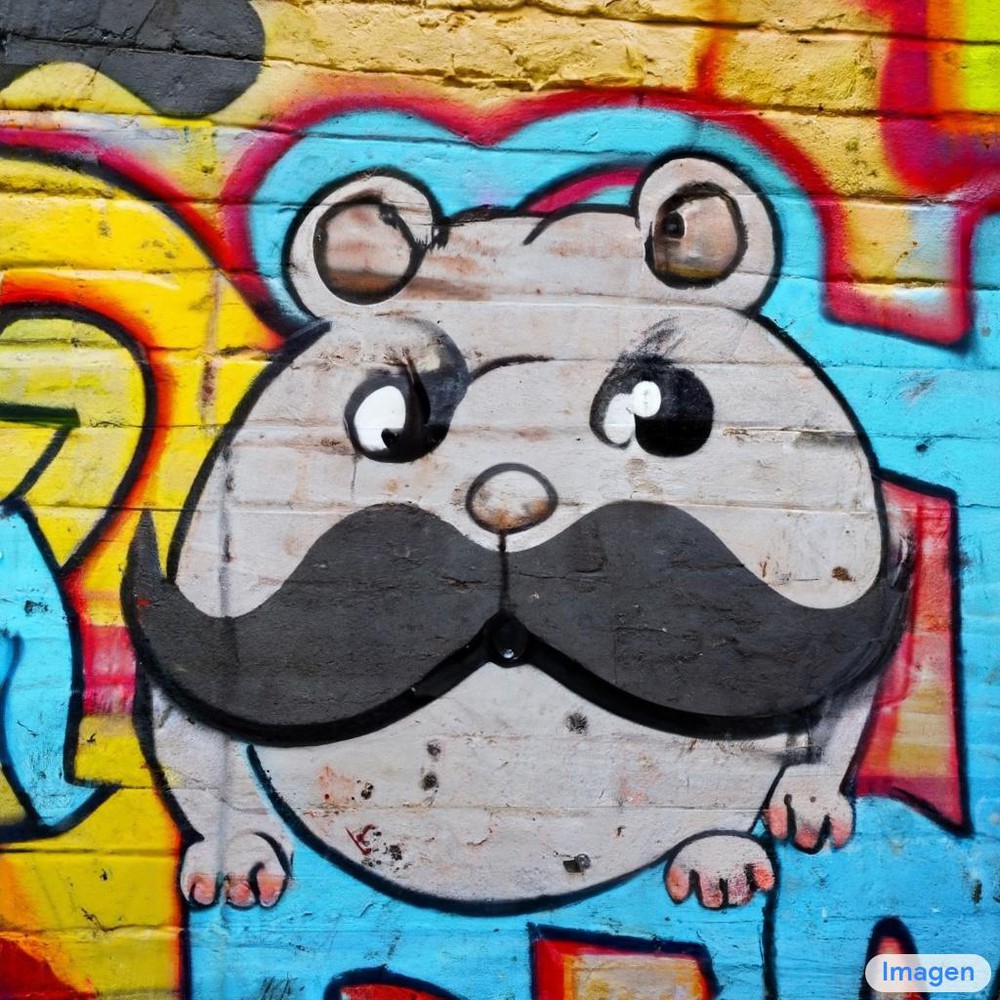} \caption{A dslr picture of colorful graffiti showing a hamster with a moustache.} \end{subfigure} &
\begin{subfigure}[t]{0.31\textwidth} \centering \includegraphics[width=\textwidth]{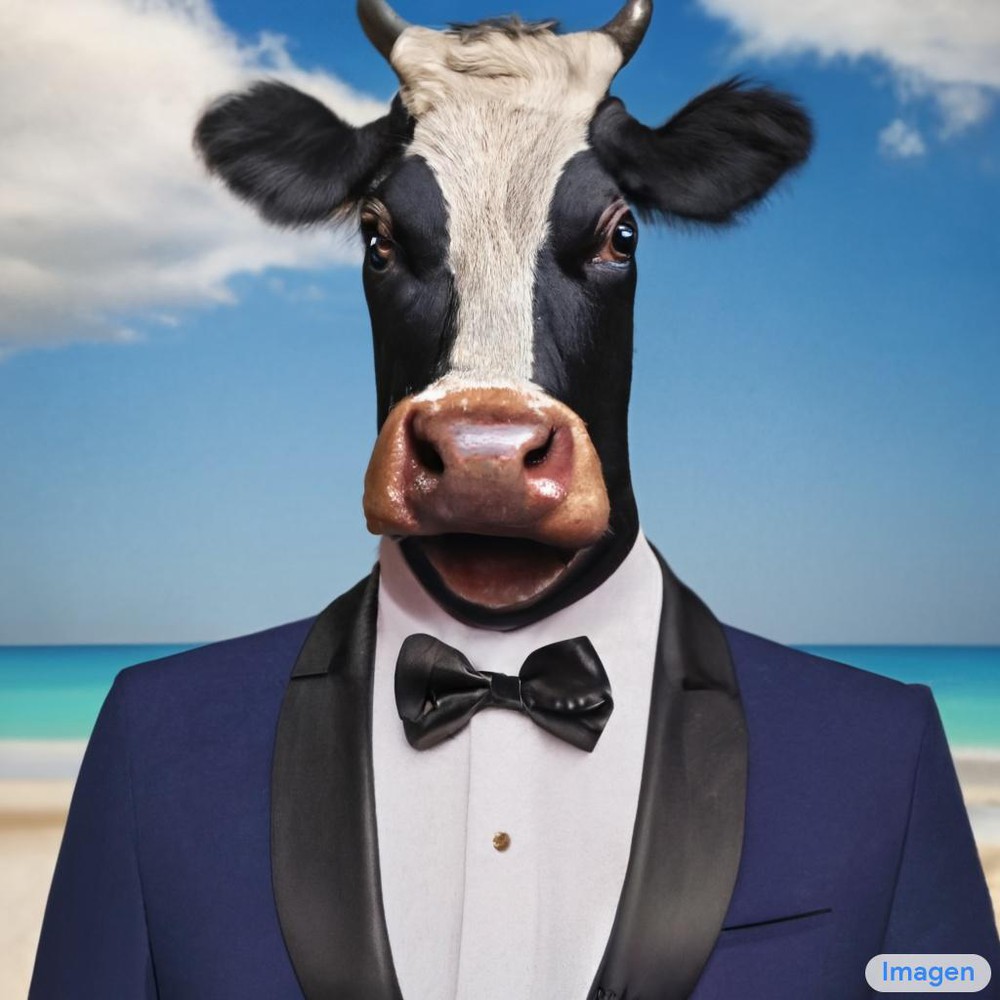} \caption{A photo of a person with the head of a cow, wearing a tuxedo and black bowtie. Beach wallpaper in the background.} \end{subfigure}
\end{tabular}
\newgeometry{
textheight=9in,
textwidth=5.5in,
top=1in,
headheight=12pt,
headsep=25pt,
footskip=30pt
}%
\caption{Select $1024\times1024$ \name samples for various text inputs.}\label{fig:full_page_collage_app2}
\end{figure}%

\begin{figure}[p]
\vspace*{-2cm}

\newgeometry{left=0cm,top=0cm,right=0cm,bottom=0cm}%
\setlength{\tabcolsep}{2.0pt}
\captionsetup[subfigure]{labelformat=empty}
\hspace*{-7.8cm}
\setlength{\tabcolsep}{2.0pt}
\centering
\begin{tabular}{ccc}
\begin{subfigure}[t]{0.31\textwidth} \centering \includegraphics[width=\textwidth]{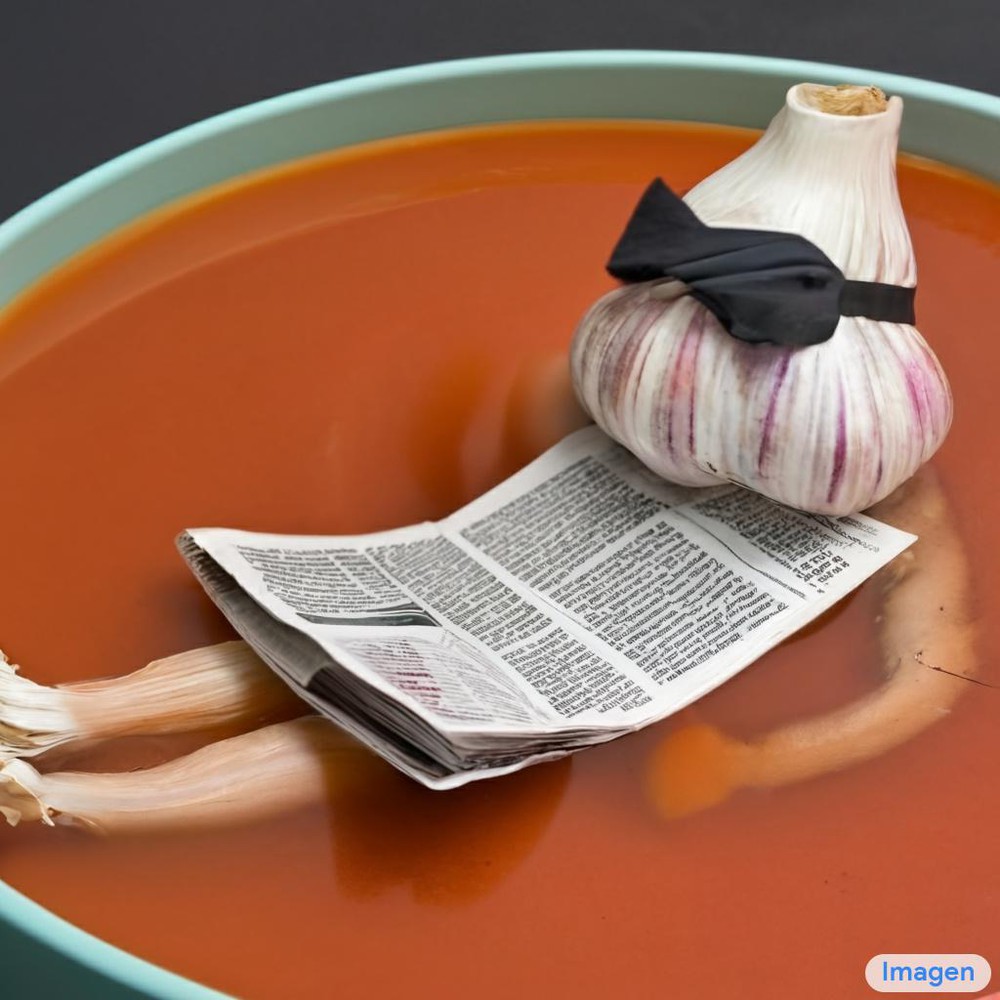} \caption{A relaxed garlic with a blindfold reading a newspaper while floating in a pool of tomato soup.} \end{subfigure} &
\begin{subfigure}[t]{0.31\textwidth} \centering \includegraphics[width=\textwidth]{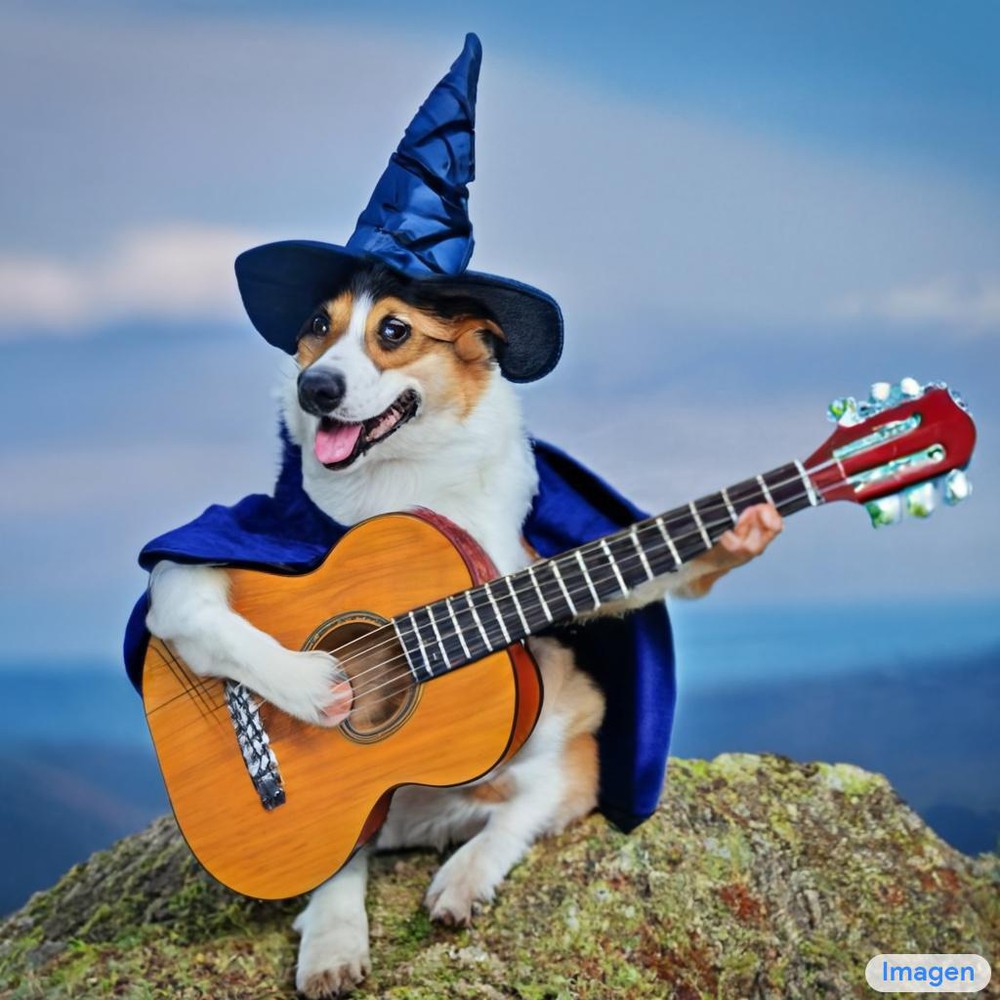} \caption{A photo of a corgi dog wearing a wizard hat playing guitar on the top of a mountain.} \end{subfigure} &
\begin{subfigure}[t]{0.31\textwidth} \centering \includegraphics[width=\textwidth]{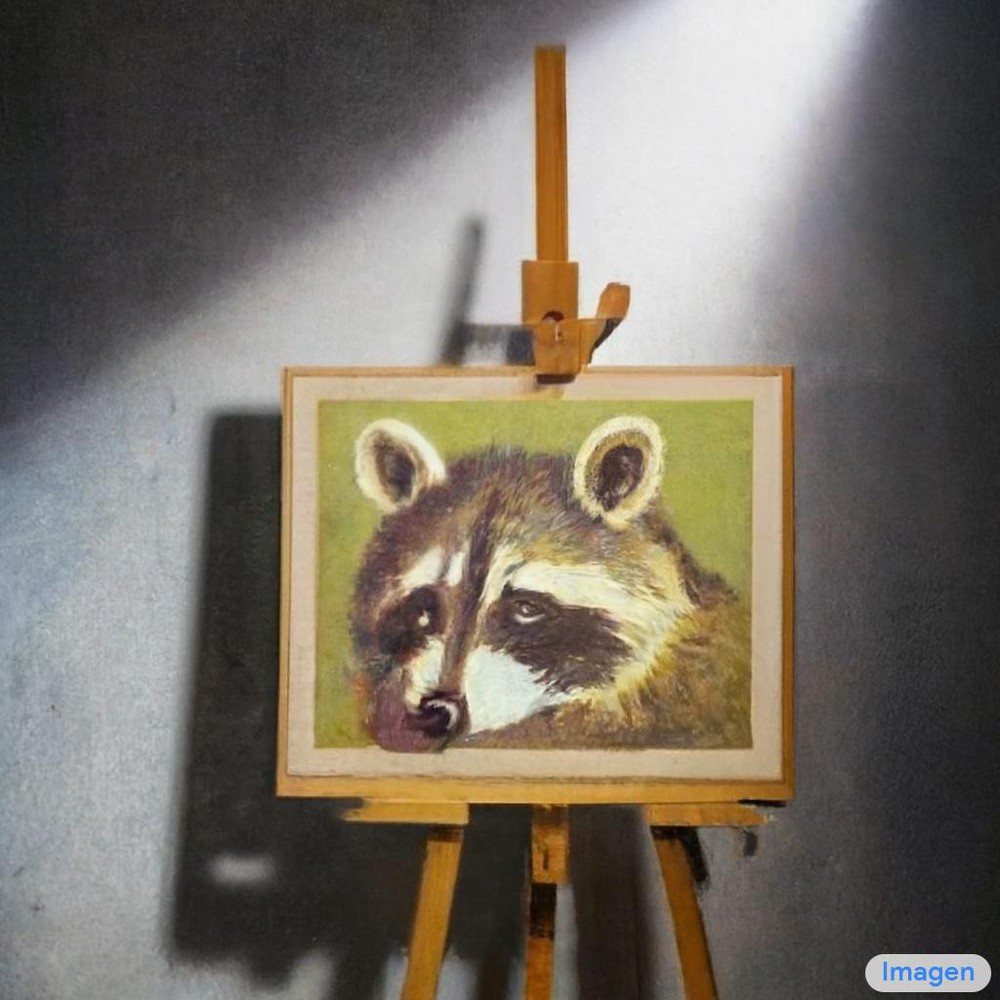} \caption{A single beam of light enter the room from the ceiling. The beam of light is illuminating an easel. On the easel there is a Rembrandt painting of a raccoon.} \end{subfigure} \\
\addlinespace
\begin{subfigure}[t]{0.31\textwidth} \centering \includegraphics[width=\textwidth]{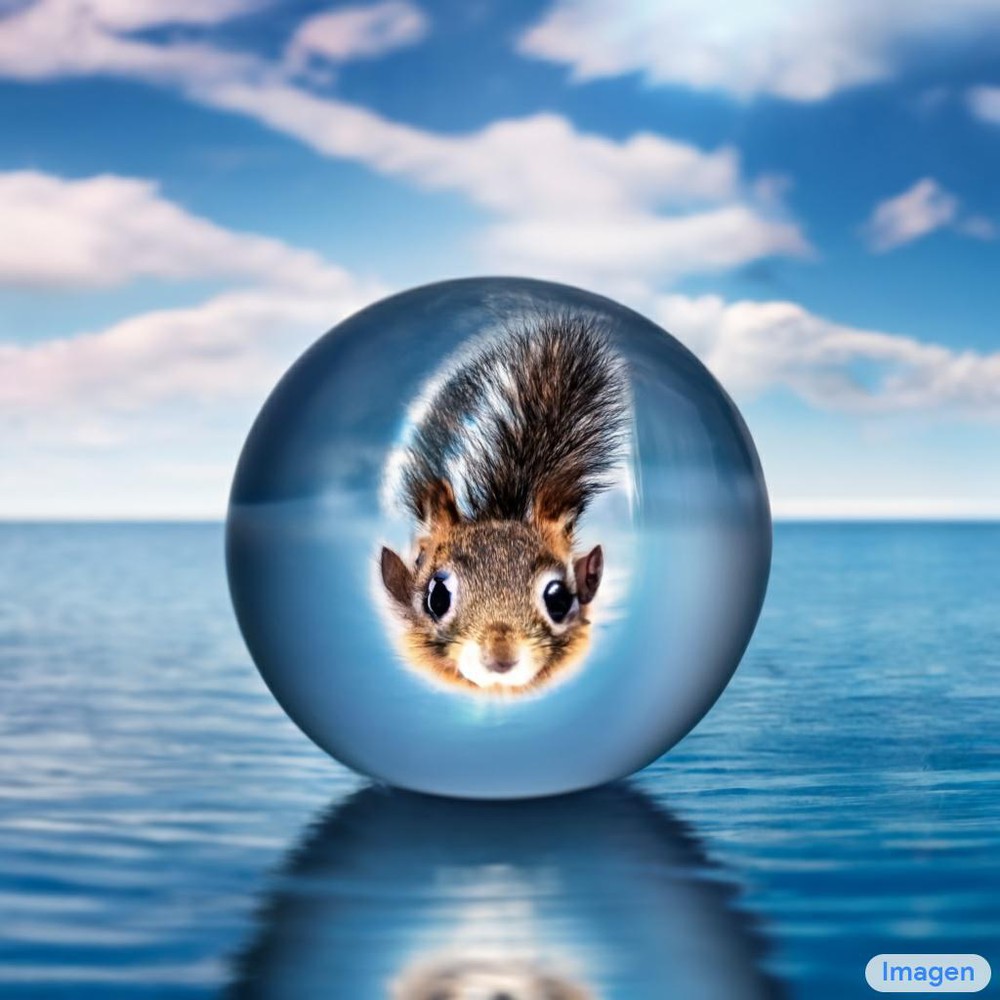} \caption{A squirrel is inside a giant bright shiny crystal ball in on the surface of blue ocean. There are few clouds in the sky.} \end{subfigure}  &
\begin{subfigure}[t]{0.31\textwidth} \centering \includegraphics[width=\textwidth]{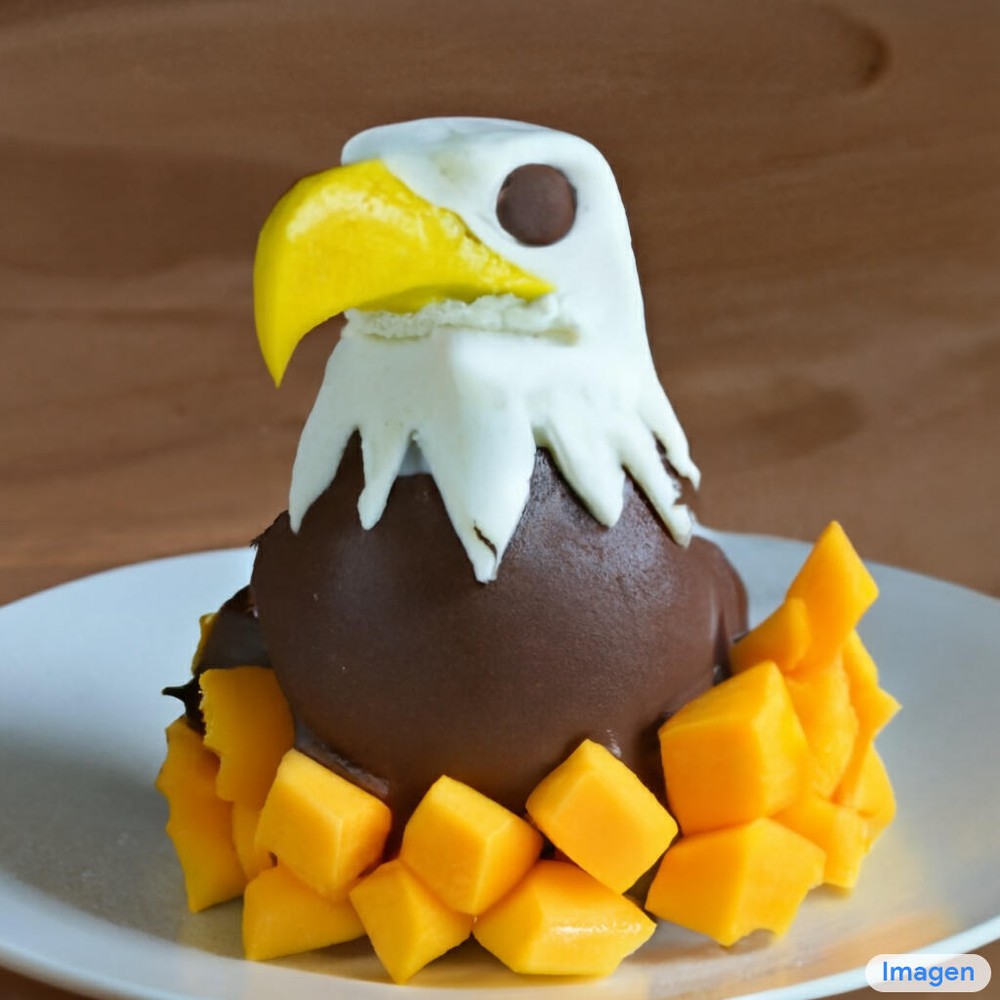} \caption{A bald eagle made of chocolate powder, mango, and whipped cream.} \end{subfigure} &
\begin{subfigure}[t]{0.31\textwidth} \centering \includegraphics[width=\textwidth]{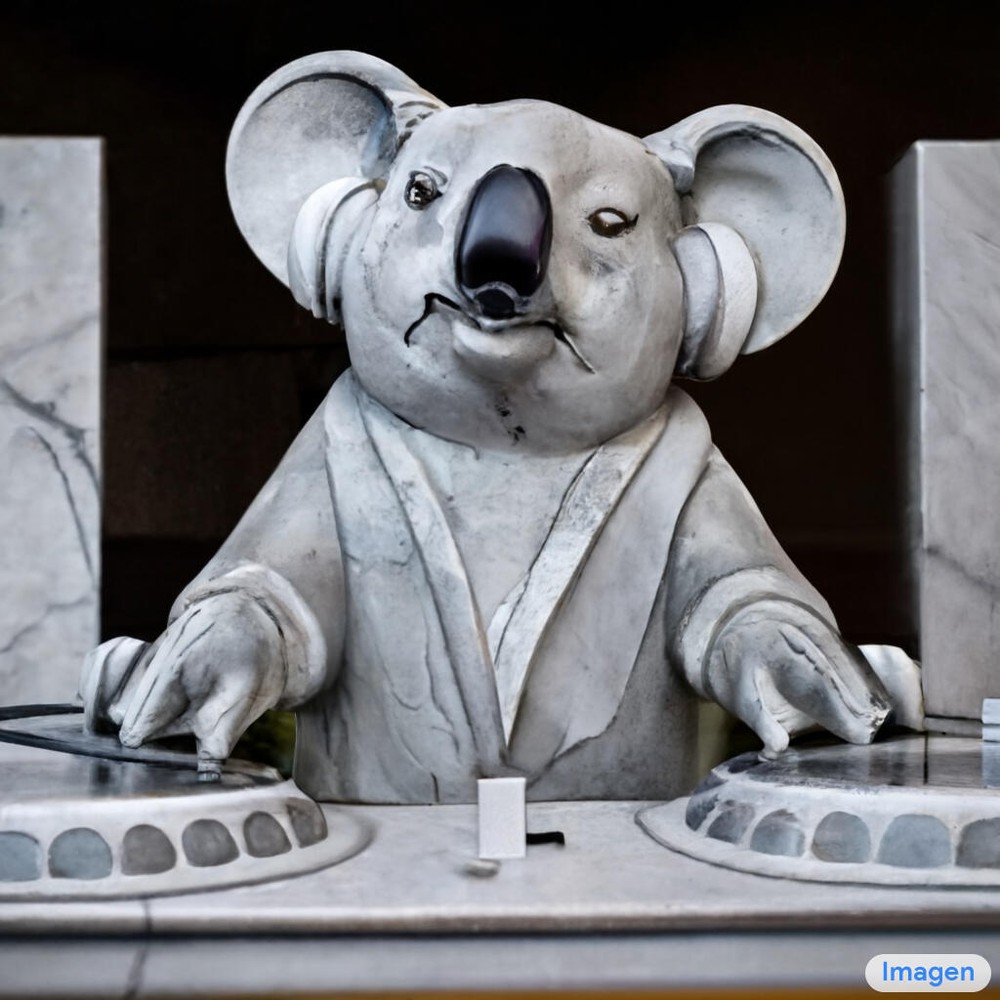} \caption{A marble statue of a Koala DJ in front of a marble statue of a turntable. The Koala has wearing large marble headphones.} \end{subfigure} \\
\addlinespace
\begin{subfigure}[t]{0.31\textwidth} \centering \includegraphics[width=\textwidth]{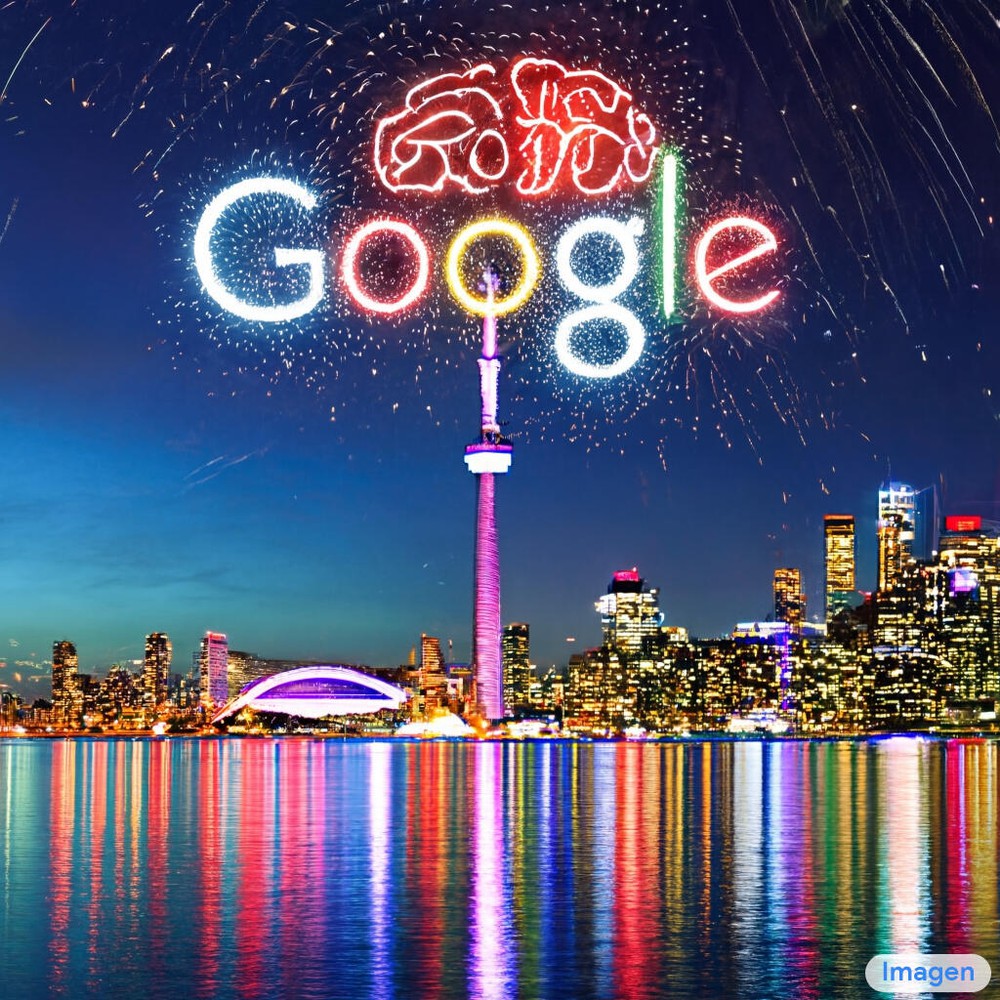} \caption{The Toronto skyline with Google brain logo written in fireworks.} \end{subfigure} &
\begin{subfigure}[t]{0.31\textwidth} \centering \includegraphics[width=\textwidth]{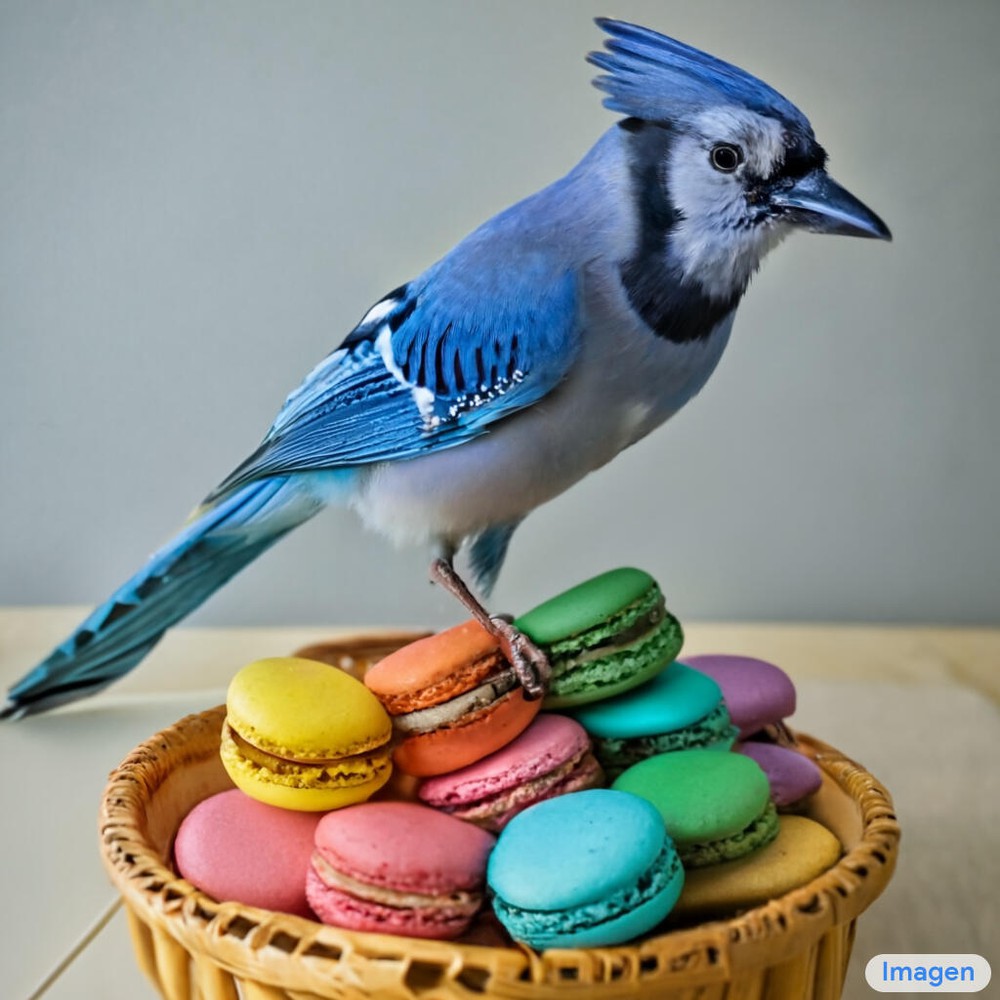} \caption{A blue jay standing on a large basket of rainbow macarons.} \end{subfigure} &
\begin{subfigure}[t]{0.31\textwidth} \centering \includegraphics[width=\textwidth]{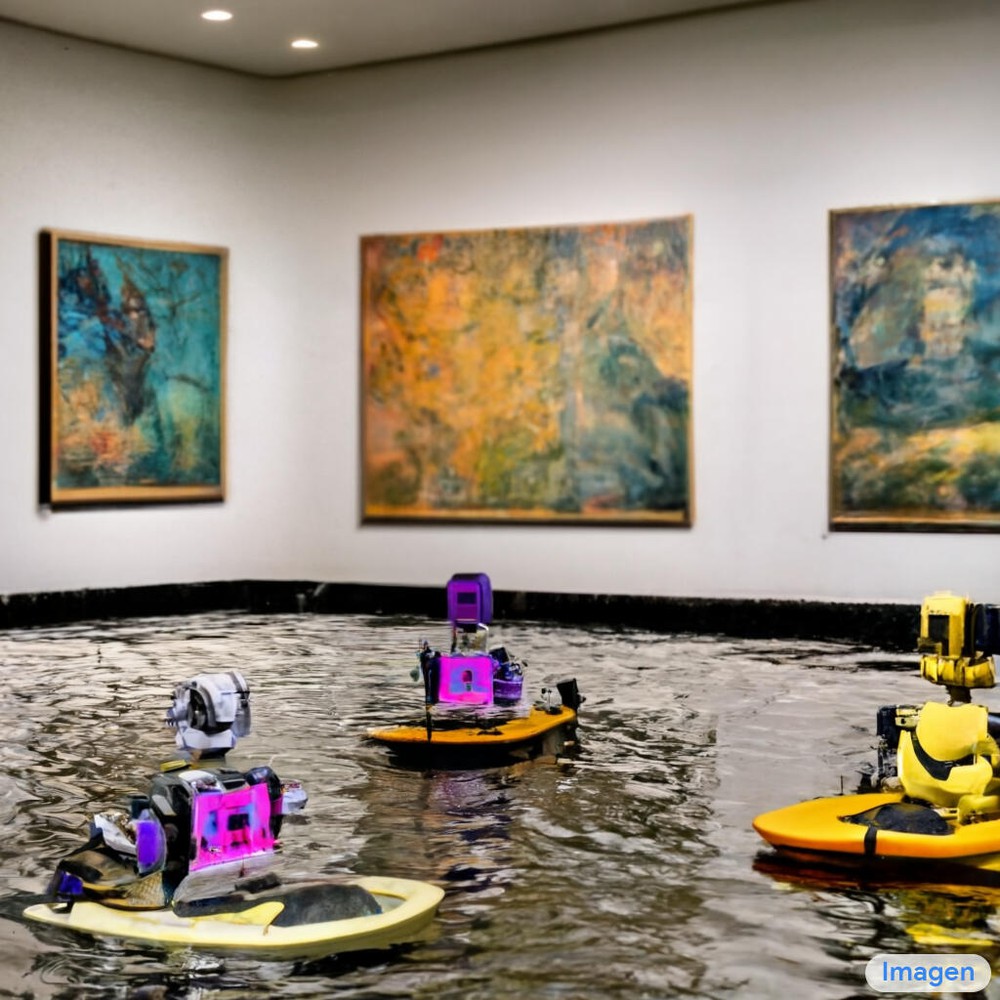} \caption{An art gallery displaying Monet paintings. The art gallery is flooded. Robots are going around the art gallery using paddle boards.} \end{subfigure} \\
\end{tabular}
\newgeometry{
textheight=9in,
textwidth=5.5in,
top=1in,
headheight=12pt,
headsep=25pt,
footskip=30pt
}%
\caption{Select $1024\times1024$ \name samples for various text inputs.}\label{fig:full_page_collage_app3}
\end{figure}%

\begin{figure}[t]
    \centering
    \input{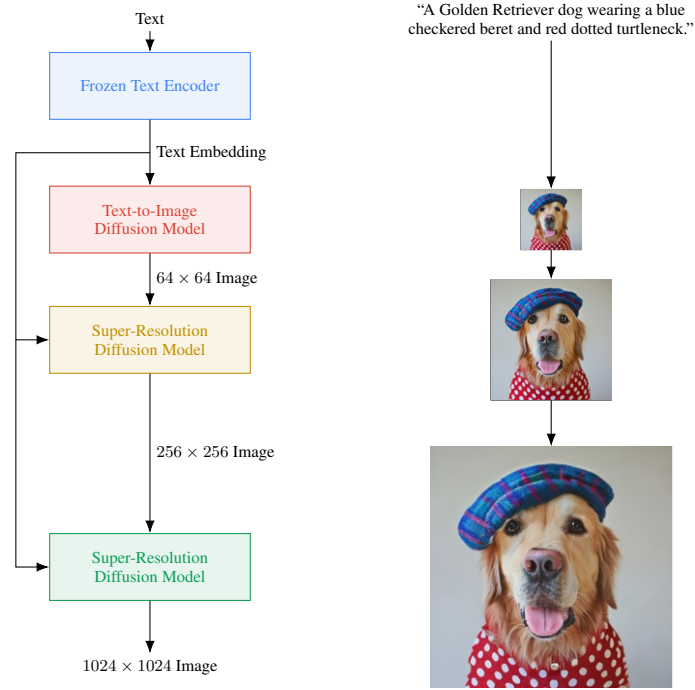}
    \caption{Visualization of \name. \name uses a frozen text encoder to encode the input text into text embeddings. A conditional diffusion model maps the text embedding into a $64\times64$ image. \name further utilizes text-conditional super-resolution diffusion models to upsample the image, first  $64\times64\rightarrow256\times256$, and then  $256\times256\rightarrow1024\times1024$.}
    \label{fig:main_diagram}
\end{figure}

\FloatBarrier

\section{Background}
\label{sec:app_background}
\label{sec:app_background_diffusion}

Diffusion models are latent variable models with latents $\bz = \{\bz_t \,|\, t \in [0,1]\}$ that obey a \emph{forward process} $q(\bz|\bx)$ starting at data $\bx \sim p(\bx)$. This forward process is a Gaussian process that satisfies the Markovian structure:
\begin{align}
    q(\bz_t|\bx) = \mathcal{N}(\bz_t; \alpha_t \bx, \sigma_t^2 \bI), \quad
    q(\bz_t | \bz_s) = \mathcal{N}(\bz_t; (\alpha_t/\alpha_s)\bz_s, \sigma_{t|s}^2\bI)
\end{align}
where $0 \leq s < t \leq 1$, $\sigma^2_{t|s} = (1-e^{\lambda_t-\lambda_s})\sigma_t^2$, and $\alpha_t, \sigma_t$ specify a differentiable \emph{noise schedule} whose log signal-to-noise-ratio,
i.e., $\lambda_t = \log[\alpha_t^2/\sigma_t^2]$,  decreases with $t$ until $q(\bz_1) \approx \mathcal{N}(\bzero, \bI)$. 
For generation, the diffusion model is learned to \emph{reverse} this forward process.

Learning to reverse the forward process can be reduced to learning to denoise $\bz_t\sim q(\bz_t|\bx)$ into an estimate $\hat\bx_\theta(\bz_t, \lambda_t, \bc) \approx \bx$ for all $t$, where $\bc$ is an optional conditioning signal (such as text embeddings or a low resolution image) drawn from the dataset jointly with~$\bx$. This is accomplished training~$\hat\bx_\theta$ using a weighted squared error loss
\begin{align}
    \Eb{\bepsilon,t}{w(\lambda_t) \|\hat\bx_\theta(\bz_t, \lambda_t, \bc) - \bx \|^2_2}
\end{align}
where $t \sim \mathcal{U}([0, 1])$, $\bepsilon \sim \mathcal{N}(\bzero,\bI)$, and $\bz_t = \alpha_t\bx + \sigma_t\bepsilon$. This reduction of generation to denoising is justified as optimizing a weighted variational lower bound on the data log likelihood under the diffusion model, or as a form of denoising score matching \citep{vincent2011connection, song2019generative, ho2020denoising, kingma2021variational}. We use the $\bepsilon$-prediction parameterization, defined as $\hat\bx_\theta(\bz_t,\lambda_t,\bc) =  (\bz_t - \sigma_t \bepsilon_\theta(\bz_t,\lambda_t,\bc))/\alpha_t$, and we impose a squared error loss on $\bepsilon_\theta$ in $\bepsilon$ space with $t$ sampled according to a cosine schedule~\citep{nichol2021improved}. This corresponds to a particular weighting $w(\lambda_t)$ and leads to a scaled score estimate $\bepsilon_\theta(\bz_t,\lambda_t,\bc) \approx -\sigma_t \nabla_{\bz_t}\log p(\bz_t|\bc)$, where $p(\bz_t|\bc)$ is the true density of $\bz_t$ given $\bc$ under the forward process starting at~$\bx \sim p(\bx)$~\citep{ho2020denoising,kingma2021variational,song2020score}. Related model designs include the work of~\citep{tzen2019neural,kadkhodaie2020solving,kadkhodaie2021stochastic}.

To sample from the diffusion model, we start at $\bz_1 \sim \mathcal{N}(\bzero, \bI)$ and use the discrete time ancestral sampler \citep{ho2020denoising} and DDIM~\citep{song2020denoising} for certain models. DDIM follows the deterministic update rule
\begin{align}
    \bz_s = \alpha_s \hat\bx_\theta(\bz_t,\lambda_t,\bc) + \frac{\sigma_s}{\sigma_t}(\bz_t - \alpha_t\hat\bx_\theta(\bz_t,\lambda_t,\bc)) \label{eq:ddim_sampler}
\end{align}
where $s < t$ follow a uniformly spaced sequence from 1 to 0. The ancestral sampler arises from a reversed description of the forward process; noting that $q(\bz_s|\bz_t,\bx) = \mathcal{N}(\bz_s; \tilde\bmu_{s|t}(\bz_t,\bx), \tilde\sigma^2_{s|t}\bI)$, where
$\tilde\bmu_{s|t}(\bz_t,\bx) = e^{\lambda_t-\lambda_s}(\alpha_s/\alpha_t)\bz_t + (1-e^{\lambda_t-\lambda_s})\alpha_s\bx$ and $\tilde\sigma^2_{s|t} = (1-e^{\lambda_t-\lambda_s})\sigma_s^2$, it follows the stochastic update rule
\begin{align}
    \bz_s &= \tilde\bmu_{s|t}(\bz_t, \hat\bx_\theta(\bz_t,\lambda_t,\bc)) + \sqrt{(\tilde\sigma^2_{s|t})^{1-\gamma} (\sigma^2_{t|s})^\gamma}\, \bepsilon  \label{eq:ancestral_sampler}
\end{align}
where $\bepsilon \sim \mathcal{N}(\bzero,\bI)$, and $\gamma$ controls the stochasticity of the sampler~\citep{nichol2021improved}.

\section{Architecture Details}
\subsection{Efficient U-Net}
\label{sec:unet_vs_efficient_unet_appendix}
We introduce a new architectural variant, which we term Efficient U-Net, for our super-resolution models. We find our Efficient U-Net to be simpler, converges faster, and is more memory efficient compared to some prior implementations \cite{nichol2021improved}, especially for high resolutions. We make several key modifications to the U-Net architecture, such as shifting of model parameters from high resolution blocks to low resolution, scaling the skip connections by $\sfrac{1}{\sqrt{2}}$ similar to \cite{song2020score, saharia2021image} and reversing the order of downsampling/upsampling operations in order to improve the speed of the forward pass.
Efficient U-Net makes several key modifications to the typical U-Net model used in \cite{dhariwal2021diffusion, sahariac-palette}: 
\begin{itemize}
    \item We shift the model parameters from the high resolution blocks to the low resolution blocks, via adding more residual blocks for the lower resolutions. Since lower resolution blocks typically have many more channels, this allows us to increase the model capacity through more model parameters, without egregious memory and computation costs.
    \item When using large number of residual blocks at lower-resolution (e.g. we use 8 residual blocks at lower-resolutions compared to typical 2-3 residual blocks used in standard U-Net architectures \cite{dhariwal2021diffusion, saharia2021image}) we find that scaling the skip connections by $\sfrac{1}{\sqrt{2}}$ similar to \cite{song2020score, saharia2021image} significantly improves convergence speed.
    \item In a typical U-Net's downsampling block, the downsampling operation happens after the convolutions, and in an upsampling block, the upsampling operation happens prior the convolution. We reverse this order for both downsampling and upsampling blocks in order to significantly improve the speed of the forward pass of the U-Net, and find no performance degradation. 
\end{itemize}  
With these key simple modifications, Efficient U-Net is simpler, converges faster, and is more memory efficient compared to some prior U-Net implementations. \cref{fig:efficient_unet_fullarch} shows the full architecture of Efficient U-Net, while  \Cref{fig:efficient_unet_dblock,fig:efficient_unet_ublock} show detailed description of the Downsampling and Upsampling blocks of Efficient U-Net respectively. See  \cref{sec:unet_vs_efficientunet} for results.

\section{\benchmarkname}
\label{appendix:benchmark}

In this section, we describe our new benchmark for fine-grained analysis of text-to-image models, namely,  \benchmarkname. \benchmarkname consists of 11 categories with approximately 200 text prompts. This is large enough to test the model well, while small enough to easily perform trials with human raters. Table \ref{tab:drawbench-categories} enumerates these categories along with description and few examples. We release the full set of samples \href{https://docs.google.com/spreadsheets/d/1y7nAbmR4FREi6npB1u-Bo3GFdwdOPYJc617rBOxIRHY/edit#gid=0}{\color{blue}{here}}.

For evaluation on this benchmark, we conduct an independent human evaluation run for each category. For each prompt, the rater is shown two sets of images - one from Model A, and second from Model B. Each set contains 8 random (non-cherry picked)  generations from the corresponding model. The rater is asked two questions - 
\begin{enumerate}[noitemsep,nolistsep]
    \item Which set of images is of higher quality?
    \item Which set of images better represents the text caption : \{Text Caption\}?
\end{enumerate}
where the questions are designed to measure: 1) image fidelity, and 2) image-text alignment. For each question, the rater is asked to select from three choices:
\begin{enumerate}[noitemsep,nolistsep]
    \item I prefer set A.
    \item I am indifferent.
    \item I prefer set B.
\end{enumerate}

We aggregate scores from 25 raters for each category (totalling to $25 \times 11 = 275$ raters). We do not perform any post filtering of the data to identify unreliable raters, both for expedience and because the task was straightforward to explain and execute. 

\setlength{\tabcolsep}{7pt}
\begin{table}[t]
    \centering
    \resizebox{\textwidth}{!}{
    {\small
    \begin{tabular}{lcc}
    \toprule
    \bfseries{Category} & \bfseries{Description} & \bfseries{Examples}  \\
    \midrule
    \multirow{2}{*}{Colors} & Ability to generate objects & ``A blue colored dog.'' \\
    &  with specified colors. &  ``A black apple and a green backpack.'' \\
    \midrule
    \multirow{2}{*}{Counting} & Ability to generate specified & ``Three cats and one dog sitting on the grass.'' \\
    &  number of objects. & ``Five cars on the street.'' \\
    \midrule
    \multirow{2}{*}{Conflicting} & Ability to generate conflicting & ``A horse riding an astronaut.'' \\
    & interactions b/w objects. & ``A panda making latte art.'' \\
    \midrule
    \multirow{2}{*}{DALL-E \cite{ramesh-dalle}} & Subset of challenging prompts & ``A triangular purple flower pot.'' \\
    & from \cite{ramesh-dalle}. & ``A cross-section view of a brain.'' \\
    \midrule
    \multirow{2}{*}{Description} & Ability to understand complex and long & ``A small vessel propelled on water by oars, sails, or an engine.'' \\
    & text prompts describing objects. & ``A mechanical or electrical device for measuring time.'' \\
    \midrule
    \multirow{2}{*}{Marcus et al. \cite{marcus-arxiv-2022}} & Set of challenging prompts & ``A pear cut into seven pieces arranged in a ring.'' \\
    & from \cite{marcus-arxiv-2022}. & ``Paying for a quarter-sized pizza with a pizza-sized quarter.'' \\
    \midrule
    \multirow{2}{*}{Misspellings} & Ability to understand & ``Rbefraigerator.'' \\
    &  misspelled prompts. & ``Tcennis rpacket.'' \\
    \midrule
    \multirow{2}{*}{Positional} & Ability to generate objects with & ``A car on the left of a bus.'' \\
    &  specified spatial positioning. & ``A stop sign on the right of a refrigerator.'' \\
    \midrule
    \multirow{2}{*}{Rare Words} & \multirow{2}{*}{Ability to understand rare words\footnote{https://www.merriam-webster.com/topics/obscure-words}.} & ``Artophagous.'' \\
    & & ``Octothorpe.'' \\
    \midrule
    \multirow{2}{*}{Reddit} & Set of challenging prompts from & ``A yellow and black bus cruising through the rainforest.'' \\
    & DALLE-2 Reddit\footnote{https://www.reddit.com/r/dalle2/}.& ``A medieval painting of the wifi not working.'' \\
    \midrule
    \multirow{2}{*}{Text} & \multirow{2}{*}{Ability to generate quoted text.} & ``A storefront with 'Deep Learning' written on it.'' \\
    & & ``A sign that says 'Text to Image'.'' \\
    \bottomrule
    \end{tabular}
    }
    }
    \vspace*{0.3cm}
    \caption{Description and examples of the 11 categories in \benchmarkname.}
    \label{tab:drawbench-categories}
\end{table}

\section{\name Detailed Abalations and Analysis}
\label{sec:experiments_analysis}
In this section, we perform ablations and provide a detailed analysis of \name.
\subsection{Pre-trained Text Encoders}
\label{sec:frozen_text_encoders}
We explore several families of pre-trained text encoders: BERT \cite{devlin-naacl-2019}, T5 \cite{raffel-jmlr-2020}, and CLIP \cite{radford-icml-2021}. 
There are several key differences between these encoders. BERT is trained on a smaller text-only corpus (approximately 20 GB, Wikipedia and BooksCorpus \cite{zhu-iccv-2015}) with a masking objective, and has relatively small model variants (upto 340M parameters). T5 is trained on a much larger C4 text-only corpus (approximately 800 GB) with a denoising objective, and has larger model variants (up to 11B parameters). The CLIP model\footnote{\url{https://github.com/openai/CLIP/blob/main/model-card.md}} is trained on an image-text corpus with an image-text contrastive objective. For T5 we use the encoder part for the contextual embeddings. For CLIP, we use the penultimate layer of the text encoder to get contextual embeddings. Note that we freeze the weights of these text encoders (i.e., we use off the shelf text encoders, without any fine-tuning on the text-to-image generation task). We explore a variety of model sizes for these text encoders.

We train a $64\times64$, 300M parameter diffusion model, conditioned on the text embeddings generated from BERT (base, and large), T5 (small, base, large, XL, and XXL), and CLIP (ViT-L/14). We observe that scaling the size of the language model text encoders generally results in better image-text alignment as captured by the CLIP score as a function of number of training steps (see \cref{Fig:EmbeddingTrainingCurves}).
One can see that the best CLIP scores are obtained with the T5-XXL text encoder. 

Since guidance weights are used to control image quality and text alignment, we also report ablation results using curves that show the trade-off between CLIP and FID scores as a function of the guidance weights (see \cref{fig:embedding_comparison_pareto}). We observe that larger variants of T5 encoder results in both better image-text alignment, and image fidelity. This emphasizes the effectiveness of large frozen text encoders for text-to-image models. Interestingly, we also observe that the T5-XXL encoder is on-par with the CLIP encoder when measured with CLIP and FID-10K on MS-COCO.

\textbf{T5-XXL vs CLIP on DrawBench}: We further compare T5-XXL and CLIP on \benchmarkname to perform a more comprehensive comparison of the abilities of these two text encoders. In our initial evaluations we observed that the 300M parameter models significantly underperformed on \benchmarkname. We believe this is primarily because \benchmarkname prompts are considerably more difficult than MS-COCO prompts. 

In order to perform a meaningful comparison, we train 64$\times$64 1B parameter diffusion models with T5-XXL and CLIP text encoders for this evaluation. \cref{fig:drawit_t5_vs_clip} shows the results. We find that raters are considerably more likely to prefer the generations from the model trained with the T5-XXL encoder over the CLIP text encoder, especially for image-text alignment. This indicates that language models are better than text encoders trained on image-text contrastive objectives in encoding complex and compositional text prompts. 
\cref{fig:t5xx_vs_clip_drawbench_details} shows the category specific comparison between the two models. We observe that human raters prefer T5-XXL samples over CLIP samples in all 11 categories for image-text alignment  demonstrating the effectiveness of large language models as text encoders for text to image generation.  

\begin{figure}[t]
    \centering
    \begin{subfigure}[t]{0.49\textwidth}
    \begin{tikzpicture}
\definecolor{google_blue}{RGB}{66,133,244}
\definecolor{google_red}{RGB}{219,68,55}
\definecolor{google_yellow}{RGB}{194,144,0}
\definecolor{google_green}{RGB}{15,157,88}

\begin{axis}[
width=\textwidth,
height=0.75\textwidth,
  xlabel=CLIP Score,
  ylabel=FID-10K,
  legend columns=1,
  legend pos=north west,
  legend cell align=left,
  legend style={font=\tiny},
  xmin=0.216,
  ymin=7.5,
  ymax=29,
  xmax=0.295,
  every axis plot/.append style={very thick}
]

\addplot +[mark=*, solid, google_yellow, mark options={scale=0.4}] coordinates {(0.2186, 17.1279)(0.2272, 14.9052)(0.2332, 14.4287)(0.2379, 14.0655)(0.2418, 14.1945)(0.2499, 16.339)(0.2543, 18.8731)(0.2563, 21.2413)(0.2586, 22.5724)(0.2596, 23.7404)(0.2604, 24.7911)(0.2609, 25.5612)(0.2615, 26.0648)};

\addplot +[mark=*, solid, google_green, mark options={scale=0.4}] coordinates {(0.2242, 15.7681)(0.2338, 13.4837)(0.2412, 12.5508)(0.2461, 12.2109)(0.25, 12.7537)(0.2594, 15.1684)(0.2638, 17.8411)(0.2666, 19.9616)(0.2683, 21.6277)(0.2691, 22.8355)(0.2701, 23.7544)(0.271, 24.3638)(0.2714, 25.1119)};

\addplot +[mark=*, solid, google_red, mark options={scale=0.4}] coordinates {(0.2316, 14.4068)(0.2422, 12.2246)(0.2499, 11.2804)(0.2552, 11.3188)(0.2593, 11.6776)(0.2685, 15.0103)(0.2731, 17.9712)(0.2749, 20.454)(0.2767, 22.132)(0.2774, 23.5054)(0.2783, 24.6424)(0.2788, 25.485)(0.2792, 26.1941)};

\addplot +[mark=*, solid, google_blue, mark options={scale=0.4}] coordinates {(0.2392, 13.33)(0.2508, 11.0229)(0.2584, 10.688)(0.2637, 11.2378)(0.268, 12.0375)(0.2764, 16.3064)(0.2797, 19.7633)(0.2822, 22.2284)(0.283, 24.3269)(0.2836, 25.5622)(0.2843, 26.6161)(0.2845, 27.1637)(0.2847, 28.0816)};

\addplot +[mark=square*, dotted, black, mark options={scale=0.4}] coordinates {(0.2386, 13.417) (0.2496, 11.0894) (0.257, 10.5947) (0.263, 11.1963) (0.2668, 11.7067) (0.2753, 15.7015) (0.2789, 19.187) (0.2809, 21.5909)(0.2823, 23.1788) (0.2825, 24.6649) (0.2835, 25.5378) (0.2839, 26.3481) (0.2841, 26.8251)};

\legend{T5-Small, T-Large, T5-XL, T5-XXL, CLIP}
\end{axis}
\end{tikzpicture}
    \caption{Pareto curves comparing various text encoders.}
     \label{fig:embedding_comparison_pareto}
    \end{subfigure}\hfill
    \begin{subfigure}[t]{0.49\textwidth}
     \centering
     \begin{tikzpicture}

\definecolor{google_blue}{RGB}{66,133,244}
\definecolor{google_red}{RGB}{219,68,55}
\definecolor{google_yellow}{RGB}{194,144,0}
\definecolor{google_green}{RGB}{15,157,88}

\begin{axis}[
name=fidelity,
width=\textwidth,
height=0.75\textwidth,
ybar,
ymin=0.0,
ymax=1.0,
xtick={1, 2},
xtick style={draw=none},
xticklabels={Alignment, Fidelity},
xticklabel style={yshift=-9pt},
ytick={0,0.5,1},
yticklabels={0\%, 50\%, 100\%},
legend style={font=\tiny},
legend cell align=left,
legend image code/.code={
    \draw [#1] (0cm, -0.1cm) rectangle (0.2cm, 0.25cm);
},
legend style={
draw=none,
at={(0.5, 0.925)}, anchor=north,
legend columns=3,
/tikz/every even column/.append style={column sep=0.2cm}
},
bar width=0.4cm,
enlarge x limits={abs=0.4},
]

\addplot+[
    fill=google_blue,
    fill opacity=0.2,
    draw=google_blue,
    error bars/error bar style={google_blue},
    error bars/.cd,
    y dir=both,
    y explicit
] coordinates {
(1.0,0.5867) +- (1.0, 0.02300842393320771)
(2.0,0.530) +- (2.0, 0.02074325857730084)
};
\addplot+[
fill=google_green,fill opacity=0.2,draw=google_green,
    error bars/error bar style={google_green},
    error bars/.cd,
    y dir=both,
    y explicit
] coordinates {
(1.0,0.4133) +- (1.0, 0.02158669186220008)
(2.0,0.470) +- (2.0, 0.019597550393515743)
};

\legend{T5-XXL, CLIP}
\end{axis}
\end{tikzpicture}
      \caption{Comparing T5-XXL and CLIP on \benchmarkname.}
      \label{fig:drawit_t5_vs_clip}
    \end{subfigure}
    \caption{Comparison between text encoders for text-to-image generation. For \cref{fig:embedding_comparison_pareto}, we sweep over guidance values of $[1, 1.25, 1.5, 1.75, 2, 3, 4, 5, 6, 7, 8, 9, 10]$}
\end{figure}
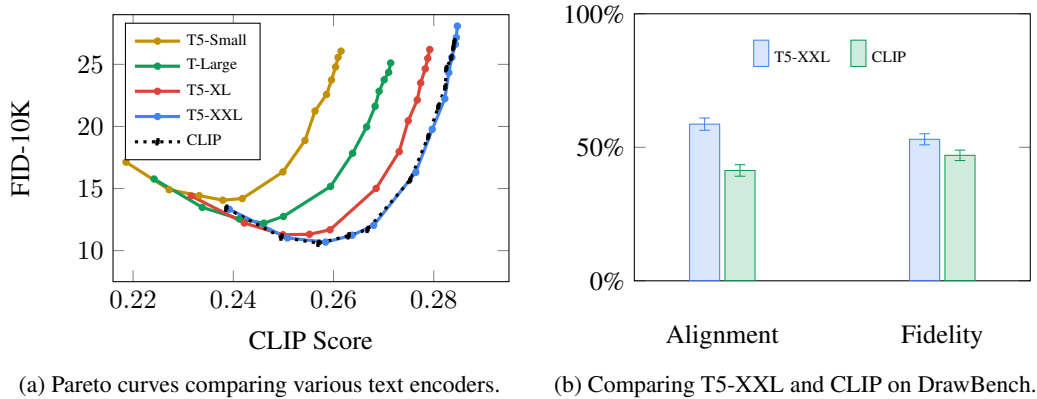

\begin{figure}[t]
    \centering
    \begin{tikzpicture}

\definecolor{google_blue}{RGB}{66,133,244}
\definecolor{google_red}{RGB}{219,68,55}
\definecolor{google_yellow}{RGB}{194,144,0}
\definecolor{google_green}{RGB}{15,157,88}

\begin{axis}[width=10cm, height=5cm,
  xlabel=Training Steps,
  ylabel=CLIP Score,
  legend columns=4,
  legend pos=south east,
  legend style={font=\tiny},
  legend cell align=left,
  xmin=0,
  ymin=0.22,
  xmax=500000,
  every axis plot/.append style={thick}
]
\addplot +[mark=none, dashed, google_blue] table [x=Step, y=Clip Score, col sep=comma] {embedding_comparison/bert_base.csv};
\addplot +[mark=none, dashed, google_red] table [x=Step, y=Clip Score, col sep=comma] {embedding_comparison/bert_large.csv};

\addplot +[mark=none, solid, google_yellow] table [x=Step, y=Clip Score, col sep=comma] {embedding_comparison/t5_small.csv};
\addplot +[mark=none, solid, google_blue] table [x=Step, y=Clip Score, col sep=comma] {embedding_comparison/t5_base.csv};
\addplot +[mark=none, solid, google_red] table [x=Step, y=Clip Score, col sep=comma] {embedding_comparison/t5_large.csv};
\addplot +[mark=none, solid, google_green] table [x=Step, y=Clip Score, col sep=comma] {embedding_comparison/t5_xl.csv};
\addplot +[mark=none, solid, gray] table [x=Step, y=Clip Score, col sep=comma] {embedding_comparison/t5_xxl.csv};
\addplot +[mark=none, dotted] table [x=Step, y=Clip Score, col sep=comma] {embedding_comparison/clip.csv};
\addplot[mark=none, dashdotted] coordinates {(0,0.27089) (1e6,0.27089)};
\legend{BERT Base, BERT Large, T5 Small, T5 Base, T5 Large, T5 XL, T5 XXL, CLIP, Reference}
\end{axis}
\end{tikzpicture}
    \caption{Training convergence comparison between text encoders for text-to-image generation.}
    \label{Fig:EmbeddingTrainingCurves}
\end{figure}
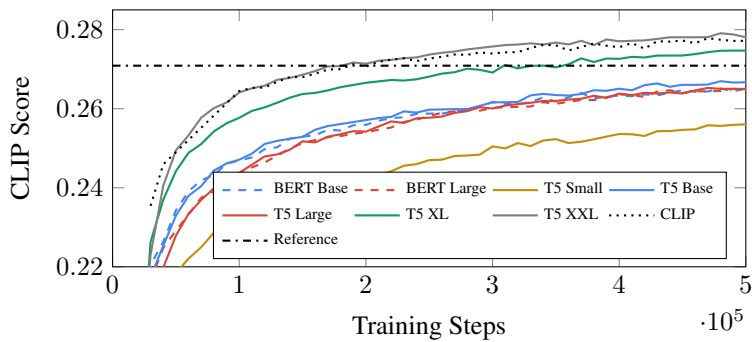

\begin{figure}
    \centering
    \begin{subfigure}[t]{\textwidth}
    \begin{tikzpicture}

\definecolor{google_blue}{RGB}{66,133,244}
\definecolor{google_red}{RGB}{219,68,55}
\definecolor{google_yellow}{RGB}{194,144,0}
\definecolor{google_green}{RGB}{15,157,88}

\begin{axis}[
name=fidelity,
width=\textwidth,
height=0.5\textwidth,
ybar stacked,
ymin=-0.1,
ymax=1.1,
xtick=data,
xtick style={draw=none},
symbolic x coords={Colors,Conflicting,Counting,DALL-E \cite{ramesh-dalle},Descriptions,Marcus et al. \cite{marcus-arxiv-2022},Misspellings,Positional,Rare Words,Reddit,Text},
xticklabels={},
x tick label style={rotate=30},
ytick={0,0.5,1},
yticklabels={0\%, 50\%, 100\%},
legend style={font=\tiny},
legend cell align=left,
legend image code/.code={
    \draw [#1] (0cm, -0.1cm) rectangle (0.2cm, 0.25cm);
},
legend style={
draw=none,
at={(0.5, 1.1)},anchor=south,
legend columns=3,
/tikz/every even column/.append style={column sep=0.2cm}
},
bar width=0.6cm,
]

\addplot+[
    fill=google_blue,
    fill opacity=0.2,
    draw=google_blue
] coordinates {
(Colors, 0.579)
(Conflicting, 0.517)
(Counting, 0.543)
(DALL-E \cite{ramesh-dalle}, 0.559)
(Descriptions, 0.749)
(Marcus et al. \cite{marcus-arxiv-2022}, 0.519)
(Misspellings, 0.632)
(Positional, 0.522)
(Rare Words, 0.519)
(Reddit, 0.512)
(Text, 0.785)
};

\addplot+[
fill=google_green,
fill opacity=0.2,
draw=google_green
] coordinates {
(Colors, 0.421)
(Conflicting, 0.483)
(Counting, 0.456)
(DALL-E \cite{ramesh-dalle}, 0.441)
(Descriptions, 0.251)
(Marcus et al. \cite{marcus-arxiv-2022}, 0.481)
(Misspellings, 0.368)
(Positional, 0.478)
(Rare Words, 0.481)
(Reddit, 0.488)
(Text, 0.214)
};

\legend{T5-XXL, CLIP}
\end{axis}

\end{tikzpicture}
    \caption{Alignment}
    \label{fig:t5xxl_vs_clip_detailed_alignment_alignment}
    \vspace*{0.5cm}
    \end{subfigure}
    \begin{subfigure}[t]{\textwidth}
    \begin{tikzpicture}

\definecolor{google_blue}{RGB}{66,133,244}
\definecolor{google_red}{RGB}{219,68,55}
\definecolor{google_yellow}{RGB}{194,144,0}
\definecolor{google_green}{RGB}{15,157,88}

\begin{axis}[
name=fidelity,
width=\textwidth,
height=0.5\textwidth,
ybar stacked,
ymin=-0.1,
ymax=1.1,
xtick=data,
xtick style={draw=none},
symbolic x coords={Color,Conflicting,Count,DALL-E \cite{ramesh-dalle},Descriptions,Marcus et al. \cite{marcus-arxiv-2022},Misspellings,Positional,Rarewords,Reddit,Text},
x tick label style={rotate=60},
ytick={0,0.5,1},
yticklabels={0\%, 50\%, 100\%},
legend style={font=\tiny},
legend cell align=left,
legend image code/.code={
    \draw [#1] (0cm, -0.1cm) rectangle (0.2cm, 0.25cm);
},
legend style={
draw=none,
at={(0.5, 1.1)},anchor=south,
legend columns=3,
/tikz/every even column/.append style={column sep=0.2cm}
},
bar width=0.6cm,
]
\addplot+[
    fill=google_blue,
    fill opacity=0.2,
    draw=google_blue
] coordinates {
(Color, 0.552)
(Conflicting, 0.476)
(Count, 0.532)
(DALL-E \cite{ramesh-dalle}, 0.509)
(Descriptions, 0.671)
(Marcus et al. \cite{marcus-arxiv-2022}, 0.453)
(Misspellings, 0.591)
(Positional, 0.538)
(Rarewords, 0.536)
(Reddit, 0.457)
(Text, 0.507)
};

\addplot+[
fill=google_green,
fill opacity=0.2,
draw=google_green
] coordinates {
(Color, 0.448)
(Conflicting, 0.524)
(Count, 0.468)
(DALL-E \cite{ramesh-dalle}, 0.491)
(Descriptions, 0.329)
(Marcus et al. \cite{marcus-arxiv-2022}, 0.547)
(Misspellings, 0.409)
(Positional, 0.462)
(Rarewords, 0.464)
(Reddit, 0.543)
(Text, 0.493)
};

\legend{}
\end{axis}
\end{tikzpicture}
    \caption{Fidelity}
    \label{fig:t5xxl_vs_clip_detailed_alignment_fidelity}
    \end{subfigure}
    \caption{T5-XXL vs. CLIP text encoder on \benchmarkname \subref{fig:t5xxl_vs_clip_detailed_alignment_alignment}) image-text alignment, and \subref{fig:t5xxl_vs_clip_detailed_alignment_fidelity}) image fidelity.}
    \label{fig:t5xx_vs_clip_drawbench_details}
\end{figure}

\subsection{Classifier-free Guidance and the Alignment-Fidelity Trade-off}

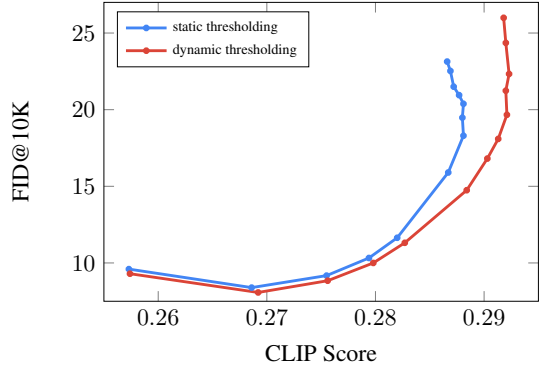
\begin{figure}
\centering
\begin{tikzpicture}[scale=0.9]

\definecolor{google_blue}{RGB}{66,133,244}
\definecolor{google_red}{RGB}{219,68,55}
\definecolor{google_yellow}{RGB}{194,144,0}
\definecolor{google_green}{RGB}{15,157,88}

\begin{axis}[
width=8cm,
height=6cm,
  xlabel=CLIP Score,
  ylabel=FID@10K,
  legend columns=1,
  legend pos=north west,
  legend style={font=\tiny},
  legend cell align=left,
  xmin=0.255,
  ymin=7.5,
  ymax=27,
  xmax=0.295,
  every axis plot/.append style={very thick}
]
\addplot +[mark=*, solid, google_blue, mark options={scale=0.4}] coordinates {(0.2573, 9.5976)(0.2686, 8.3913)(0.2755, 9.1741)(0.2794, 10.3238)(0.282, 11.6418)(0.2867, 15.9009)(0.2881, 18.2998)(0.288, 19.4781)(0.2881, 20.3822)(0.2877, 20.9476)(0.2872, 21.4995)(0.2869, 22.526)(0.2866, 23.1381)};

\addplot +[mark=*, solid, google_red, mark options={scale=0.4}] coordinates {(0.2574, 9.2972)(0.2692, 8.0746)(0.2756, 8.8401)(0.2798, 10.0002)(0.2827, 11.3142)(0.2884, 14.7455)(0.2903, 16.8102)(0.2913, 18.092)(0.2921, 19.6666)(0.292, 21.2374)(0.2923, 22.3307)(0.292, 24.3588)(0.2918, 25.9931)};

\legend{static thresholding, dynamic thresholding}
\end{axis}
\end{tikzpicture}
\caption{CLIP Score vs FID trade-off across various $\hat{\bx}_0$ thresholding methods for the 64$\times$64 model. We sweep over guidance values of $[1, 1.25, 1.5, 1.75, 2, 3, 4, 5, 6, 7, 8, 9, 10]$.
}
\label{fig:clip_vs_scaled_clip}
\end{figure}

\begin{figure}[tb]
\centering
\begin{subfigure}{.33\textwidth}
\centering
\setlength{\tabcolsep}{0.1pt}
\begin{tabular}{cccc}
\includegraphics[width=0.24\textwidth]{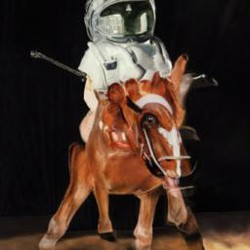} &
\includegraphics[width=0.24\textwidth]{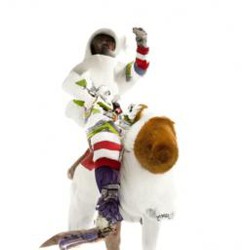} &
\includegraphics[width=0.24\textwidth]{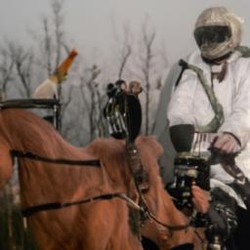} &
\includegraphics[width=0.24\textwidth]{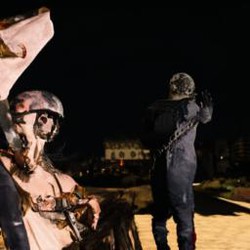} \\ [-3pt]

\includegraphics[width=0.24\textwidth]{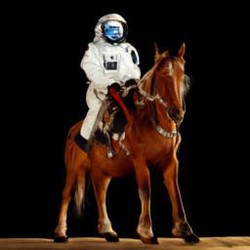} &
\includegraphics[width=0.24\textwidth]{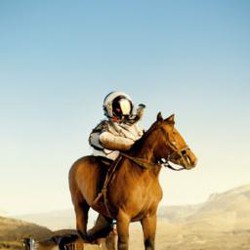} &
\includegraphics[width=0.24\textwidth]{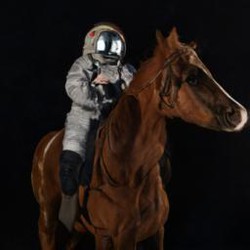} &
\includegraphics[width=0.24\textwidth]{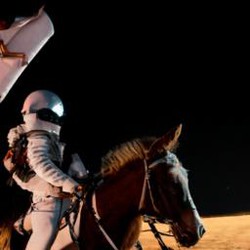} \\ [-3pt]

\includegraphics[width=0.24\textwidth]{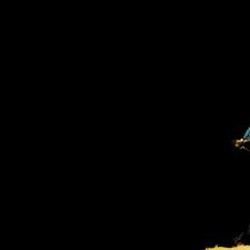} &
\includegraphics[width=0.24\textwidth]{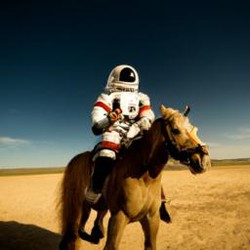} &
\includegraphics[width=0.24\textwidth]{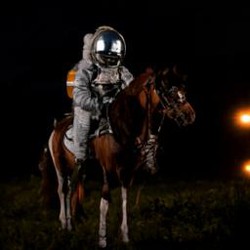} &
\includegraphics[width=0.24\textwidth]{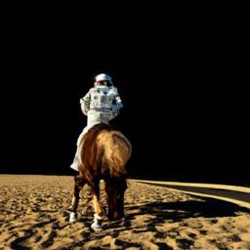} \\ [-3pt]

\includegraphics[width=0.24\textwidth]{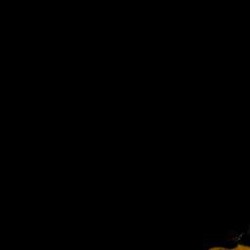} &
\includegraphics[width=0.24\textwidth]{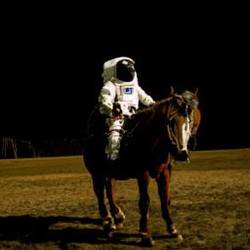} &
\includegraphics[width=0.24\textwidth]{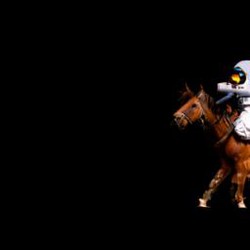} &
\includegraphics[width=0.24\textwidth]{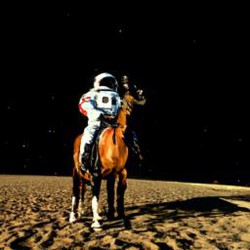} \\ [-3pt]

\includegraphics[width=0.24\textwidth]{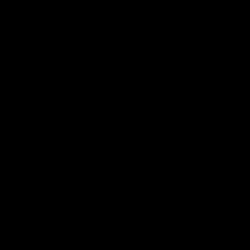} &
\includegraphics[width=0.24\textwidth]{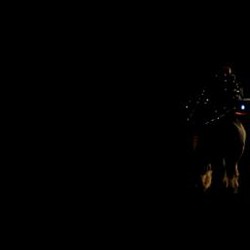} &
\includegraphics[width=0.24\textwidth]{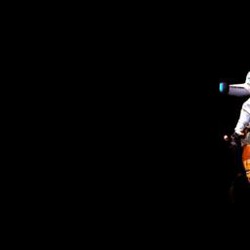} &
\includegraphics[width=0.24\textwidth]{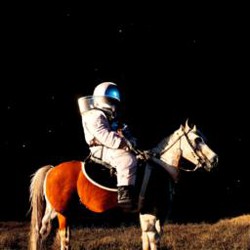} \\
\end{tabular}
\vspace*{-0.2cm}
\caption{No thresholding.}
\end{subfigure}%
\hfill
\begin{subfigure}{.33\textwidth}
\centering
\setlength{\tabcolsep}{0.1pt}
\begin{tabular}{ccccc}
\includegraphics[width=0.24\textwidth]{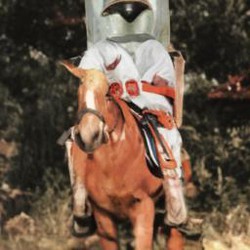} &
\includegraphics[width=0.24\textwidth]{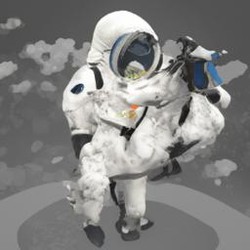} &
\includegraphics[width=0.24\textwidth]{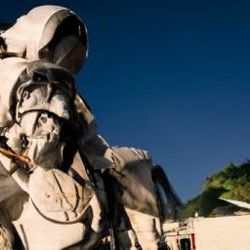} &
\includegraphics[width=0.24\textwidth]{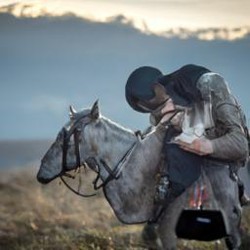} \\ [-3pt]

\includegraphics[width=0.24\textwidth]{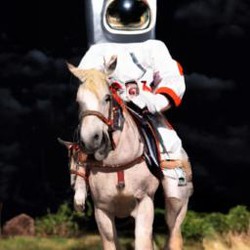} &
\includegraphics[width=0.24\textwidth]{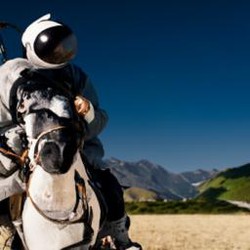} &
\includegraphics[width=0.24\textwidth]{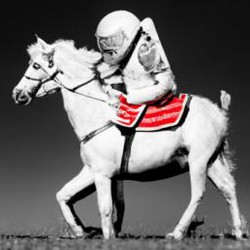} &
\includegraphics[width=0.24\textwidth]{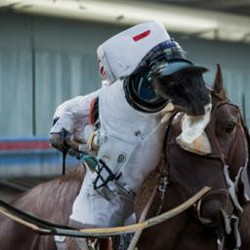} \\ [-3pt]

\includegraphics[width=0.24\textwidth]{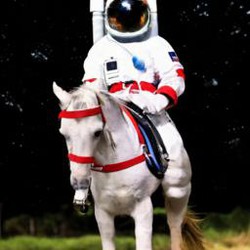} &
\includegraphics[width=0.24\textwidth]{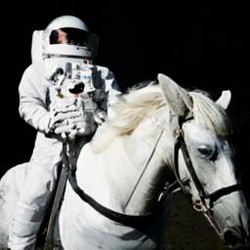} &
\includegraphics[width=0.24\textwidth]{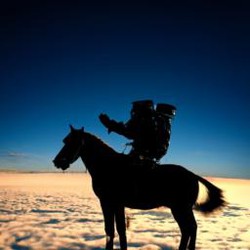} &
\includegraphics[width=0.24\textwidth]{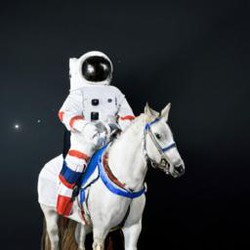} \\ [-3pt]

\includegraphics[width=0.24\textwidth]{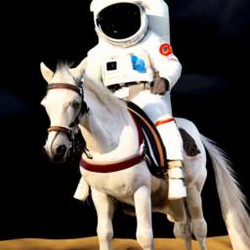} &
\includegraphics[width=0.24\textwidth]{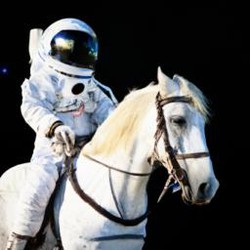} &
\includegraphics[width=0.24\textwidth]{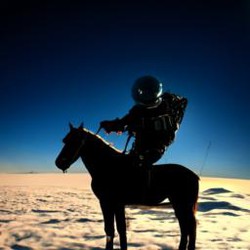} &
\includegraphics[width=0.24\textwidth]{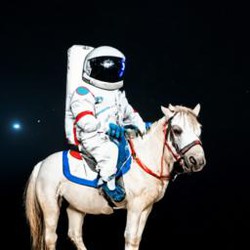} \\ [-3pt]

\includegraphics[width=0.24\textwidth]{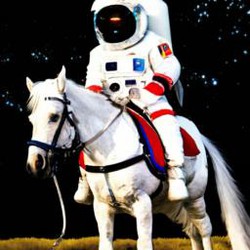} &
\includegraphics[width=0.24\textwidth]{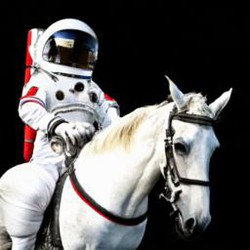} &
\includegraphics[width=0.24\textwidth]{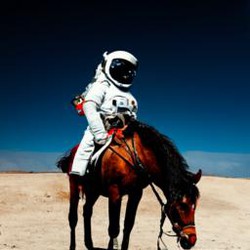} &
\includegraphics[width=0.24\textwidth]{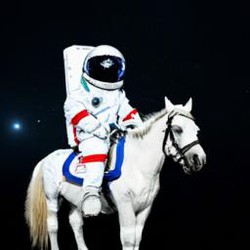} \\
\end{tabular}
\vspace*{-0.2cm}
\caption{Static thresholding.}
\end{subfigure}%
\hfill
\begin{subfigure}{.33\textwidth}
\centering
\setlength{\tabcolsep}{0.1pt}
\begin{tabular}{cccc}
\includegraphics[width=0.24\textwidth]{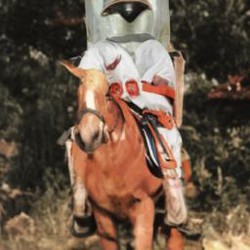} &
\includegraphics[width=0.24\textwidth]{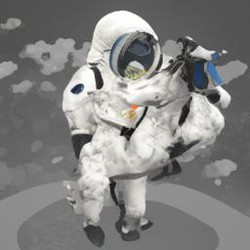} &
\includegraphics[width=0.24\textwidth]{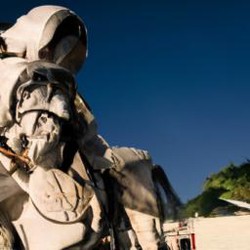} &
\includegraphics[width=0.24\textwidth]{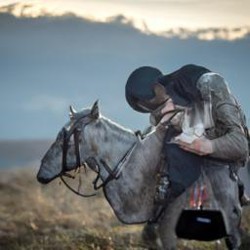} \\ [-3pt]

\includegraphics[width=0.24\textwidth]{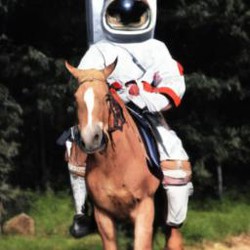} &
\includegraphics[width=0.24\textwidth]{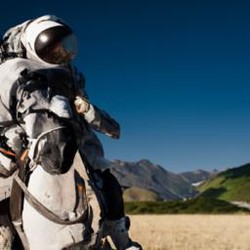} &
\includegraphics[width=0.24\textwidth]{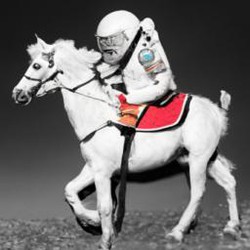} &
\includegraphics[width=0.24\textwidth]{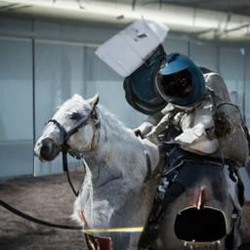} \\ [-3pt]

\includegraphics[width=0.24\textwidth]{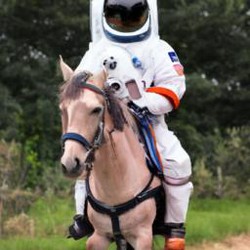} &
\includegraphics[width=0.24\textwidth]{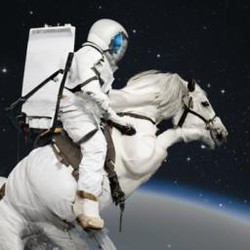} &
\includegraphics[width=0.24\textwidth]{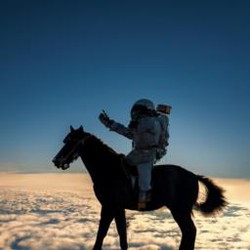} &
\includegraphics[width=0.24\textwidth]{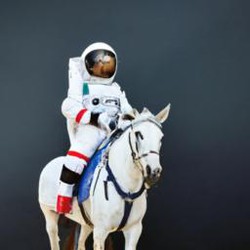} \\ [-3pt]

\includegraphics[width=0.24\textwidth]{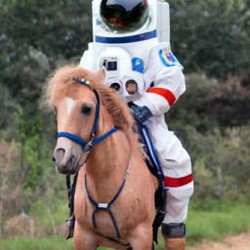} &
\includegraphics[width=0.24\textwidth]{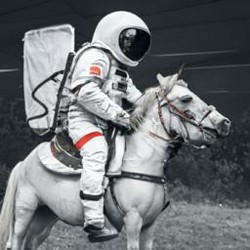} &
\includegraphics[width=0.24\textwidth]{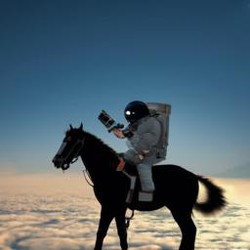} &
\includegraphics[width=0.24\textwidth]{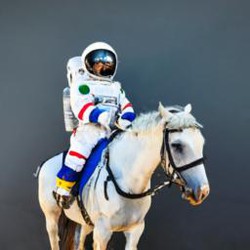} \\ [-3pt]

\includegraphics[width=0.24\textwidth]{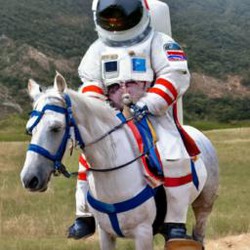} &
\includegraphics[width=0.24\textwidth]{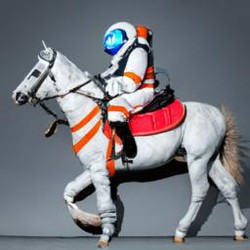} &
\includegraphics[width=0.24\textwidth]{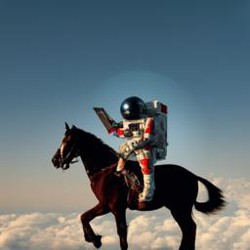} &
\includegraphics[width=0.24\textwidth]{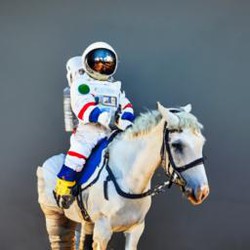} \\
\end{tabular}
\vspace*{-0.2cm}
\caption{Dynamic thresholding.}
\end{subfigure}

\caption{Thresholding techniques on $256\times256$ samples for ``A photo of an astronaut riding a horse.'' Guidance weights increase from 1 to 5 as we go from top to bottom. No thresholding results in poor images with high guidance weights. Static thresholding is an improvement but still leads to oversaturated samples. Our dynamic thresholding leads to the highest quality images.  See \cref{fig:clip_vs_scaled_clip_samples} for more  qualitative comparison.}
\label{fig:clip_noclip_scaledclip}
\end{figure}

\begin{figure}[tb]
\centering
\begin{subfigure}{.495\textwidth}
\centering
\setlength{\tabcolsep}{0.1pt}
\begin{tabular}{cccc}
\includegraphics[width=0.24\textwidth]{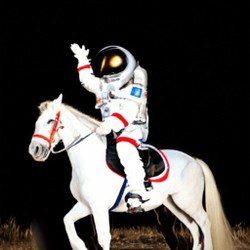} &
\includegraphics[width=0.24\textwidth]{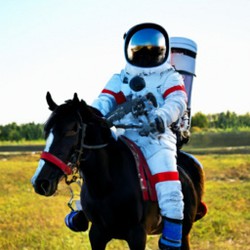} &
\includegraphics[width=0.24\textwidth]{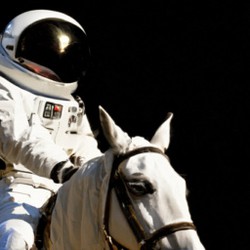} &
\includegraphics[width=0.24\textwidth]{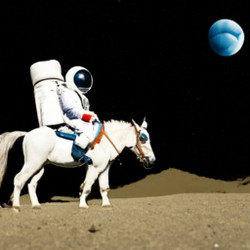} \\ [-3pt]

\includegraphics[width=0.24\textwidth]{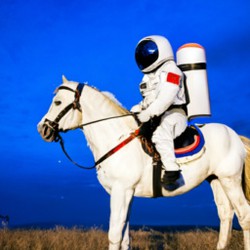} &
\includegraphics[width=0.24\textwidth]{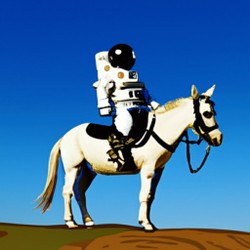} &
\includegraphics[width=0.24\textwidth]{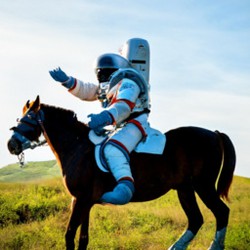} &
\includegraphics[width=0.24\textwidth]{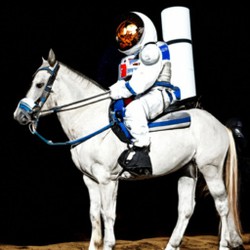} \\ [-3pt]

\includegraphics[width=0.24\textwidth]{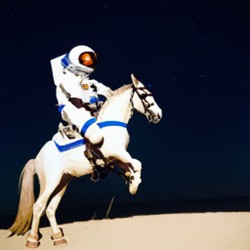} &
\includegraphics[width=0.24\textwidth]{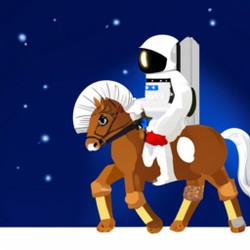} &
\includegraphics[width=0.24\textwidth]{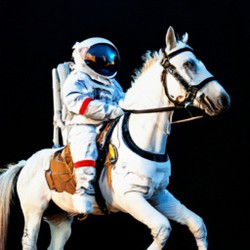} &
\includegraphics[width=0.24\textwidth]{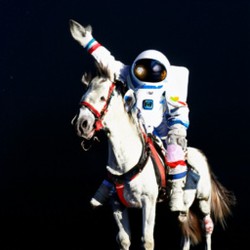} \\ [-3pt]

\includegraphics[width=0.24\textwidth]{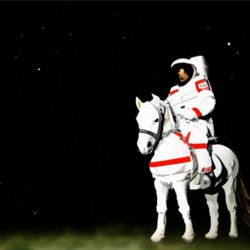} &
\includegraphics[width=0.24\textwidth]{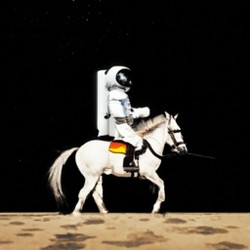} &
\includegraphics[width=0.24\textwidth]{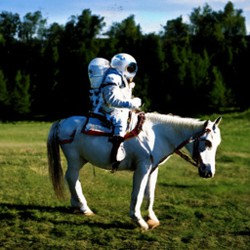} &
\includegraphics[width=0.24\textwidth]{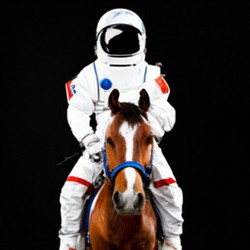} \\
\end{tabular}
\vspace*{-0.2cm}
\caption{Samples using static thresholding.}
\end{subfigure}%
\begin{subfigure}{.495\textwidth}
\centering
\setlength{\tabcolsep}{0.1pt}
\begin{tabular}{cccc}
\includegraphics[width=0.24\textwidth]{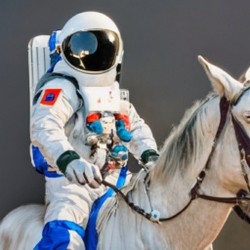} &
\includegraphics[width=0.24\textwidth]{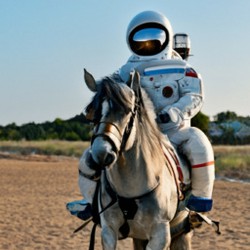} &
\includegraphics[width=0.24\textwidth]{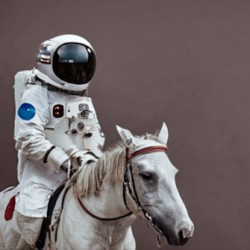} &
\includegraphics[width=0.24\textwidth]{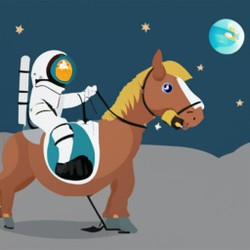} \\[-3pt] 

\includegraphics[width=0.24\textwidth]{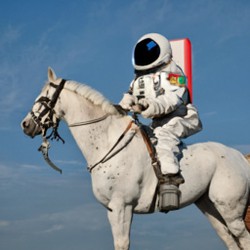} &
\includegraphics[width=0.24\textwidth]{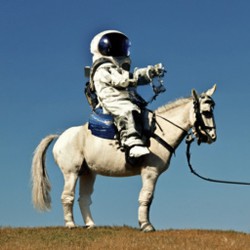} &
\includegraphics[width=0.24\textwidth]{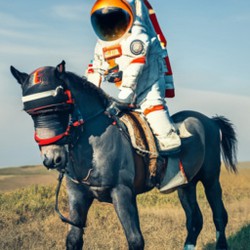} &
\includegraphics[width=0.24\textwidth]{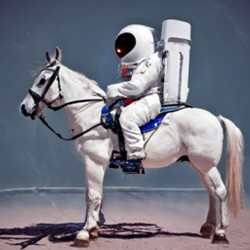} \\[-3pt]

\includegraphics[width=0.24\textwidth]{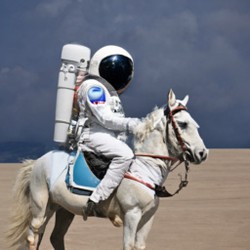} &
\includegraphics[width=0.24\textwidth]{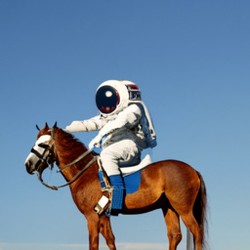} &
\includegraphics[width=0.24\textwidth]{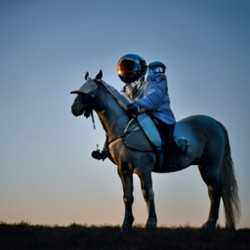} &
\includegraphics[width=0.24\textwidth]{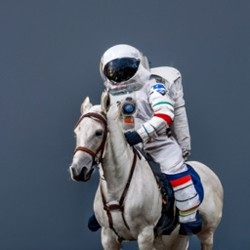} \\[-3pt]

\includegraphics[width=0.24\textwidth]{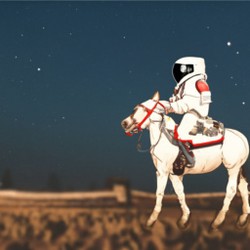} &
\includegraphics[width=0.24\textwidth]{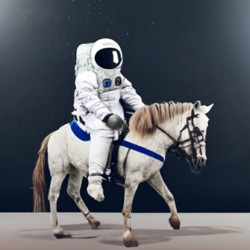} &
\includegraphics[width=0.24\textwidth]{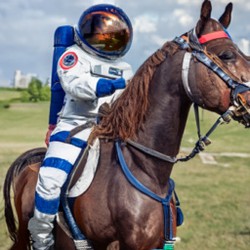} &
\includegraphics[width=0.24\textwidth]{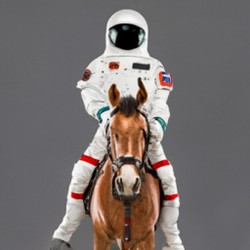} \\
\end{tabular}
\vspace*{-0.2cm}
\caption{Samples using dynamic thresholding ($p=99.5$)}
\end{subfigure}
\caption{Static vs.\ dynamic thresholding on non-cherry picked $256\times256$ samples using a guidance weight of 5 for both the base model and the super-resolution model, using the same random seed. The text prompt used for these samples is ``A photo of an astronaut riding a horse.'' When using high guidance weights, static thresholding often leads to oversaturated samples, while our dynamic thresholding yields more natural looking images.}
\label{fig:clip_vs_scaled_clip_samples}
\end{figure}

We observe that classifier-free guidance \cite{ho2021classifierfree} is a key contributor to generating samples with strong image-text alignment, this is also consistent with the observations of \cite{ramesh-dalle, ramesh-dalle2}. There is typically a trade-off between image fidelity and image-text alignment, as we iterate over the guidance weight. 
While previous work has typically used relatively small guidance weights, \name uses relatively large guidance weights for all three diffusion models.  We found this to yield a good balance of sample quality and alignment.
However, naive use of large guidance weights often produces relatively poor results. To enable the effective use of larger guidance we introduce several innovations, as described below.

\textbf{Thresholding Techniques}: First, we compare various thresholding methods used with classifier-free guidance. Fig.  \ref{fig:clip_vs_scaled_clip} compares the CLIP vs.\ FID-10K score pareto frontiers for various thresholding methods of the base text-to-image $64\times64$ model. We observe that our dynamic thresholding technique results in significantly better CLIP scores, and comparable or better FID scores than the static thresholding technique for a wide range of guidance weights. \cref{fig:clip_noclip_scaledclip} shows qualitative samples for thresholding techniques.

\begin{figure}[t]
    \centering
    \begin{subfigure}[t]{0.49\textwidth}
    \begin{tikzpicture}[scale=0.7]

\definecolor{google_blue}{RGB}{66,133,244}
\definecolor{google_red}{RGB}{219,68,55}
\definecolor{google_yellow}{RGB}{194,144,0}
\definecolor{google_green}{RGB}{15,157,88}

\begin{axis}[width=8cm, height=6cm,
  xlabel=CLIP Score,
  ylabel=FID-10K,
  legend columns=1,
  legend pos=south west,
  legend style={font=\tiny},
  xmin=0.285,
  ymin=10.0,
  ymax=13.5,
  xmax=0.305,
  xtick = {0.29, 0.30},
  xticklabels = {0.29, 0.30},
  every axis plot/.append style={very thick}
]
\addplot +[mark=*, solid, google_blue, mark options={scale=0.4}] coordinates {(0.2921, 11.5138)(0.294, 10.9842)(0.2954, 10.6339)(0.2966, 10.4398)(0.297, 10.3053)(0.2979, 10.2523)(0.2989, 10.2793)(0.2998, 10.4203)(0.3011, 10.8419)};

\addplot +[mark=*, solid, google_red, mark options={scale=0.4}] coordinates {(0.2903, 12.3925)(0.2942, 11.526)(0.2962, 11.1867)(0.2975, 11.1969)(0.2981, 11.0975)(0.299, 11.1523)(0.2999, 11.2705)(0.3007, 11.6897)(0.3015, 12.176)};

\addplot +[mark=*, solid, google_green, mark options={scale=0.4}] coordinates {(0.2888, 12.9354)(0.2944, 11.5655)(0.297, 11.3662)(0.2984, 11.4112)(0.299, 11.5807)(0.2999, 11.658)(0.301, 11.8975)(0.3016, 12.4118)(0.3025, 13.143)};

\legend{$\mathrm{aug\_level} = 0.0$, $\mathrm{aug\_level} = 0.1$, $\mathrm{aug\_level} = 0.2$}
\end{axis}
\end{tikzpicture}
    \caption{Comparison between different values of $\mathrm{aug\_level}$.}
     \label{fig:256x256_t_eval_and_gw}
    \end{subfigure}
    \begin{subfigure}[t]{0.49\textwidth}
     \centering
     \begin{tikzpicture}[scale=0.7]

\definecolor{google_blue}{RGB}{66,133,244}
\definecolor{google_red}{RGB}{219,68,55}
\definecolor{google_yellow}{RGB}{194,144,0}
\definecolor{google_green}{RGB}{15,157,88}

\begin{axis}[width=8cm, height=6cm,
  xlabel=CLIP Score,
  ylabel=FID-10K,
  legend columns=1,
  legend pos=south west,
  legend style={font=\tiny},
  xmin=0.285,
  ymin=9.4,
  ymax=13.5,
  xmax=0.297,
  xtick = {0.286, 0.29, 0.295},
  xticklabels = {0.286, 0.29, 0.295},
  every axis plot/.append style={very thick}
]
\addplot +[mark=*, solid, google_blue, mark options={scale=0.4}] coordinates {(0.2899, 11.5946)(0.29, 11.45)(0.2902, 11.3181)(0.2907, 11.1728)(0.2908, 11.1542)(0.2911, 11.1761)(0.2913, 11.038)(0.2915, 11.1443)(0.292, 11.341)};
\addplot +[mark=*, solid, google_red, mark options={scale=0.4}] coordinates {(0.2903, 12.029)(0.2918, 11.4921)(0.2928, 11.2132)(0.2934, 10.9514)(0.2936, 10.8039)(0.2942, 10.6215)(0.2947, 10.6294)(0.2955, 10.7032)(0.296, 10.974)};
\addplot +[mark=*, solid, google_green, mark options={scale=0.4}] coordinates {(0.2883, 12.4993)(0.2912, 11.7119)(0.2931, 11.5148)(0.2942, 11.3467)(0.2946, 11.2633)(0.2952, 11.3083)(0.2957, 11.3226)(0.2961, 11.5953)(0.2964, 11.9972)};
\addplot +[mark=*, solid, google_yellow, mark options={scale=0.4}] coordinates {(0.2861, 13.1623)(0.2907, 11.9674)(0.2927, 11.745)(0.2941, 11.5647)(0.2944, 11.5996)(0.2954, 11.7032)(0.2958, 12.0862)(0.2962, 12.3539)(0.2964, 12.8817)};
\legend{ A,  B ($\mathrm{aug\_level} = 0.0$),  B ($\mathrm{aug\_level} = 0.1$),  B ($\mathrm{aug\_level} = 0.2$)}
\end{axis}
\end{tikzpicture}
      \caption{Comparison between training with no noise augmentation ``A'' vs noise augmentation ``B''}
      \label{fig:super_res_ablation}
    \end{subfigure}
    \caption{CLIP vs FID-10K pareto curves showing the impact of noise augmentation on our $64\times64 \rightarrow 256\times256$ model. For each study, we sweep over guidance values of $[1, 3, 5, 7, 8, 10, 12, 15, 18]$}
\end{figure}
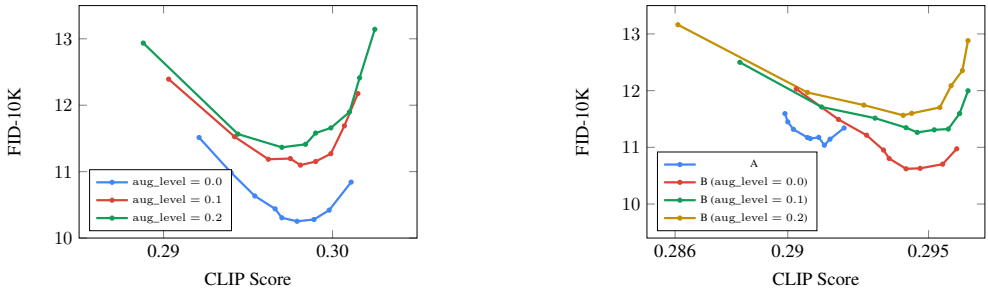

\begin{figure}[t]
\captionsetup[subfigure]{labelformat=empty}
\centering
\begin{tabular}{c|@{\hskip 6pt}c@{\hskip 2pt}c@{\hskip 2pt}c}
Input & Unmodified & Oil Painting & Illustration \\
\begin{subfigure}[t]{0.24\textwidth} \centering \includegraphics[width=\textwidth]{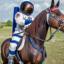} \caption{} \end{subfigure} &
\begin{subfigure}[t]{0.24\textwidth} \centering \includegraphics[width=\textwidth]{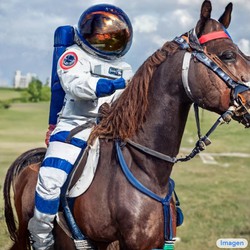} \caption{} \end{subfigure} &
\begin{subfigure}[t]{0.24\textwidth} \centering \includegraphics[width=\textwidth]{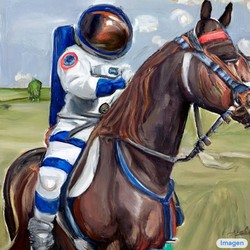} \caption{} \end{subfigure} &
\begin{subfigure}[t]{0.24\textwidth} \centering \includegraphics[width=\textwidth]{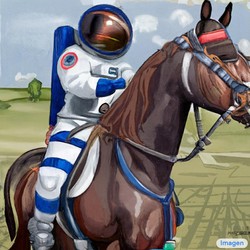} \caption{} \end{subfigure} \\[-12pt]

\begin{subfigure}[t]{0.24\textwidth} \centering \includegraphics[width=\textwidth]{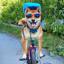} \caption{} \end{subfigure} &
\begin{subfigure}[t]{0.24\textwidth} \centering \includegraphics[width=\textwidth]{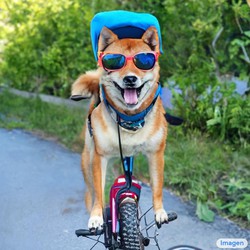} \caption{} \end{subfigure} &
\begin{subfigure}[t]{0.24\textwidth} \centering \includegraphics[width=\textwidth]{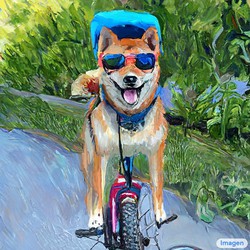} \caption{} \end{subfigure} &
\begin{subfigure}[t]{0.24\textwidth} \centering \includegraphics[width=\textwidth]{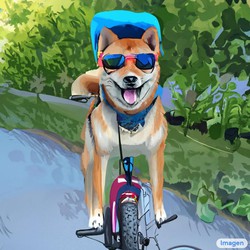} \caption{} \end{subfigure} \\[-12pt]

\begin{subfigure}[t]{0.24\textwidth} \centering \includegraphics[width=\textwidth]{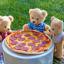} \caption{} \end{subfigure} &
\begin{subfigure}[t]{0.24\textwidth} \centering \includegraphics[width=\textwidth]{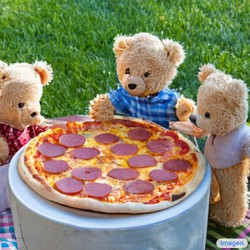} \caption{} \end{subfigure} &
\begin{subfigure}[t]{0.24\textwidth} \centering \includegraphics[width=\textwidth]{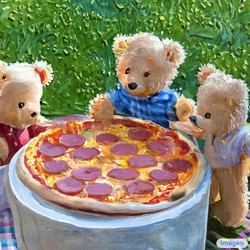} \caption{} \end{subfigure} &
\begin{subfigure}[t]{0.24\textwidth} \centering \includegraphics[width=\textwidth]{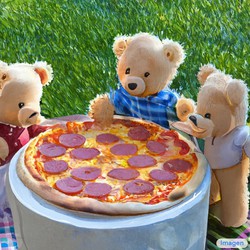} \caption{} \end{subfigure} \\

\end{tabular}
\caption{Super-resolution variations for some $64\times64$ generated images. We first generate the 64$\times$64 image using ``A photo of ... .''. Given generated $64\times64$ images, we condition both the super-resolution models on different prompts in order to generate different upsampled variations. e.g. for oil painting we condition the super-resolution models on the prompt ``An oil painting of ... .''. Through a combination of large guidance weights and $\mathrm{aug\_level} = 0.3$ for both super-res models we can generate different styles based on the style query through text.}
\label{fig:super_res_variations}
\end{figure}

\textbf{Guidance for Super-Resolution}: We further analyze the impact of classifier-free guidance for our $64\times64 \rightarrow 256\times256$ model. \cref{fig:256x256_t_eval_and_gw} shows the pareto frontiers for CLIP vs. FID-10K score for the $64\times64 \rightarrow 256\times256$ super-resolution model. $\mathrm{aug\_level}$ specifies the level of noise augmentation applied to the input low-resolution image during inference ($\mathrm{aug\_level} = 0$ means no noise). We observe that $\mathrm{aug\_level} = 0$ gives the best FID score for all values of guidance weight. Furthermore, for all values of $\mathrm{aug\_level}$, we observe that FID improves considerably with increasing guidance weight upto around $7-10$. While generation using larger values of $\mathrm{aug\_level}$ gives slightly worse FID, it allows more varied range of CLIP scores, suggesting more diverse generations by the super-resolution model. In practice, for our best samples, we generally use $\mathrm{aug\_level}$ in $[0.1, 0.3]$. Using large values of $\mathrm{aug\_level}$ and high guidance weights for the super-resolution models, \name can create different variations of a given $64\times64$ image by altering the prompts to the super-resolution models (See \cref{fig:super_res_variations} for examples).

\textbf{Impact of Conditioning Augmentation}: \cref{fig:super_res_ablation} shows the impact of training super-resolution models with noise conditioning augmentation. Training with no noise augmentation generally results in worse CLIP and FID scores, suggesting noise conditioning augmentation is critical to attaining best sample quality similar to prior work \cite{ho2021cascaded}. Interestingly, the model trained without noise augmentation has much less variations in CLIP and FID scores across different guidance weights compared to the model trained with conditioning augmentation. We hypothesize that this is primarily because strong noise augmented training reduces the low-resolution image conditioning signal considerably, encouraging higher degree of dependence on conditioned text for the model. 

\subsection{Impact of Model Size}

Fig. \ref{fig:model_size_comparison} plots the CLIP-FID score trade-off curves for various model sizes of the $64\times64$ text-to-image U-Net model. We train each of the models with a batch size of 2048, and 400K training steps. As we scale from 300M parameters to 2B parameters for the U-Net model, we obtain better trade-off curves with increasing model capacity. Interestingly, scaling the frozen text encoder model size yields more improvement in model quality over scaling the U-Net model size. Scaling with a frozen text encoder is also easier since the text embeddings can be computed and stored offline during training.

\subsubsection{Impact of Text Conditioning Schemas}
\label{sec:text_conditioning_ablation}

\begin{figure}[t]
    \centering
    \begin{subfigure}[t]{0.49\textwidth}
    \begin{tikzpicture}[scale=0.7]

\definecolor{google_blue}{RGB}{66,133,244}
\definecolor{google_red}{RGB}{219,68,55}
\definecolor{google_yellow}{RGB}{194,144,0}
\definecolor{google_green}{RGB}{15,157,88}

\begin{axis}[width=8cm, height=6cm,
  xlabel=CLIP Score,
  ylabel=FID-10K,
  legend columns=1,
  legend pos=north west,
  legend cell align=left,
  legend style={font=\tiny},
  xmin=0.23,
  ymin=7.5,
  ymax=29,
  xmax=0.295,
  every axis plot/.append style={very thick}
]
\addplot +[mark=*, solid, google_blue, mark options={scale=0.4}] coordinates {(0.2331, 12.9559)(0.2431, 10.874)(0.2497, 10.556)(0.254, 10.9103)(0.2573, 11.6425)(0.2651, 15.0507)(0.2684, 17.7926)(0.2705, 19.9939)(0.2716, 21.5785)(0.2726, 22.7001)(0.2733, 23.0793)(0.2734, 23.8903)(0.2738, 24.4911)};

\addplot +[mark=*, solid, google_red, mark options={scale=0.4}] coordinates {(0.2395, 12.0989)(0.2503, 10.0628)(0.2569, 9.9583)(0.2624, 10.6117)(0.2654, 11.3276)(0.2729, 15.1281)(0.2763, 18.5341)(0.2779, 20.1554)(0.2788, 21.6792)(0.2796, 22.8844)(0.28, 23.4682)(0.2802, 24.2021)(0.2805, 24.8261)};

\addplot +[mark=*, solid, google_yellow, mark options={scale=0.4}] coordinates {(0.2451, 11.3483)(0.2572, 9.4933)(0.2652, 9.4017)(0.2699, 10.2998)(0.2738, 11.3422)(0.281, 16.4762)(0.2837, 19.563)(0.285, 21.9303)(0.286, 23.5618)(0.2862, 24.8776)(0.2865, 25.403)(0.2862, 26.2234)(0.2865, 26.5037)};

\legend{Mean Pooling, Attention Pooling, Cross Attention}
\end{axis}
\end{tikzpicture}
    \caption{Comparison between different text encoders.}
     \label{fig:text_cond_comparison}
    \end{subfigure}
    \begin{subfigure}[t]{0.49\textwidth}
     \centering
     \begin{tikzpicture}[scale=0.7]

\definecolor{google_blue}{RGB}{66,133,244}
\definecolor{google_red}{RGB}{219,68,55}
\definecolor{google_yellow}{RGB}{194,144,0}
\definecolor{google_green}{RGB}{15,157,88}

\begin{axis}[width=8cm, height=6cm,
  xlabel=CLIP Score,
  ylabel=FID-10K,
  legend columns=1,
  legend pos=north west,
  legend style={font=\tiny},
  xmin=0.24,
  ymin=7.5,
  ymax=29,
  xmax=0.295,
  every axis plot/.append style={very thick}
]
\addplot +[mark=*, solid, google_blue, mark options={scale=0.4}] coordinates {(0.2451, 11.3483)(0.2572, 9.4933)(0.2652, 9.4017)(0.2699, 10.2998)(0.2738, 11.3422)(0.281, 16.4762)(0.2837, 19.563)(0.285, 21.9303)(0.286, 23.5618)(0.2862, 24.8776)(0.2865, 25.403)(0.2862, 26.2234)(0.2865, 26.5037)};
\addplot +[mark=*, solid, google_red, mark options={scale=0.4}] coordinates {(0.2479, 11.1512)(0.2599, 9.4931)(0.2672, 9.7014)(0.2721, 10.7983)(0.2756, 11.8326)(0.2822, 17.2267)(0.2847, 20.8375)(0.2861, 22.856)(0.2867, 24.6328)(0.2869, 25.723)(0.2869, 26.5768)(0.2868, 26.8506)(0.2873, 27.3912)};
\addplot +[mark=*, solid, google_green, mark options={scale=0.4}] coordinates {(0.2511, 10.4663)(0.2627, 9.4659)(0.27, 9.9582)(0.2748, 11.2437)(0.2781, 12.6562)(0.2836, 17.3972)(0.2863, 20.673)(0.2869, 22.4689)(0.2873, 23.7)(0.2873, 24.6056)(0.2876, 25.4052)(0.2876, 26.0697)(0.2876, 26.2802)};
\addplot +[mark=*, solid, google_yellow, mark options={scale=0.4}] coordinates {(0.2535, 10.1896)(0.2645, 9.3721)(0.2715, 9.9115)(0.2761, 11.2531)(0.28, 12.6531)(0.2844, 17.0384)(0.2866, 20.061)(0.2873, 21.9676)(0.2879, 23.0162)(0.2881, 23.936)(0.288, 24.1819)(0.288, 25)(0.2879, 25.5056)};

\legend{300M, 500M, 1B, 2B}
\end{axis}
\end{tikzpicture}
      \caption{Comparison between different model sizes.}
      \label{fig:model_size_comparison}
    \end{subfigure}
    \caption{CLIP vs FID-10K pareto curves for different ablation studies for the base $64\times64$ model. For each study, we sweep over guidance values of $[1, 1.25, 1.5, 1.75, 2, 3, 4, 5, 6, 7, 8, 9, 10]$}
\end{figure}
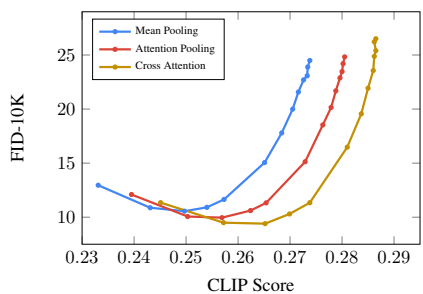
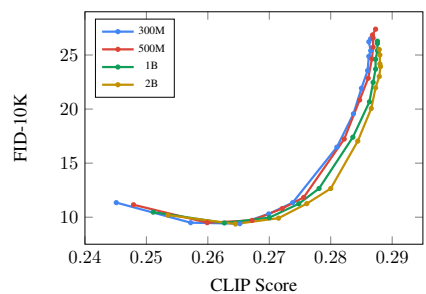

We ablate various schemas for conditioning the frozen text embeddings in the base $64\times64$ text-to-image diffusion model. Fig. \ref{fig:text_cond_comparison} compares the CLIP-FID pareto curves for mean pooling, attention pooling, and cross attention. We find using any pooled embedding configuration (mean or attention pooling) performs noticeably worse compared to attending over the sequence of contextual embeddings in the attention layers. We implement the cross attention by concatenating the text embedding sequence to the key-value pairs of each self-attention layer in the base $64\times64$ and $64\times64 \rightarrow 256\times256$ models. %
For our $256\times256 \rightarrow 1024\times1024$ model, since we have no self-attention layers, we simply added explicit cross-attention layers to attend over the text embeddings. We found this to improve both fidelity and image-text alignment with minimal computational costs.

\subsubsection{Comparison of U-Net vs Efficient U-Net}
\label{sec:unet_vs_efficientunet}
\begin{figure}
\centering
\begin{tikzpicture}[scale=0.75]

\definecolor{google_blue}{RGB}{66,133,244}
\definecolor{google_red}{RGB}{219,68,55}
\definecolor{google_yellow}{RGB}{194,144,0}
\definecolor{google_green}{RGB}{15,157,88}

\begin{axis}[width=10cm, height=5cm,
  xlabel=TPU Training Days,
  ylabel=FID-2K,
  legend columns=1,
  legend pos=north east,
  legend style={font=\tiny},
  xmin=0.0,
  ymin=18.0,
  ymax=42.0,
  xmax=4.5,
  every axis plot/.append style={thick}
]
\addplot +[mark=*, solid, google_blue, mark options={scale=0.4}] coordinates {(0.5, 40.4)(1.0, 28.4)(1.5, 25.3)(2.0, 24.2)(2.5, 23.0)(3.0, 21.5)(4.0, 21.2)};
\addplot +[mark=*, solid, google_red, mark options={scale=0.4}] coordinates {(0.5, 27.7)(1.0, 22.8)(1.5, 21.0)(2.0, 20.3)} ;
\legend{U-Net, Efficient U-Net}
\end{axis}
\end{tikzpicture}
\caption{Comparison of convergence speed of U-Net vs Efficient U-Net on the $64\times64 \rightarrow 256\times256$ super-resolution task.}
\label{fig:unet_vs_Efficient}
\end{figure}
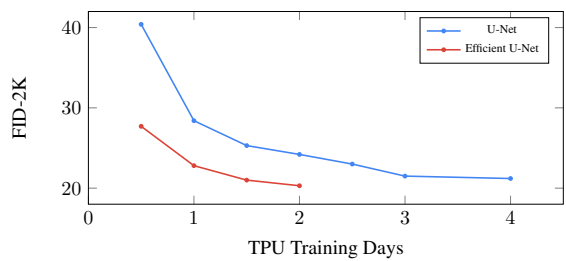

We compare the performance of U-Net with our new Efficient U-Net on the task of $64\times64 \rightarrow 256\times256$ super-resolution task. Fig. \ref{fig:unet_vs_Efficient} compares the training convergence of the two architectures. We observe that Efficient U-Net converges significantly faster than U-Net, and obtains better performance overall. Our Efficient U-Net is also $\times2-3$ faster at sampling.

\section{Comparison to GLIDE and DALL-E 2}
\label{sec:benchmark_comparison}

\begin{figure}[t]
\centering
\begin{tabular}{c@{\hskip 1pt}c@{\hskip 8pt}c@{\hskip 1pt}c}
\multicolumn{2}{c}{\name (Ours)} & \multicolumn{2}{c}{DALL-E 2 \cite{ramesh-dalle2}} \\
{\includegraphics[width=0.2\textwidth]{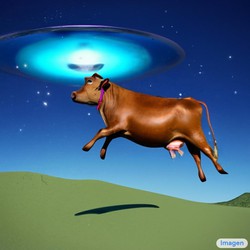}} &
{\includegraphics[width=0.2\textwidth]{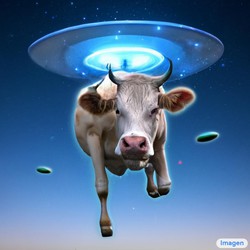}} &
{\includegraphics[width=0.2\textwidth]{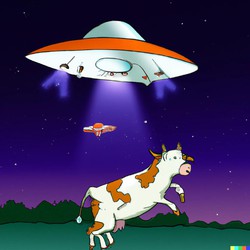}} &
{\includegraphics[width=0.2\textwidth]{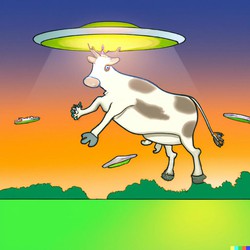} }\\
{\includegraphics[width=0.2\textwidth]{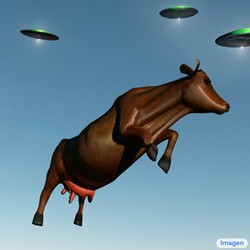}} &
{\includegraphics[width=0.2\textwidth]{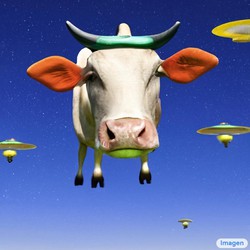}} &
{\includegraphics[width=0.2\textwidth]{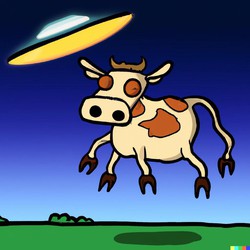}} &
{\includegraphics[width=0.2\textwidth]{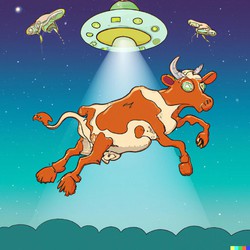}}\\
\multicolumn{4}{c}{Hovering cow abducting aliens.} \\
\multicolumn{4}{c}{} \\
{\includegraphics[width=0.2\textwidth]{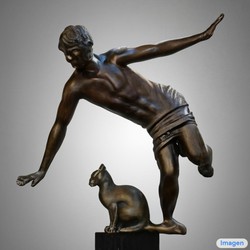}} &
{\includegraphics[width=0.2\textwidth]{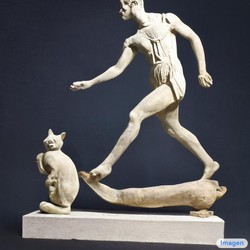}} &
{\includegraphics[width=0.2\textwidth]{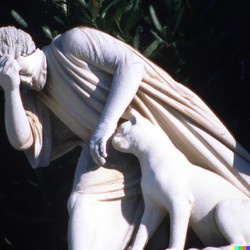}} &
{\includegraphics[width=0.2\textwidth]{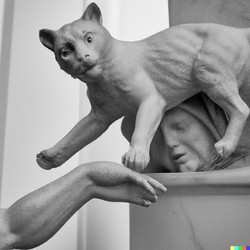} }\\
{\includegraphics[width=0.2\textwidth]{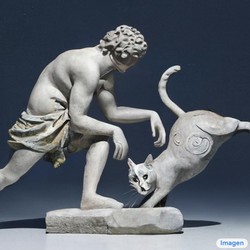}} &
{\includegraphics[width=0.2\textwidth]{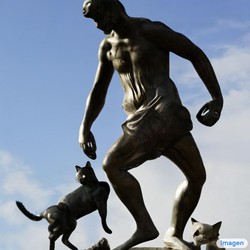}} &
{\includegraphics[width=0.2\textwidth]{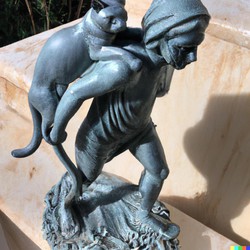}} &
{\includegraphics[width=0.2\textwidth]{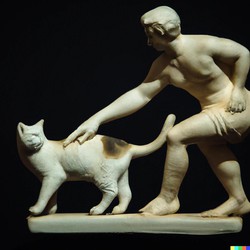}}\\
\multicolumn{4}{c}{Greek statue of a man tripping over a cat.}
\end{tabular}

\caption{Example qualitative comparisons between \name and DALL-E 2 \cite{ramesh-dalle2} on \benchmarkname prompts from Reddit category.}
\label{fig:drawit_vs_dalle2_reddit}
\end{figure}

\begin{figure}[t]
\centering
\begin{tabular}{c@{\hskip 1pt}c@{\hskip 8pt}c@{\hskip 1pt}c}
\multicolumn{2}{c}{\name (Ours)} & \multicolumn{2}{c}{DALL-E 2 \cite{ramesh-dalle2}} \\
{\includegraphics[width=0.2\textwidth]{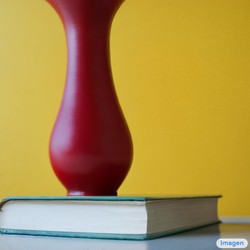}} &
{\includegraphics[width=0.2\textwidth]{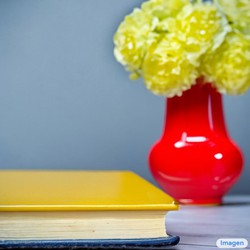}} &
{\includegraphics[width=0.2\textwidth]{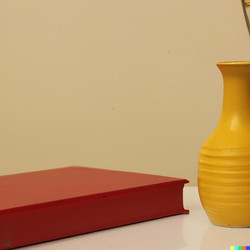}} &
{\includegraphics[width=0.2\textwidth]{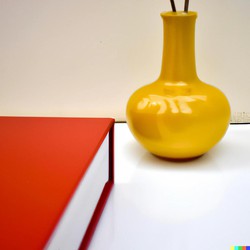} }\\
{\includegraphics[width=0.2\textwidth]{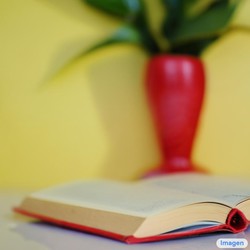}} &
{\includegraphics[width=0.2\textwidth]{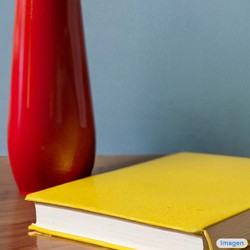}} &
{\includegraphics[width=0.2\textwidth]{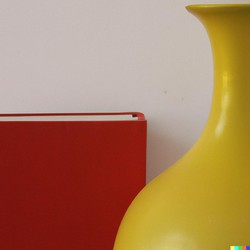}} &
{\includegraphics[width=0.2\textwidth]{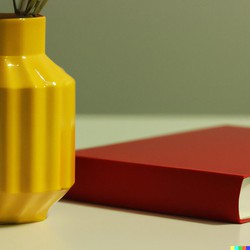}}\\
\multicolumn{4}{c}{A yellow book and a red vase.} \\
\multicolumn{4}{c}{} \\
{\includegraphics[width=0.2\textwidth]{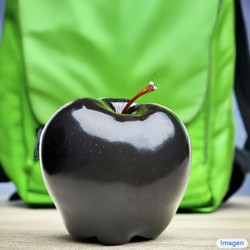}} &
{\includegraphics[width=0.2\textwidth]{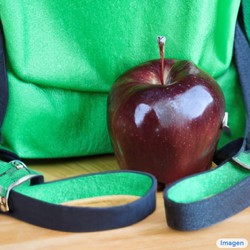}} &
{\includegraphics[width=0.2\textwidth]{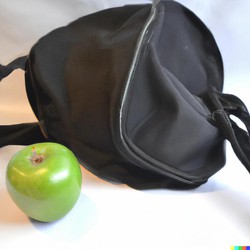}} &
{\includegraphics[width=0.2\textwidth]{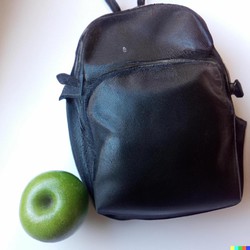} }\\
{\includegraphics[width=0.2\textwidth]{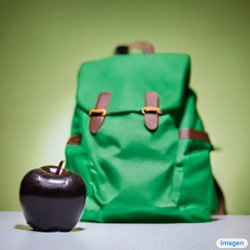}} &
{\includegraphics[width=0.2\textwidth]{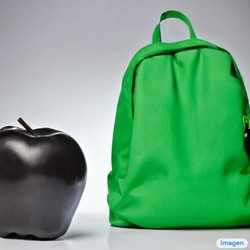}} &
{\includegraphics[width=0.2\textwidth]{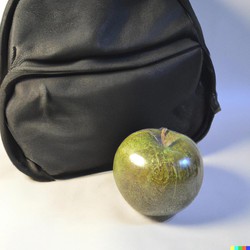}} &
{\includegraphics[width=0.2\textwidth]{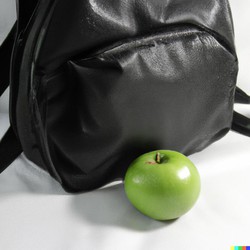}}\\
\multicolumn{4}{c}{A black apple and a green backpack.}
\end{tabular}

\caption{Example qualitative comparisons between \name and DALL-E 2 \cite{ramesh-dalle2} on \benchmarkname prompts from Colors category. We observe that DALL-E 2 generally struggles with correctly assigning the colors to the objects especially for prompts with more than one object.}
\label{fig:drawit_vs_dalle2_colors}
\end{figure}

\begin{figure}[t]
\centering
\begin{tabular}{c@{\hskip 1pt}c@{\hskip 8pt}c@{\hskip 1pt}c}
\multicolumn{2}{c}{\name (Ours)} & \multicolumn{2}{c}{DALL-E 2 \cite{ramesh-dalle2}} \\
{\includegraphics[width=0.2\textwidth]{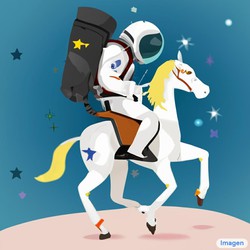}} &
{\includegraphics[width=0.2\textwidth]{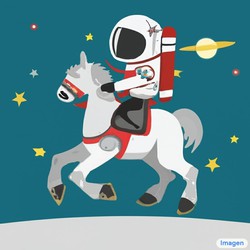}} &
{\includegraphics[width=0.2\textwidth]{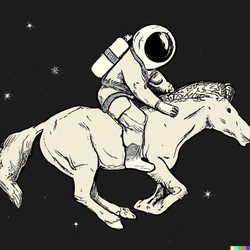}} &
{\includegraphics[width=0.2\textwidth]{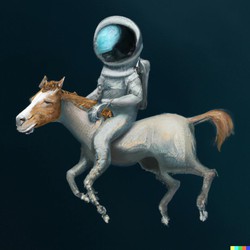} }\\
{\includegraphics[width=0.2\textwidth]{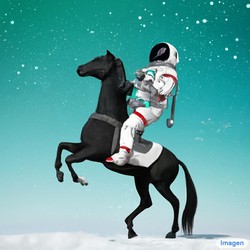}} &
{\includegraphics[width=0.2\textwidth]{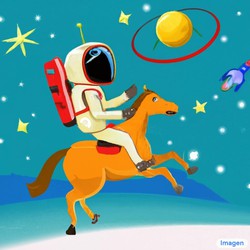}} &
{\includegraphics[width=0.2\textwidth]{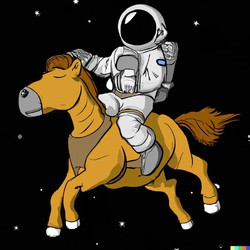}} &
{\includegraphics[width=0.2\textwidth]{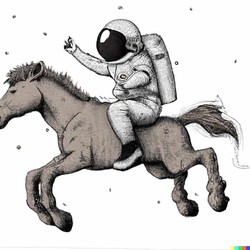}}\\
\multicolumn{4}{c}{A horse riding an astronaut.} \\
\multicolumn{4}{c}{} \\
{\includegraphics[width=0.2\textwidth]{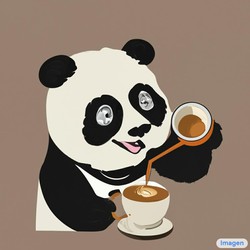}} &
{\includegraphics[width=0.2\textwidth]{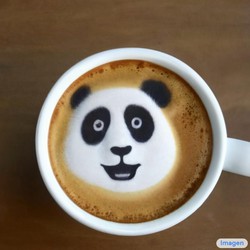}} &
{\includegraphics[width=0.2\textwidth]{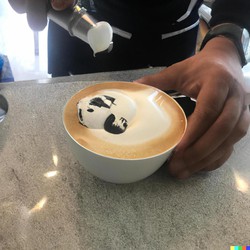}} &
{\includegraphics[width=0.2\textwidth]{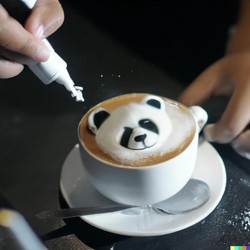} }\\
{\includegraphics[width=0.2\textwidth]{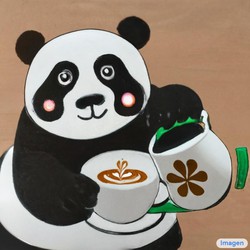}} &
{\includegraphics[width=0.2\textwidth]{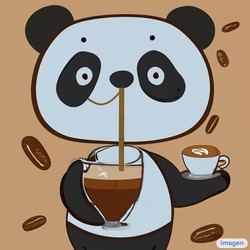}} &
{\includegraphics[width=0.2\textwidth]{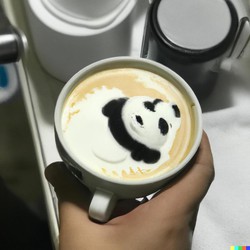}} &
{\includegraphics[width=0.2\textwidth]{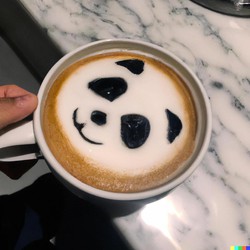}}\\
\multicolumn{4}{c}{A panda making latte art.}
\end{tabular}

\caption{Example qualitative comparisons between \name and DALL-E 2 \cite{ramesh-dalle2} on \benchmarkname prompts from Conflicting category. We observe that both DALL-E 2 and \name struggle generating well aligned images for this category. However, \name often generates some well aligned samples, e.g. ``A panda making latte art.''.}
\label{fig:drawit_vs_dalle2_conflicting}
\end{figure}

\begin{figure}[t]
\centering
\begin{tabular}{c@{\hskip 1pt}c@{\hskip 8pt}c@{\hskip 1pt}c}
\multicolumn{2}{c}{\name (Ours)} & \multicolumn{2}{c}{DALL-E 2 \cite{ramesh-dalle2}} \\
{\includegraphics[width=0.2\textwidth]{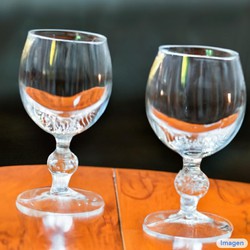}} &
{\includegraphics[width=0.2\textwidth]{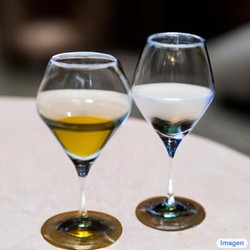}} &
{\includegraphics[width=0.2\textwidth]{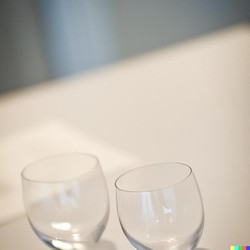}} &
{\includegraphics[width=0.2\textwidth]{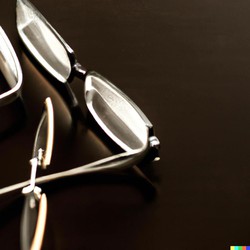} }\\
{\includegraphics[width=0.2\textwidth]{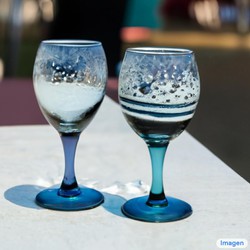}} &
{\includegraphics[width=0.2\textwidth]{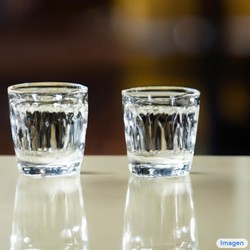}} &
{\includegraphics[width=0.2\textwidth]{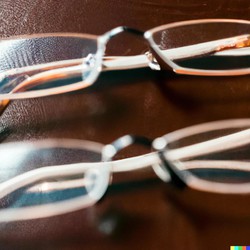}} &
{\includegraphics[width=0.2\textwidth]{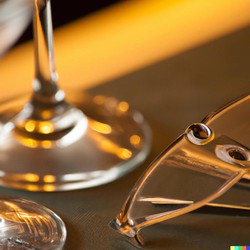}}\\
\multicolumn{4}{c}{A couple of glasses are sitting on a table.} \\
\multicolumn{4}{c}{} \\
{\includegraphics[width=0.2\textwidth]{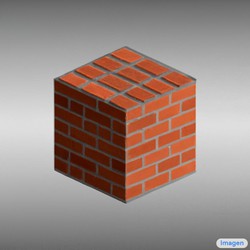}} &
{\includegraphics[width=0.2\textwidth]{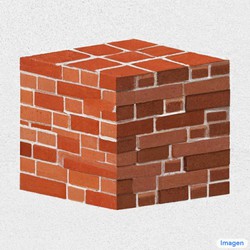}} &
{\includegraphics[width=0.2\textwidth]{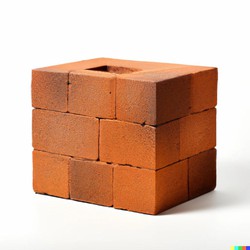}} &
{\includegraphics[width=0.2\textwidth]{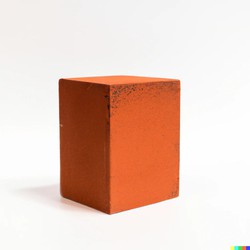} }\\
{\includegraphics[width=0.2\textwidth]{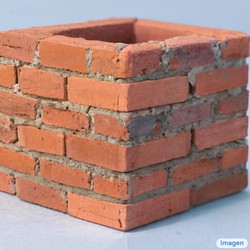}} &
{\includegraphics[width=0.2\textwidth]{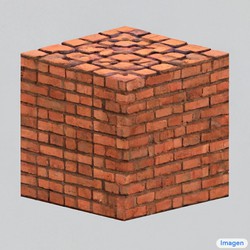}} &
{\includegraphics[width=0.2\textwidth]{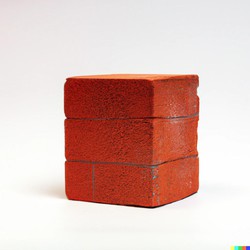}} &
{\includegraphics[width=0.2\textwidth]{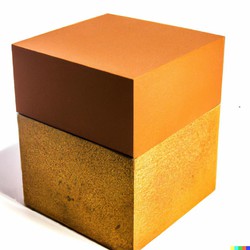}}\\
\multicolumn{4}{c}{A cube made of brick. A cube with the texture of brick.}
\end{tabular}

\caption{Example qualitative comparisons between \name and DALL-E 2 \cite{ramesh-dalle2} on \benchmarkname prompts from DALL-E category.}
\label{fig:drawit_vs_dalle2_dalle}
\end{figure}

\begin{figure}[t]
\centering
\begin{tabular}{c@{\hskip 1pt}c@{\hskip 8pt}c@{\hskip 1pt}c}
\multicolumn{2}{c}{\name (Ours)} & \multicolumn{2}{c}{DALL-E 2 \cite{ramesh-dalle2}} \\
{\includegraphics[width=0.2\textwidth]{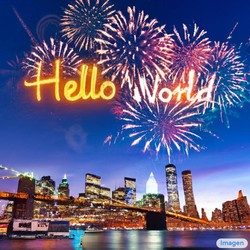}} &
{\includegraphics[width=0.2\textwidth]{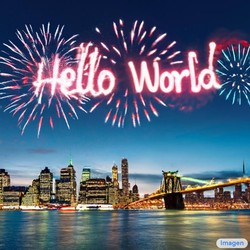}} &
{\includegraphics[width=0.2\textwidth]{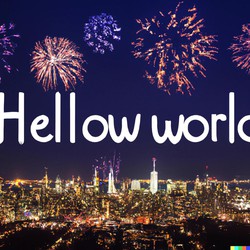}} &
{\includegraphics[width=0.2\textwidth]{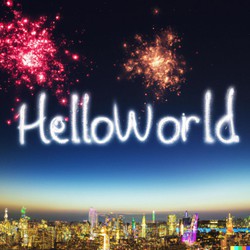} }\\
{\includegraphics[width=0.2\textwidth]{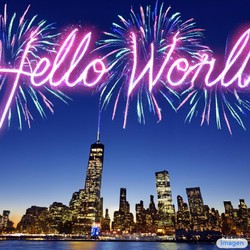}} &
{\includegraphics[width=0.2\textwidth]{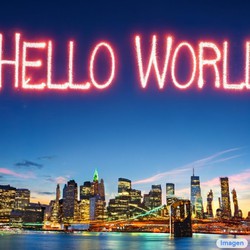}} &
{\includegraphics[width=0.2\textwidth]{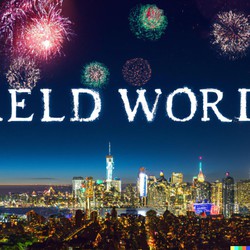}} &
{\includegraphics[width=0.2\textwidth]{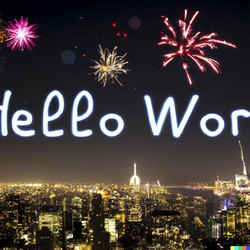}}\\
\multicolumn{4}{c}{New York Skyline with Hello World written with fireworks on the sky.} \\
\multicolumn{4}{c}{} \\
{\includegraphics[width=0.2\textwidth]{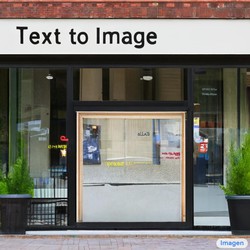}} &
{\includegraphics[width=0.2\textwidth]{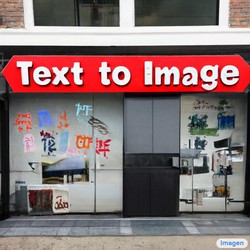}} &
{\includegraphics[width=0.2\textwidth]{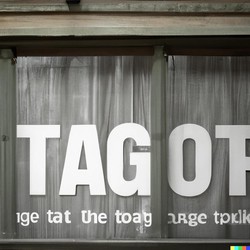}} &
{\includegraphics[width=0.2\textwidth]{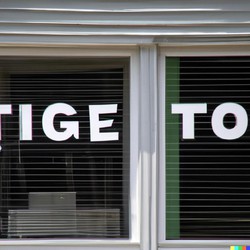} }\\
{\includegraphics[width=0.2\textwidth]{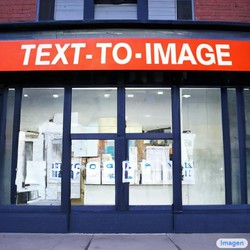}} &
{\includegraphics[width=0.2\textwidth]{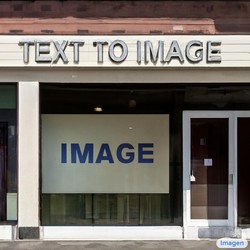}} &
{\includegraphics[width=0.2\textwidth]{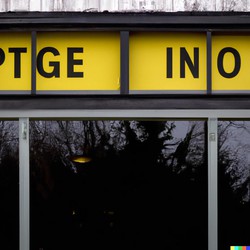}} &
{\includegraphics[width=0.2\textwidth]{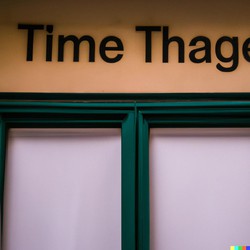}}\\
\multicolumn{4}{c}{A storefront with Text to Image written on it.}
\end{tabular}

\caption{Example qualitative comparisons between \name and DALL-E 2 \cite{ramesh-dalle2} on \benchmarkname prompts from Text category. \name is significantly better than DALL-E 2 in prompts with quoted text.}
\label{fig:drawit_vs_dalle2_text}
\end{figure}

\begin{figure}[htb]
\centering
\begin{tabular}{c@{\hskip 1pt}c@{\hskip 8pt}c@{\hskip 1pt}c}
\multicolumn{2}{c}{\name (Ours)} & \multicolumn{2}{c}{GLIDE \cite{nichol-glide}} \\
{\includegraphics[width=0.2\textwidth]{figures/drawit_vs_dalle2/cow_alien/drawit_1.jpeg}} &
{\includegraphics[width=0.2\textwidth]{figures/drawit_vs_dalle2/cow_alien/drawit_2.jpeg}} &
{\includegraphics[width=0.2\textwidth]{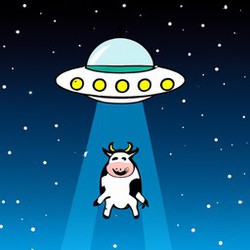}} &
{\includegraphics[width=0.2\textwidth]{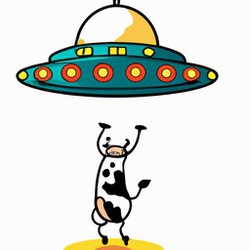}} \\
{\includegraphics[width=0.2\textwidth]{figures/drawit_vs_dalle2/cow_alien/drawit_3.jpeg}} &
{\includegraphics[width=0.2\textwidth]{figures/drawit_vs_dalle2/cow_alien/drawit_4.jpeg}} &
{\includegraphics[width=0.2\textwidth]{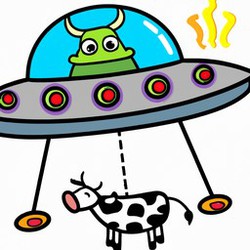}} &
{\includegraphics[width=0.2\textwidth]{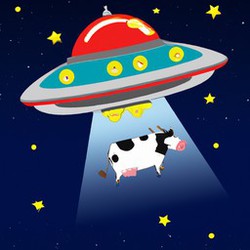}} \\
\multicolumn{4}{c}{Hovering cow abducting aliens.} \\
\multicolumn{4}{c}{} \\
{\includegraphics[width=0.2\textwidth]{figures/drawit_vs_dalle2/greek_statue/drawit_1.jpeg}} &
{\includegraphics[width=0.2\textwidth]{figures/drawit_vs_dalle2/greek_statue/drawit_2.jpeg}} &
{\includegraphics[width=0.2\textwidth]{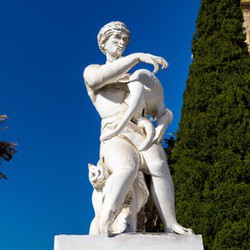}} &
{\includegraphics[width=0.2\textwidth]{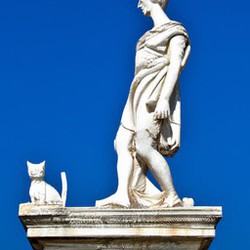}} \\
{\includegraphics[width=0.2\textwidth]{figures/drawit_vs_dalle2/greek_statue/drawit_3.jpeg}} &
{\includegraphics[width=0.2\textwidth]{figures/drawit_vs_dalle2/greek_statue/drawit_4.jpeg}} &
{\includegraphics[width=0.2\textwidth]{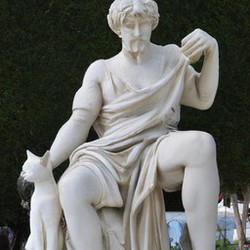}} &
{\includegraphics[width=0.2\textwidth]{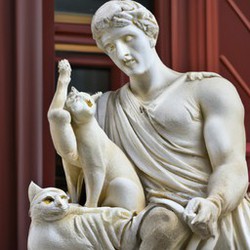}} \\
\multicolumn{4}{c}{Greek statue of a man tripping over a cat.}
\end{tabular}
\caption{Example qualitative comparisons between \name and GLIDE \cite{nichol-glide} on \benchmarkname prompts from Reddit category.}
\label{fig:drawit_vs_glide_reddit}
\end{figure}

\begin{figure}[t]
\centering
\begin{tabular}{c@{\hskip 1pt}c@{\hskip 8pt}c@{\hskip 1pt}c}
\multicolumn{2}{c}{\name (Ours)} & \multicolumn{2}{c}{GLIDE \cite{nichol-glide}} \\
{\includegraphics[width=0.2\textwidth]{figures/drawit_vs_dalle2/yellowbook_redvase/drawit_1.jpeg}} &
{\includegraphics[width=0.2\textwidth]{figures/drawit_vs_dalle2/yellowbook_redvase/drawit_2.jpeg}} &
{\includegraphics[width=0.2\textwidth]{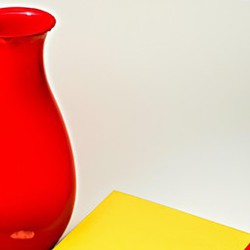}} &
{\includegraphics[width=0.2\textwidth]{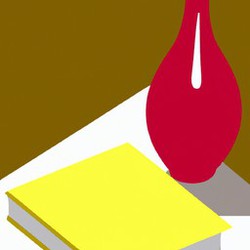}} \\
{\includegraphics[width=0.2\textwidth]{figures/drawit_vs_dalle2/yellowbook_redvase/drawit_3.jpeg}} &
{\includegraphics[width=0.2\textwidth]{figures/drawit_vs_dalle2/yellowbook_redvase/drawit_4.jpeg}} &
{\includegraphics[width=0.2\textwidth]{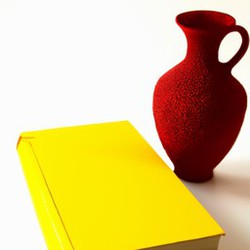}} &
{\includegraphics[width=0.2\textwidth]{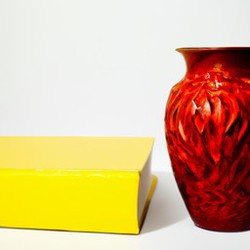}} \\
\multicolumn{4}{c}{A yellow book and a red vase.} \\
\multicolumn{4}{c}{} \\
{\includegraphics[width=0.2\textwidth]{figures/drawit_vs_dalle2/blackapple_greenbag/drawit_1.jpeg}} &
{\includegraphics[width=0.2\textwidth]{figures/drawit_vs_dalle2/blackapple_greenbag/drawit_2.jpeg}} &
{\includegraphics[width=0.2\textwidth]{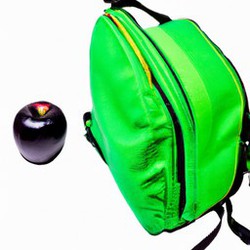}} &
{\includegraphics[width=0.2\textwidth]{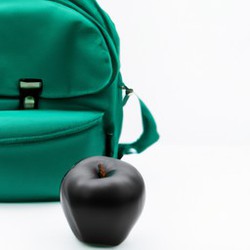}} \\
{\includegraphics[width=0.2\textwidth]{figures/drawit_vs_dalle2/blackapple_greenbag/drawit_3.jpeg}} &
{\includegraphics[width=0.2\textwidth]{figures/drawit_vs_dalle2/blackapple_greenbag/drawit_4.jpeg}} &
{\includegraphics[width=0.2\textwidth]{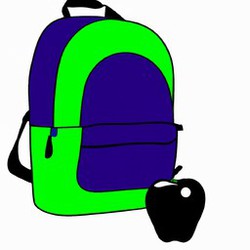}} &
{\includegraphics[width=0.2\textwidth]{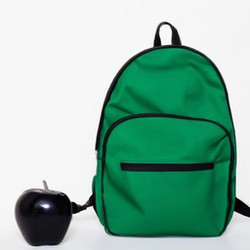}} \\
\multicolumn{4}{c}{A black apple and a green backpack.}
\end{tabular}

\caption{Example qualitative comparisons between \name and GLIDE \cite{nichol-glide} on \benchmarkname prompts from Colors category. We observe that GLIDE is better than DALL-E 2 in assigning the colors to the objects.}
\label{fig:drawit_vs_glide_colors}
\end{figure}

\begin{figure}[t]
\centering
\begin{tabular}{c@{\hskip 1pt}c@{\hskip 8pt}c@{\hskip 1pt}c}
\multicolumn{2}{c}{\name (Ours)} & \multicolumn{2}{c}{GLIDE \cite{nichol-glide}} \\
{\includegraphics[width=0.2\textwidth]{figures/drawit_vs_dalle2/horse_riding_astronaut/drawit_1.jpeg}} &
{\includegraphics[width=0.2\textwidth]{figures/drawit_vs_dalle2/horse_riding_astronaut/drawit_2.jpeg}} &
{\includegraphics[width=0.2\textwidth]{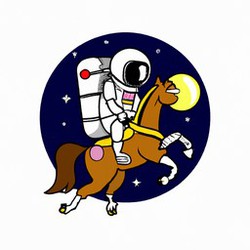}} &
{\includegraphics[width=0.2\textwidth]{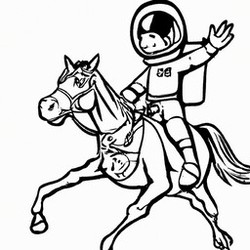}} \\
{\includegraphics[width=0.2\textwidth]{figures/drawit_vs_dalle2/horse_riding_astronaut/drawit_3.jpeg}} &
{\includegraphics[width=0.2\textwidth]{figures/drawit_vs_dalle2/horse_riding_astronaut/drawit_4.jpeg}} &
{\includegraphics[width=0.2\textwidth]{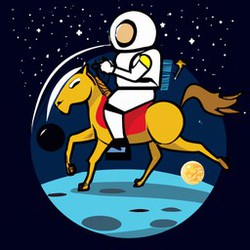}} &
{\includegraphics[width=0.2\textwidth]{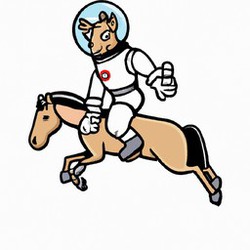}} \\
\multicolumn{4}{c}{A horse riding an astronaut.} \\
\multicolumn{4}{c}{} \\
{\includegraphics[width=0.2\textwidth]{figures/drawit_vs_dalle2/panda_latte_art/drawit_1.jpeg}} &
{\includegraphics[width=0.2\textwidth]{figures/drawit_vs_dalle2/panda_latte_art/drawit_2.jpeg}} &
{\includegraphics[width=0.2\textwidth]{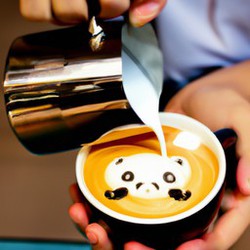}} &
{\includegraphics[width=0.2\textwidth]{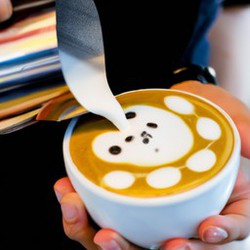}} \\
{\includegraphics[width=0.2\textwidth]{figures/drawit_vs_dalle2/panda_latte_art/drawit_3.jpeg}} &
{\includegraphics[width=0.2\textwidth]{figures/drawit_vs_dalle2/panda_latte_art/drawit_4.jpeg}} &
{\includegraphics[width=0.2\textwidth]{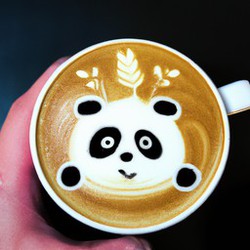}} &
{\includegraphics[width=0.2\textwidth]{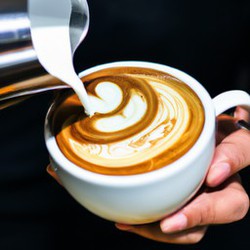}} \\
\multicolumn{4}{c}{A panda making latte art.}
\end{tabular}

\caption{Example qualitative comparisons between \name and GLIDE \cite{nichol-glide} on \benchmarkname prompts from Conflicting category.}
\label{fig:drawit_vs_glide_conflicting}
\end{figure}

\begin{figure}[t]
\centering
\begin{tabular}{c@{\hskip 1pt}c@{\hskip 8pt}c@{\hskip 1pt}c}
\multicolumn{2}{c}{\name (Ours)} & \multicolumn{2}{c}{GLIDE \cite{nichol-glide}} \\
{\includegraphics[width=0.2\textwidth]{figures/drawit_vs_dalle2/glasses_on_table/drawit_1.jpeg}} &
{\includegraphics[width=0.2\textwidth]{figures/drawit_vs_dalle2/glasses_on_table/drawit_2.jpeg}} &
{\includegraphics[width=0.2\textwidth]{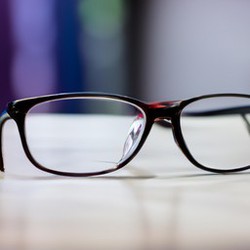}} &
{\includegraphics[width=0.2\textwidth]{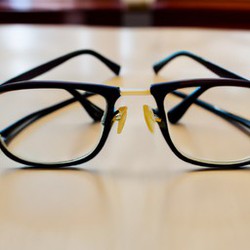}} \\
{\includegraphics[width=0.2\textwidth]{figures/drawit_vs_dalle2/glasses_on_table/drawit_3.jpeg}} &
{\includegraphics[width=0.2\textwidth]{figures/drawit_vs_dalle2/glasses_on_table/drawit_4.jpeg}} &
{\includegraphics[width=0.2\textwidth]{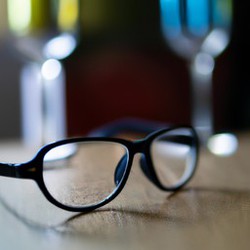}} &
{\includegraphics[width=0.2\textwidth]{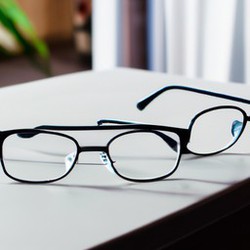}} \\
\multicolumn{4}{c}{A couple of glasses are sitting on a table.} \\
\multicolumn{4}{c}{} \\
{\includegraphics[width=0.2\textwidth]{figures/drawit_vs_dalle2/cube_brick/drawit_1.jpeg}} &
{\includegraphics[width=0.2\textwidth]{figures/drawit_vs_dalle2/cube_brick/drawit_2.jpeg}} &
{\includegraphics[width=0.2\textwidth]{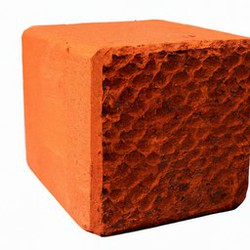}} &
{\includegraphics[width=0.2\textwidth]{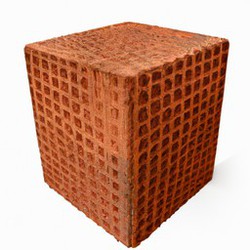}} \\
{\includegraphics[width=0.2\textwidth]{figures/drawit_vs_dalle2/cube_brick/drawit_3.jpeg}} &
{\includegraphics[width=0.2\textwidth]{figures/drawit_vs_dalle2/cube_brick/drawit_4.jpeg}} &
{\includegraphics[width=0.2\textwidth]{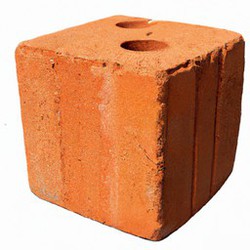}} &
{\includegraphics[width=0.2\textwidth]{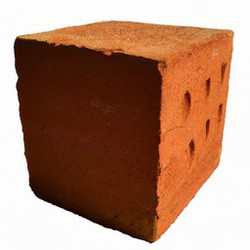}} \\
\\
\multicolumn{4}{c}{A cube made of brick. A cube with the texture of brick.}
\end{tabular}

\caption{Example qualitative comparisons between \name and GLIDE \cite{nichol-glide} on \benchmarkname prompts from DALL-E category.}
\label{fig:drawit_vs_glide_dalle}
\end{figure}

\begin{figure}[t]
\centering
\begin{tabular}{c@{\hskip 1pt}c@{\hskip 8pt}c@{\hskip 1pt}c}
\multicolumn{2}{c}{\name (Ours)} & \multicolumn{2}{c}{GLIDE \cite{nichol-glide}} \\
{\includegraphics[width=0.2\textwidth]{figures/drawit_vs_dalle2/hello_world/drawit_1.jpeg}} &
{\includegraphics[width=0.2\textwidth]{figures/drawit_vs_dalle2/hello_world/drawit_2.jpeg}} &
{\includegraphics[width=0.2\textwidth]{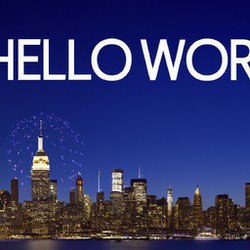}} &
{\includegraphics[width=0.2\textwidth]{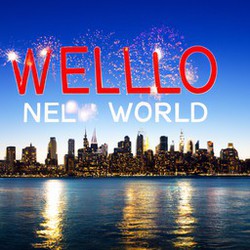}} \\
{\includegraphics[width=0.2\textwidth]{figures/drawit_vs_dalle2/hello_world/drawit_3.jpeg}} &
{\includegraphics[width=0.2\textwidth]{figures/drawit_vs_dalle2/hello_world/drawit_4.jpeg}} &
{\includegraphics[width=0.2\textwidth]{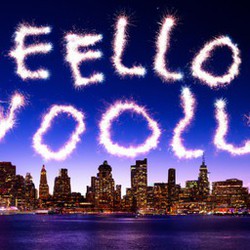}} &
{\includegraphics[width=0.2\textwidth]{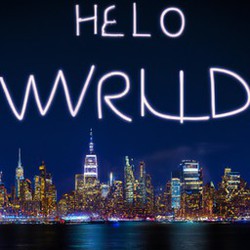}} \\
\multicolumn{4}{c}{New York Skyline with Hello World written with fireworks on the sky.} \\
\multicolumn{4}{c}{} \\
{\includegraphics[width=0.2\textwidth]{figures/drawit_vs_dalle2/text_to_image/drawit_1.jpeg}} &
{\includegraphics[width=0.2\textwidth]{figures/drawit_vs_dalle2/text_to_image/drawit_2.jpeg}} &
{\includegraphics[width=0.2\textwidth]{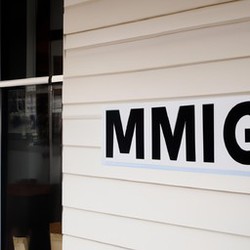}} &
{\includegraphics[width=0.2\textwidth]{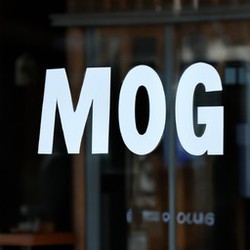}} \\
{\includegraphics[width=0.2\textwidth]{figures/drawit_vs_dalle2/text_to_image/drawit_3.jpeg}} &
{\includegraphics[width=0.2\textwidth]{figures/drawit_vs_dalle2/text_to_image/drawit_4.jpeg}} &
{\includegraphics[width=0.2\textwidth]{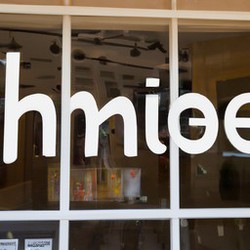}} &
{\includegraphics[width=0.2\textwidth]{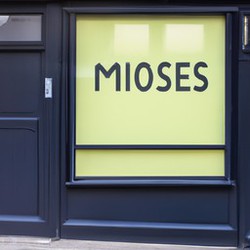}} \\
\multicolumn{4}{c}{A storefront with Text to Image written on it.}
\end{tabular}

\caption{Example qualitative comparisons between \name and GLIDE \cite{nichol-glide} on \benchmarkname prompts from Text category. \name is significantly better than GLIDE too in prompts with quoted text.}
\label{fig:drawit_vs_glide_text}
\end{figure}

\cref{fig:drawit_vs_dalle2_detailed_alignment} shows category wise comparison between \name and DALL-E 2 \cite{ramesh-dalle2} on \benchmarkname. We observe that human raters clearly prefer \name over DALL-E 2 in 7 out of 11 categories for text alignment. For sample fidelity, they prefer \name over DALL-E 2 in all 11 categories. \Cref{fig:drawit_vs_dalle2_reddit,fig:drawit_vs_dalle2_colors,fig:drawit_vs_dalle2_conflicting,fig:drawit_vs_dalle2_dalle,fig:drawit_vs_dalle2_text} show few qualitative comparisons between \name and DALL-E 2 samples used for this human evaluation study. Some of the categories where \name has a considerably larger preference over DALL-E 2 include Colors, Positional, Text, DALL-E and Descriptions. The authors in \cite{ramesh-dalle2} identify some of these limitations of DALL-E~2, specifically they observe that DALLE-E~2 is worse than GLIDE \cite{nichol-glide} in binding attributes to objects such as colors, and producing coherent text from the input prompt (cf.\ the discussion of limitations in \cite{ramesh-dalle2}). To this end, we also perform quantitative and qualitative comparison with GLIDE \cite{nichol-glide} on \benchmarkname. See \cref{fig:drawit_vs_glide_detailed_alignment} for category wise human evaluation comparison between \name and GLIDE. See \Cref{fig:drawit_vs_glide_reddit,fig:drawit_vs_glide_colors,fig:drawit_vs_glide_conflicting,fig:drawit_vs_glide_dalle,fig:drawit_vs_glide_text} for qualitative comparisons. \name outperforms GLIDE on 8 out of 11 categories on image-text alignment, and 10 out of 11 categories on image fidelity. We observe that GLIDE is considerably better than DALL-E~2 in binding attributes to objects corroborating the observation by  \citep{ramesh-dalle2}.

\begin{figure}
    \centering
    \begin{subfigure}[t]{\textwidth}
    \begin{tikzpicture}

\definecolor{google_blue}{RGB}{66,133,244}
\definecolor{google_red}{RGB}{219,68,55}
\definecolor{google_yellow}{RGB}{194,144,0}
\definecolor{google_green}{RGB}{15,157,88}

\begin{axis}[
name=fidelity,
width=\textwidth,
height=0.5\textwidth,
ybar stacked,
ymin=-0.1,
ymax=1.1,
xtick=data,
xtick style={draw=none},
symbolic x coords={Colors,Conflicting,Counting,DALL-E \cite{ramesh-dalle},Descriptions,Marcus et al. \cite{marcus-arxiv-2022},Misspellings,Positional,Rare Words,Reddit,Text},
xticklabels={},
x tick label style={rotate=30},
ytick={0,0.5,1},
yticklabels={0\%, 50\%, 100\%},
legend style={font=\tiny},
legend cell align=left,
legend image code/.code={
    \draw [#1] (0cm, -0.1cm) rectangle (0.2cm, 0.25cm);
},
legend style={
draw=none,
at={(0.5, 1.1)},anchor=south,
legend columns=3,
/tikz/every even column/.append style={column sep=0.2cm}
},
bar width=0.6cm,
]

\addplot+[
    fill=google_blue,
    fill opacity=0.2,
    draw=google_blue
] coordinates {
(Colors, 0.736)
(Conflicting, 0.562)
(Counting, 0.672)
(DALL-E \cite{ramesh-dalle}, 0.627)
(Descriptions, 0.746)
(Marcus et al. \cite{marcus-arxiv-2022}, 0.444)
(Misspellings, 0.369)
(Positional, 0.691)
(Rare Words, 0.344)
(Reddit, 0.499)
(Text, 0.972)
};

\addplot+[
fill=google_green,
fill opacity=0.2,
draw=google_green
] coordinates {
(Colors, 0.264)
(Conflicting, 0.438)
(Counting, 0.328)
(DALL-E \cite{ramesh-dalle}, 0.372)
(Descriptions, 0.254)
(Marcus et al. \cite{marcus-arxiv-2022}, 0.556)
(Misspellings, 0.631)
(Positional, 0.309)
(Rare Words, 0.655)
(Reddit, 0.501)
(Text, 0.028)
};

\legend{\name, DALL-E 2}
\end{axis}

\end{tikzpicture}
    \caption{Alignment}
    \label{fig:drawit_vs_dalle2_detailed_alignment_alignment}
    \vspace*{0.5cm}
    \end{subfigure}
    \begin{subfigure}[t]{\textwidth}
    \begin{tikzpicture}

\definecolor{google_blue}{RGB}{66,133,244}
\definecolor{google_red}{RGB}{219,68,55}
\definecolor{google_yellow}{RGB}{194,144,0}
\definecolor{google_green}{RGB}{15,157,88}

\begin{axis}[
name=fidelity,
width=\textwidth,
height=0.5\textwidth,
ybar stacked,
ymin=-0.1,
ymax=1.1,
xtick=data,
xtick style={draw=none},
symbolic x coords={Color,Conflicting,Count,DALL-E \cite{ramesh-dalle},Descriptions,Marcus et al. \cite{marcus-arxiv-2022},Misspellings,Positional,Rarewords,Reddit,Text},
x tick label style={rotate=60},
ytick={0,0.5,1},
yticklabels={0\%, 50\%, 100\%},
legend style={font=\tiny},
legend cell align=left,
legend image code/.code={
    \draw [#1] (0cm, -0.1cm) rectangle (0.2cm, 0.25cm);
},
legend style={
draw=none,
at={(0.5, 1.1)},anchor=south,
legend columns=3,
/tikz/every even column/.append style={column sep=0.2cm}
},
bar width=0.6cm,
]

\addplot+[
    fill=google_blue,
    fill opacity=0.2,
    draw=google_blue
] coordinates {
(Color, 0.685)
(Conflicting, 0.619)
(Count, 0.594)
(DALL-E \cite{ramesh-dalle}, 0.704)
(Descriptions, 0.678)
(Marcus et al. \cite{marcus-arxiv-2022}, 0.705)
(Misspellings, 0.540)
(Positional, 0.721)
(Rarewords, 0.621)
(Reddit, 0.623)
(Text, 0.771)
};

\addplot+[
fill=google_green,
fill opacity=0.2,
draw=google_green
] coordinates {
(Color, 0.315)
(Conflicting, 0.381)
(Count, 0.252)
(DALL-E \cite{ramesh-dalle}, 0.296)
(Descriptions, 0.322)
(Marcus et al. \cite{marcus-arxiv-2022}, 0.295)
(Misspellings, 0.460)
(Positional, 0.279)
(Rarewords, 0.379)
(Reddit, 0.377)
(Text, 0.229)
};

\legend{}
\end{axis}

\end{tikzpicture}
    \caption{Fidelity}
    \label{fig:drawit_vs_dalle2_detailed_alignment_fidelity}
    \end{subfigure}
    \caption{\name vs DALL-E 2 on \benchmarkname \subref{fig:drawit_vs_dalle2_detailed_alignment_alignment}) image-text alignment, and \subref{fig:drawit_vs_dalle2_detailed_alignment_fidelity}) image fidelity.}
    \label{fig:drawit_vs_dalle2_detailed_alignment}
\end{figure}

\begin{figure}
    \centering
    \begin{subfigure}[t]{\textwidth}
    \begin{tikzpicture}

\definecolor{google_blue}{RGB}{66,133,244}
\definecolor{google_red}{RGB}{219,68,55}
\definecolor{google_yellow}{RGB}{194,144,0}
\definecolor{google_green}{RGB}{15,157,88}

\begin{axis}[
name=fidelity,
width=\textwidth,
height=0.5\textwidth,
ybar stacked,
ymin=-0.1,
ymax=1.1,
xtick=data,
xtick style={draw=none},
symbolic x coords={Colors,Conflicting,Counting,DALL-E \cite{ramesh-dalle},Descriptions,Marcus et al. \cite{marcus-arxiv-2022},Misspellings,Positional,Rare Words,Reddit,Text},
xticklabels={},
x tick label style={rotate=30},
ytick={0,0.5,1},
yticklabels={0\%, 50\%, 100\%},
legend style={font=\tiny},
legend cell align=left,
legend image code/.code={
    \draw [#1] (0cm, -0.1cm) rectangle (0.2cm, 0.25cm);
},
legend style={
draw=none,
at={(0.5, 1.1)},anchor=south,
legend columns=3,
/tikz/every even column/.append style={column sep=0.2cm}
},
bar width=0.6cm,
]

\addplot+[
    fill=google_blue,
    fill opacity=0.2,
    draw=google_blue
] coordinates {
(Colors, 0.429)
(Conflicting, 0.628)
(Counting, 0.609)
(DALL-E \cite{ramesh-dalle}, 0.493)
(Descriptions, 0.730)
(Marcus et al. \cite{marcus-arxiv-2022}, 0.568)
(Misspellings, 0.506)
(Positional, 0.659)
(Rare Words, 0.500)
(Reddit, 0.613)
(Text, 0.837)
};

\addplot+[
fill=google_green,
fill opacity=0.2,
draw=google_green
] coordinates {
(Colors, 0.571)
(Conflicting, 0.372)
(Counting, 0.391)
(DALL-E \cite{ramesh-dalle}, 0.507)
(Descriptions, 0.270)
(Marcus et al. \cite{marcus-arxiv-2022}, 0.432)
(Misspellings, 0.494)
(Positional, 0.341)
(Rare Words, 0.500)
(Reddit, 0.387)
(Text, 0.163)
};

\legend{\name, GLIDE}
\end{axis}

\end{tikzpicture}
    \caption{Alignment}
    \label{fig:drawit_vs_glide_detailed_alignment_alignment}
    \vspace*{0.5cm}
    \end{subfigure}
    \begin{subfigure}[t]{\textwidth}
    \begin{tikzpicture}

\definecolor{google_blue}{RGB}{66,133,244}
\definecolor{google_red}{RGB}{219,68,55}
\definecolor{google_yellow}{RGB}{194,144,0}
\definecolor{google_green}{RGB}{15,157,88}

\begin{axis}[
name=fidelity,
width=\textwidth,
height=0.5\textwidth,
ybar stacked,
ymin=-0.1,
ymax=1.1,
xtick=data,
xtick style={draw=none},
symbolic x coords={Color,Conflicting,Count,DALL-E \cite{ramesh-dalle},Descriptions,Marcus et al. \cite{marcus-arxiv-2022},Misspellings,Positional,Rarewords,Reddit,Text},
x tick label style={rotate=60},
ytick={0,0.5,1},
yticklabels={0\%, 50\%, 100\%},
legend style={font=\tiny},
legend cell align=left,
legend image code/.code={
    \draw [#1] (0cm, -0.1cm) rectangle (0.2cm, 0.25cm);
},
legend style={
draw=none,
at={(0.5, 1.1)},anchor=south,
legend columns=3,
/tikz/every even column/.append style={column sep=0.2cm}
},
bar width=0.6cm,
]

\addplot+[
    fill=google_blue,
    fill opacity=0.2,
    draw=google_blue
] coordinates {
(Color, 0.610)
(Conflicting, 0.593)
(Count, 0.674)
(DALL-E \cite{ramesh-dalle}, 0.597)
(Descriptions, 0.631)
(Marcus et al. \cite{marcus-arxiv-2022}, 0.568)
(Misspellings, 0.459)
(Positional, 0.575)
(Rarewords, 0.609)
(Reddit, 0.531)
(Text, 0.677)
};

\addplot+[
fill=google_green,
fill opacity=0.2,
draw=google_green
] coordinates {
(Color, 0.390)
(Conflicting, 0.407)
(Count, 0.326)
(DALL-E \cite{ramesh-dalle}, 0.403)
(Descriptions, 0.369)
(Marcus et al. \cite{marcus-arxiv-2022}, 0.432)
(Misspellings, 0.541)
(Positional, 0.425)
(Rarewords, 0.391)
(Reddit, 0.469)
(Text, 0.323)
};

\legend{}
\end{axis}

\end{tikzpicture}
    \caption{Fidelity}
    \label{fig:drawit_vs_glide_detailed_alignment_fidelity}
    \end{subfigure}
    \caption{\name vs GLIDE on \benchmarkname \subref{fig:drawit_vs_glide_detailed_alignment_alignment}) image-text alignment, and \subref{fig:drawit_vs_glide_detailed_alignment_fidelity}) image fidelity.}
    \label{fig:drawit_vs_glide_detailed_alignment}
\end{figure}

\FloatBarrier

\input{fig_resnet}

\clearpage
\begin{figure}
\centering
\begin{subfigure}{.49\textwidth}
  \centering
    \begin{Verbatim}[fontsize=\scriptsize]
    def sample():
      for t in reversed(range(T)):
        # Forward pass to get x0_t from z_t.
        x0_t = nn(z_t, t)
        
        # Static thresholding.
        x0_t = jnp.clip(x0_t, -1.0, 1.0)
        
        
        
        
        
        # Sampler step.
        z_tm1 = sampler_step(x0_t, z_t, t)
        z_t = z_tm1
      return x0_t
    \end{Verbatim}
  \caption{Implementation for static thresholding.}
\end{subfigure}%
\hfill%
\begin{subfigure}{.49\textwidth}
  \centering
  \begin{Verbatim}[fontsize=\scriptsize]
    def sample(p: float):
      for t in reversed(range(T)):
        # Forward pass to get x0_t from z_t.
        x0_t = nn(z_t, t)
        
        # Dynamic thresholding (ours).
        s = jnp.percentile(
            jnp.abs(x0_t), p, 
            axis=tuple(range(1, x0_t.ndim)))
        s = jnp.max(s, 1.0)
        x0_t = jnp.clip(x0_t, -s, s) / s

        # Sampler step.
        z_tm1 = sampler_step(x0_t, z_t, t)
        z_t = z_tm1
      return x0_t
  \end{Verbatim}
  \caption{Implementation for dynamic thresholding.}
\end{subfigure}
\caption{Pseudo code implementation comparing static thresholding and dynamic thresholding.}
\label{fig:clip_vs_scaled_clip_impl}
\end{figure}

\begin{figure}
\centering
\begin{subfigure}{.49\textwidth}
  \centering
  \begin{Verbatim}[fontsize=\scriptsize]
    def train_step(
        x_lr: jnp.ndarray, x_hr: jnp.ndarray):
      # Add augmentation to the low-resolution image.
      aug_level = jnp.random.uniform(0.0, 1.0)
      x_lr = apply_aug(x_lr, aug_level)
      
      # Diffusion forward process.
      t = jnp.random.uniform(0.0, 1.0)
      z_t = forward_process(x_hr, t)
      
      Optimize loss(x_hr, nn(z_t, x_lr, t, aug_level))
  \end{Verbatim}
  \caption{Training using conditioning augmentation.}
\end{subfigure}%
\hfill%
\begin{subfigure}{.49\textwidth}
  \centering
    \begin{Verbatim}[fontsize=\scriptsize]
    def sample(aug_level: float, x_lr: jnp.ndarray):
      # Add augmentation to the low-resolution image.
      x_lr = apply_aug(x_lr, aug_level)

      for t in reversed(range(T)):
        x_hr_t = nn(z_t, x_lr, t, aug_level)

        # Sampler step.
        z_tm1 = sampler_step(x_hr_t, z_t, t)
        z_t = z_tm1
      return x_hr_t
  \end{Verbatim}
  \caption{Sampling using conditioning augmentation. }
\end{subfigure}
\caption{Pseudo-code implementation for training and sampling using conditioning augmentation. Text conditioning has not been shown for brevity.}
\label{fig:cond_aug_impl}
\end{figure}

\FloatBarrier
\newpage
\section{Implementation Details}
\label{sec:impl_details_appendix}

\subsection{$64\times64$}
\textbf{Architecture}: We adapt the architecture used in \cite{dhariwal2021diffusion}. We use larger \emph{embed\_dim} for scaling up the architecture size. For conditioning on text, we use text cross attention at resolutions $[32, 16, 8]$ as well as attention pooled text embedding.

\textbf{Optimizer}: We use the Adafactor optimizer for training the base model. We use the default \href{https://optax.readthedocs.io/en/latest/api.html#adafactor}{optax.adafactor} parameters. We use a learning rate of 1e-4 with 10000 linear warmup steps.

\textbf{Diffusion}: We use the cosine noise schedule similar to \cite{nichol2021improved}. We train using continuous time steps $t \sim \mathcal{U}(0, 1)$.
\begin{Verbatim}[fontsize=\scriptsize]
 # 64 X 64 model.
 architecture = {
      "attn_resolutions": [32, 16, 8],
      "channel_mult": [1, 2, 3, 4],
      "dropout": 0,
      "embed_dim": 512,
      "num_res_blocks": 3,
      "per_head_channels": 64,
      "res_block_type": "biggan",
      "text_cross_attn_res": [32, 16, 8],
      "feature_pooling_type": "attention",
      "use_scale_shift_norm": True,
  }

  learning_rate = optax.warmup_cosine_decay_schedule(
      init_value=0.0,
      peak_value=1e-4,
      warmup_steps=10000,
      decay_steps=2500000,
      end_value=2500000)

  optimizer = optax.adafactor(lrs=learning_rate, weight_decay=0)
  diffusion_params = {
    "continuous_time": True,
    "schedule": {
      "name": "cosine",
    }
  }
\end{Verbatim}

\subsection{$64\times64 \rightarrow 256\times256$}
\textbf{Architecture}: Below is the architecture specification for our $64\times64 \rightarrow 256\times256$ super-resolution model. We use an Efficient U-Net architecture for this model.

\textbf{Optimizer}: We use the standard Adam optimizer with 1e-4 learning rate, and 10000 warmup steps.

\textbf{Diffusion}: We use the same cosine noise schedule as the base $64\times64$ model. We train using continuous time steps $t \sim \mathcal{U}(0, 1)$.

\begin{Verbatim}[fontsize=\scriptsize]
    architecture = {
        "dropout": 0.0,
        "feature_pooling_type": "attention",
        "use_scale_shift_norm": True,
        "blocks": [
            {
              "channels": 128,
              "strides": (2, 2),
              "kernel_size": (3, 3),
              "num_res_blocks": 2,
            },
            {
              "channels": 256,
              "strides": (2, 2),
              "kernel_size": (3, 3),
              "num_res_blocks": 4,
            },
            {
              "channels": 512,
              "strides": (2, 2),
              "kernel_size": (3, 3),
              "num_res_blocks": 8,
            },
            {
              "channels": 1024,
              "strides": (2, 2),
              "kernel_size": (3, 3),
              "num_res_blocks": 8,
              "self_attention": True,
              "text_cross_attention": True,
              "num_attention_heads": 8
            }
        ]
    }

    learning_rate = optax.warmup_cosine_decay_schedule(
        init_value=0.0,
        peak_value=1e-4,
        warmup_steps=10000,
        decay_steps=2500000,
        end_value=2500000)
    
    optimizer = optax.adam(
        lrs=learning_rate, b1=0.9, b2=0.999, eps=1e-8, weight_decay=0)
    
    diffusion_params = {
      "continuous_time": True,
      "schedule": {
        "name": "cosine",
      }
    }
\end{Verbatim}

\subsection{$256\times256 \rightarrow 1024\times1024$}
\textbf{Architecture}: Below is the architecture specification for our $256\times256 \rightarrow 1024\times1024$ super-resolution model. We use the same configuration as the $64\times64 \rightarrow 256\times256$ super-resolution model, except we do not use self-attention layers but rather have cross-attention layers (to the text embeddings). 

\textbf{Optimizer}: We use the standard Adam optimizer with 1e-4 learning rate, and 10000 linear warmup steps.

\textbf{Diffusion}: We use the 1000 step linear noise schedule with start and end set to 1e-4 and 0.02 respectively. We train using continuous time steps $t \sim \mathcal{U}(0, 1)$.

\begin{Verbatim}[fontsize=\scriptsize]
    "dropout": 0.0,
    "feature_pooling_type": "attention",
    "use_scale_shift_norm": true,
    "blocks"=[
        {
          "channels": 128,
          "strides": (2, 2),
          "kernel_size": (3, 3),
          "num_res_blocks": 2,
        },
        {
          "channels": 256,
          "strides": (2, 2),
          "kernel_size": (3, 3),
          "num_res_blocks": 4,
        },
        {
          "channels": 512,
          "strides": (2, 2),
          "kernel_size": (3, 3),
          "num_res_blocks": 8,
        },
        {
          "channels": 1024,
          "strides": (2, 2),
          "kernel_size": (3, 3),
          "num_res_blocks": 8,
          "text_cross_attention": True,
          "num_attention_heads": 8
        }
    ]
\end{Verbatim}

\FloatBarrier

\nocite{de2021diffusion}
\nocite{esser2021taming}
\nocite{van2017neural,razavi2019generating,yu2021vector}
\nocite{ding2021cogview,ramesh-dalle}
\nocite{reed2016generative,wang2018high,xu2018attngan,zhu2019dm,tao2020df}
\nocite{ye2021improving,zhang2021cross,zhou2021lafite}

\end{document}